\documentclass[10pt,journal,compsoc]{IEEEtran}

\usepackage[usenames]{color}
\usepackage[table,dvipsnames]{xcolor}
\usepackage{soul}
\usepackage{amssymb}
\usepackage{makecell}
\usepackage{multirow}
\usepackage{enumitem}
\usepackage{rotating}
\usepackage{array}
\usepackage{graphicx}
\usepackage{psfrag}
\usepackage{subfigure}
\usepackage{rotating}
\usepackage{url}
\usepackage{rotating, graphicx}
\usepackage{supertabular}
\usepackage[graphicx]{realboxes}
\usepackage{adjustbox}
\usepackage{threeparttable}
\usepackage{tikz}
\usepackage{amsmath}
\usepackage{booktabs}
\usepackage{lscape}
\usepackage{verbatim}
\usepackage{longtable}
\usepackage{stfloats}
\usepackage[misc]{ifsym}
\newcommand{\etal}{\textit{et al.}}
\newcommand{\ie}{\emph{i.e., }}
\newcommand{\eg}{\emph{e.g., }}
\ifCLASSOPTIONcompsoc
  \usepackage[nocompress]{cite}
\else
  \usepackage{cite}
\fi

\usepackage[top=0.4in,bottom=0.4in,left=0.5in,textwidth=7.5in]{geometry}

\ifCLASSINFOpdf
\else
\fi

\begin{document}

\title{Deep Learning for Scene Classification: A Survey}

\author{Delu~Zeng,
        Minyu~Liao,
        Mohammad~Tavakolian,
        Yulan~Guo \\
        Bolei~Zhou, Dewen Hu,
        Matti Pietik\"{a}inen, and
        Li Liu
\IEEEcompsocitemizethanks{
\IEEEcompsocthanksitem Delu Zeng (dlzeng@scut.edu.cn) and Minyu Liao (201820127075@mail.-scut.edu.cn) are with the South China University of Technology, China. Mohammad Tavakolian (mohammad.tavakolian@oulu.fi), Matti Pietik\"{a}inen (matti.pietikainen@oulu.fi) and Li Liu (li.liu@oulu.fi) are with the University of Oulu, Finland. Yulan Guo (yulan.guo@nudt.edu.cn), Dewen Hu (dwhu@nudt.edu.cn) and Li Liu are with the National University of Defense Technology. Bolei Zhou (bzhou@ie.cuhk.edu.hk) is with the Chinese University of Hong Kong, China.
\IEEEcompsocthanksitem  Li Liu is the corresponding author. Delu Zeng, Minlu Liao, and Li Liu have equal contribution to this work and are cofirst authors.
}
}


\IEEEtitleabstractindextext{%
\begin{abstract}
Scene classification, aiming at classifying a scene image to one of the predefined scene categories by comprehending the entire image, is a longstanding, fundamental and challenging problem in computer vision. The rise of large-scale datasets, which constitute the corresponding dense sampling of diverse real-world scenes, and the renaissance of deep learning techniques, which learn powerful feature representations directly from big raw data, have been bringing remarkable progress in the field of scene representation and classification. To help researchers master needed advances in this field, the goal of this paper is to provide a comprehensive survey of recent achievements in scene classification using deep learning. More than 200 major publications are included in this survey covering different aspects of scene classification, including challenges, benchmark datasets, taxonomy, and quantitative performance comparisons of the reviewed methods. In retrospect of what has been achieved so far, this paper is also concluded with a list of promising research opportunities.

\end{abstract}

\begin{IEEEkeywords}
Scene classification, Deep learning, Convolutional neural network, Scene representation, Literature survey.
\end{IEEEkeywords}}

\maketitle
\IEEEdisplaynontitleabstractindextext
\IEEEpeerreviewmaketitle

\IEEEraisesectionheading{\section{Introduction}\label{sec:introduction}}

\IEEEPARstart{T}{he} goal of scene classification is to classify a scene image\footnote{A scene is a semantically coherent view of a real-world environment that contains background and multiple objects, organized in a spatially licensed manner \cite{henderson1999high, epstein2005cortical}.} to one of the predefined scene categories (such as beach, kitchen, and bakery), based on the image's ambient content, objects, and their layout. Visual scene understanding requires reasoning about the diverse and complicated environments that we encounter in our daily life. Recognizing visual categories such as objects, actions and events is no doubt the indispensable ability of a visual system. Moreover, recognizing the scene where the objects appear is of equal importance for an intelligent system to predict the context for the recognized objects by reasoning ``What is happening? What will happen next?''. Humans are remarkably efficient at categorizing natural scenes \cite{greene2009briefest, walther2011simple}. However, it is not an easy task for machines due to the scene's semantic ambiguity, and the large intraclass variations caused by imaging conditions like variations in illumination, viewing angle and scale, imaging distance, \emph{etc}. As a longstanding, fundamental and challenging problem in computer vision, scene classification has been an active area of research for several decades, and has a wide range of applications, such as content based image retrieval \cite{vogel2007semantic, zheng2017sift}, robot navigation \cite{zhang2017learning,hou2019deep}, intelligent video surveillance \cite{zhang2012mining, sreenu2019intelligent}, augmented reality \cite{behzadan2011integrated, nee2012augmented}, and disaster detection applications~\cite{muhammad2018early}.

\begin{figure}[!htbp]
\centering
\includegraphics[width =.38\textwidth]{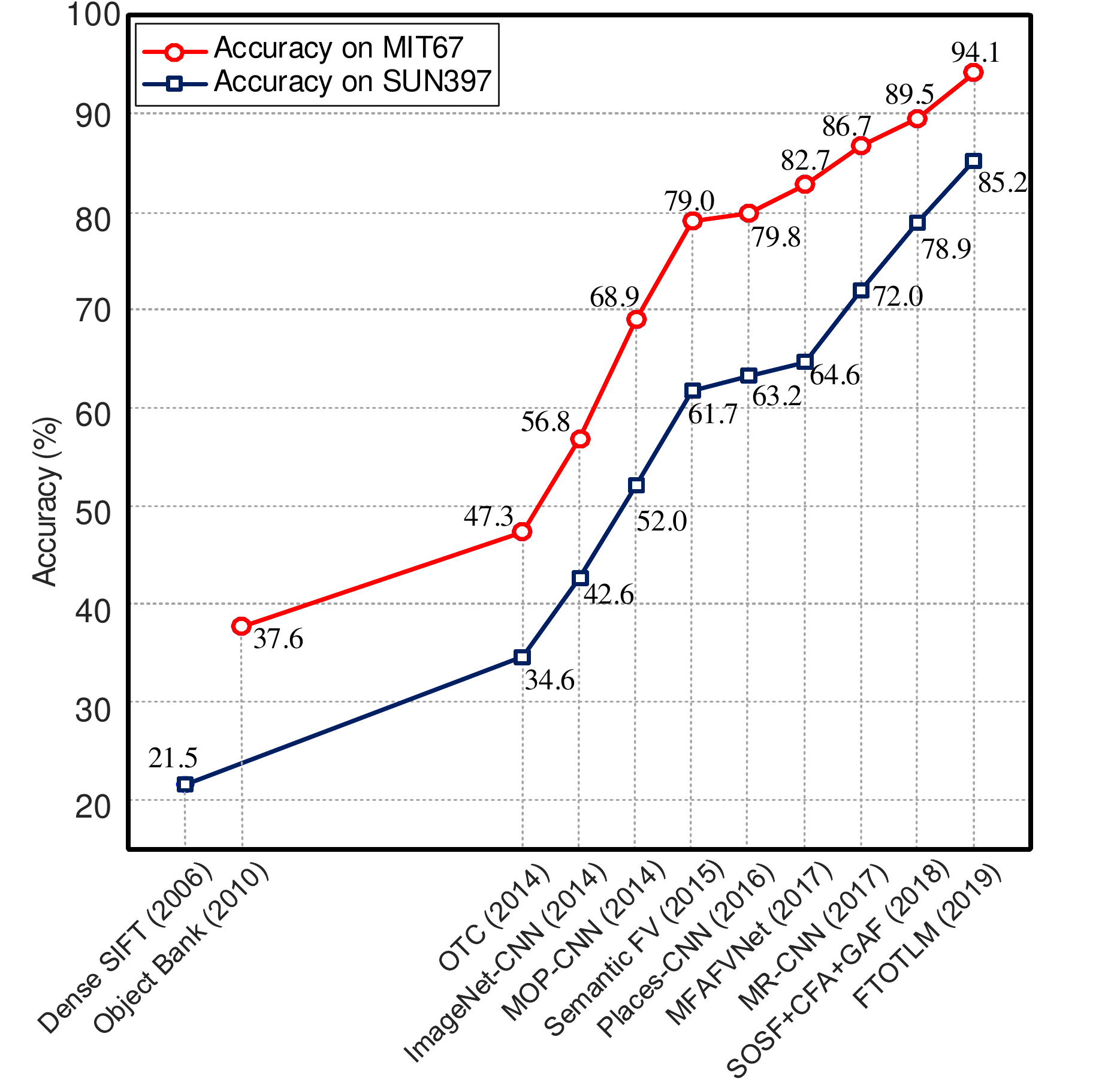}
\\
\caption{A performance overview of scene classification: we can observe a significant improvement on two benchmark datasets since the reignition of deep learning. Dense-SIFT \cite{lazebnik2006beyond}, Object Bank \cite{li2010object} and OTC \cite{margolin2014otc} are handcrafted methods, while the others are deep learning based methods.}
\label{fig:accuracy}
\end{figure}

\begin{figure*}[!htbp]
\centering
\footnotesize
\begin{tikzpicture}[xscale=0.8, yscale=0.36]

\draw [thick, -] (0, 16.5) -- (0, 1); \node [right] at (-0.5, 17) {\textbf{Deep Learning based methods for Scene Classification~(Section~\ref{sec:deeplearning})}};
\draw [thick, -] (0, 16) -- (1, 16);\node [right] at (1, 16) {\textbf{Main CNN Framework~(Section~\ref{ssec:framework})}};
\draw [thick, -] (1.5, 15.5) -- (1.5, 13);
\draw [thick, -] (1.5, 15) -- (2.5, 15);\node [right] at (2.5, 15) {\emph{Pre-trained CNN Model}: Object-centric CNNs~\cite{krizhevsky2012imagenet}, Scene-centric CNNs~\cite{zhou2014learning}};
\draw [thick, -] (1.5, 14) -- (2.5, 14);\node [right] at (2.5, 14) {\emph{Fine-tuned CNN Model}: Scale-specific CNNs~\cite{herranz2016scene}, DUCA~\cite{khan2016discriminative}, FTOTLM~\cite{liu2019novel}};
\draw [thick, -] (1.5, 13) -- (2.5, 13);\node [right] at (2.5, 13) {\emph{Specific CNN Model}: DL-CNN~\cite{liu2018dictionary}, GAP-CNN~\cite{zhou2016learning}, CFA~\cite{sun2018fusing}};

\draw [thick, -] (0, 12) -- (1, 12);\node [right] at (1, 12) {\textbf{CNN based Scene Representation~(Section~\ref{ssec:representation})}};
\draw [thick, -] (1.5, 11.5) -- (1.5, 7);
\draw [thick, -] (1.5, 11) -- (2.5, 11);\node [right] at (2.5, 11) {\emph{Global CNN Feature based Method}: Places-CNN \cite{zhou2014learning, zhou2017places}, S2ICA \cite{hayat2016spatial}, GAP-CNN~\cite{zhou2016learning}, DL-CNN~\cite{liu2018dictionary}, HLSTM~\cite{zuo2016learning}};
\draw [thick, -] (1.5, 10) -- (2.5, 10);\node [right] at (2.5, 10) {\emph{Spatially Invariant Feature based Method}: MOP-CNN \cite{gong2014multi}, MPP-CNN \cite{yoo2015multi}, SFV \cite{dixit2015scene}, MFAFVNet \cite{li2017deep}, VSAD \cite{wang2017weakly}};
\draw [thick, -] (1.5, 9) -- (2.5, 9);\node [right] at (2.5, 9) {\emph{Semantic Feature based Method}: MetaObject-CNN~\cite{wu2015harvesting}, WELDON~\cite{durand2016weldon}, SDO~\cite{cheng2018scene}, M2M BiLSTM~\cite{laranjeira2019modeling}};
\draw [thick, -] (1.5, 8) -- (2.5, 8);\node [right] at (2.5, 8) {\emph{Multi-layer Feature based Method}: DAG \cite{yang2015multi}, Hybrid CNNs \cite{xie2015hybrid}, G-MS2F \cite{tang2017g}, FTOTLM \cite{liu2019novel}};
\draw [thick, -] (1.5, 7) -- (2.5, 7);\node [right] at (2.5, 7) {\emph{Multi-view Feature based Method}: Scale-specific CNNs~\cite{herranz2016scene}, LS-DHM~\cite{guo2016locally}, MR CNN~\cite{wang2017knowledge}};

\draw [thick, -] (0, 6) -- (1, 6);\node [right] at (1, 6) {\textbf{Strategy for Improving Scene Representation~(Section~\ref{ssec:strategy})}};
\draw [thick, -] (1.5, 5.5) -- (1.5, 2);
\draw [thick, -] (1.5, 5) -- (2.5, 5);\node [right] at (2.5, 5) {\emph{Encoding Strategy}: Semantic FV~\cite{dixit2015scene}, FCV~\cite{guo2016locally}, VSAD~\cite{wang2017weakly}, MFA-FS~\cite{dixit2016object},  MFAFVNet~\cite{li2017deep}};
\draw [thick, -] (1.5, 4) -- (2.5, 4);\node [right] at (2.5, 4) {\emph{Attention Strategy}: Channel attention policy~\cite{lopez2020semantic}, Spatial attention policy~\cite{xiong2020msn}};
\draw [thick, -] (1.5, 3) -- (2.5, 3);\node [right] at (2.5, 3) {\emph{Contextual Strategy}: Sequential model~\cite{zuo2016learning}, Graph-related model~\cite{chen2020scene}};
\draw [thick, -] (1.5, 2) -- (2.5, 2);\node [right] at (2.5, 2) {\emph{Regularization Strategy}: Sparse Regularization~\cite{liu2018dictionary}, Structured Regularization~\cite{xiong2020msn}, Supervised Regularization~\cite{guo2016locally}};

\draw [thick, -] (0, 1) -- (1, 1);\node [right] at (1, 1) {\textbf{RGB-D Scene Classification~(Section~\ref{ssec:RGBD})}};
\draw [thick, -] (1.5, 0.5) -- (1.5, -1);
\draw [thick, -] (1.5, 0) -- (2.5, 0);\node [right] at (2.5, 0) {\emph{Depth-specific Feature Learning}: Fine-tuning RGB-CNNs~\cite{wang2016modality}, Weak-supervised learning~\cite{song2017combining}, Semi-supervised learning~\cite{du2019translate} };
\draw [thick, -] (1.5, -1) -- (2.5, -1);\node [right] at (2.5, -1) {\emph{Multiple Modality Fusion}: Feature-level combination~\cite{cheng2017locality},Consistent-feature based fusion~\cite{zhu2016discriminative}, Distinctive-feature based fusion~\cite{xiong2020msn}};

\end{tikzpicture}
\caption{A taxonomy of deep learning for scene classification. With the rise of large-scale datasets, powerful features are learned from pre-trained CNNs, fine-tuned CNNs, or specific CNNs, having made remarkable progress. The major features include global CNN features, spatially invariant features, semantic features, multi-layer features, and multi-view features. Meanwhile, many methods are improved via effective strategies, like encoding, attention learning, and context modeling. As a new issue, methods using RGB-D datasets, focus on learning depth specific features and fusing multiple modalities.} 
\label{fig:overview}
\end{figure*}
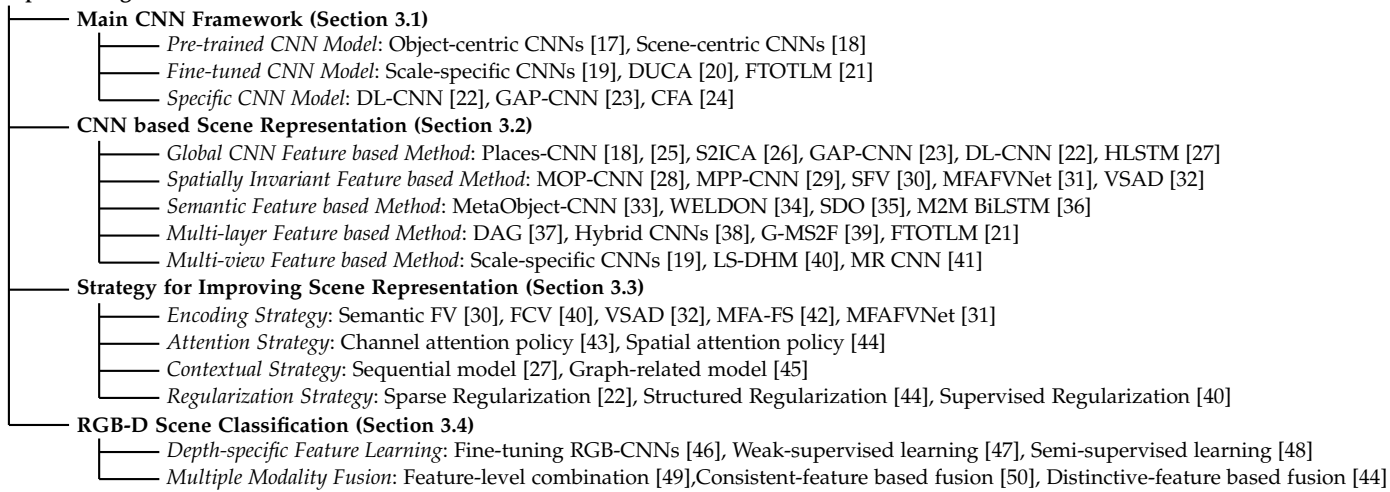

As the core of scene classification, \textbf{scene representation} is the process of  transforming a scene image into its concise descriptors (\ie features), and still attracts tremendous and increasing attention. The recent revival of interest in Artificial Neural Networks (ANNs), particularly deep learning \cite{krizhevsky2012imagenet}, has revolutionized computer vision and been ubiquitously used in various tasks like object classification and detection \cite{girshick2015fast, ren2016faster}, semantic segmentation \cite{badrinarayanan2017segnet, chen2017deeplab, cai2018menet} and scene classification \cite{gong2014multi, zhou2014learning}. In 2012, object classification with the large-scale ImageNet dataset \cite{deng2009imagenet} achieved a significant breakthrough in performance by a Deep Neural Network (DNN) named AlexNet \cite{krizhevsky2012imagenet}, which is arguably what reignited the field of ANNs and triggered the recent revolution in computer vision. Since then, research focus on scene classification has begun to move away from handcrafted feature engineering to deep learning, which can learn powerful representations directly from data. Recent advances in deep learning have opened the possibility of scene classification towards the datasets \emph{of large scale} and \emph{in the wild} \cite{zhou2014learning, xiao2010sun, zhou2017places}, and many scene representations \cite{gong2014multi, cinbis2015approximate, dixit2015scene, liu2019novel} have been proposed. As illustrated in Fig.~\ref{fig:accuracy}, deep learning has brought significant improvements in scene classification. Given the exceptionally rapid rate of progress, this article attempts to track recent advances and summarize their achievements to gain a clearer picture of the current panorama in scene classification using deep learning.

Recently, several surveys for scene classification have also been available, such as \cite{wei2016visual, cheng2017remote, xie2020scene}.  Cheng \etal~\cite{cheng2017remote} provided a recent comprehensive review of the recent progress for remote sensing image scene classification. Wei \etal~\cite{wei2016visual} carried out an experimental study of 14 scene descriptors mainly in the handcrafted feature engineering way for scene classification. Xie \etal~\cite{xie2020scene} reviewed scene recognition approaches in the past two decades, and most of discussed methods in their survey appeared in this handcrafted way. \emph{As opposed to} these existing reviews \cite{cheng2017remote, wei2016visual, xie2020scene}, this work herein summarizes the striking success and dominance in indoor/outdoor scene classification using deep learning and its related methods, but not including other scene classification tasks, \eg remote sensing scene classification \cite{nogueira2017towards, hu2015transferring, cheng2017remote}, acoustic scene classification \cite{mesaros2016tut, ren2018deep}, place classification~\cite{lowry2015visual, arandjelovic2016netvlad}, \emph{etc}. The major contributions of this work can be summarized as follows:
\begin{itemize}
  \item As far as we know, this paper is the first to specifically focus on deep learning methods for indoor/outdoor scene classification, including RGB scene classification, as well as RGB-D scene classification.
  \item We present a taxonomy (see Fig.~\ref{fig:overview}), covering the most recent and advanced progresses of deep learning for scene representation.
  \item Comprehensive comparisons of existing methods on several public datasets are provided, meanwhile we also present brief summaries and insightful discussions.
\end{itemize}
The remainder of this paper is organized as follows: Challenges and benchmark datasets are summarized in Section \ref{sec:background}. In section \ref{sec:deeplearning}, we present a taxonomy of the existing deep learning based methods. Then in section \ref{sec:performance}, we provide an overall discussion of their performance (Tables~\ref{tab:RGBresults},~\ref{tab:RGBDresults}). Followed by Section~\ref{sec:conclusion} we conclude important future research outlook.

\section{Background}
\label{sec:background}

\subsection{The Problem and Challenge}
\label{sec:problems}

Scene classification can be further dissected through analyzing its strong ties with related vision tasks, such as object classification and texture classification. As typical pattern recognition problems, these tasks all consist of feature representation and classification. However, in contrast to object classification (images are object-centric) or texture classification (images include only textures), the scene images are more complicated, and it is essential to further explore the content of scene, \eg what the semantic parts (\eg objects, textures, background) are, in what way they are organized together, and what their semantic connections with each other are. Despite over several decades of development in scene classification (shown in the appendix due to space limit), most of methods still have not been capable of performing at a level sufficient for various real-world scenes. The inherent difficulty is due to the nature of complexity and high variance of scenes. Overall, significant challenges in scene classification stem from large intraclass variations, semantic ambiguity, and computational efficiency.

\begin{figure}[!htbp]
\centering
\includegraphics[width = .48\textwidth]{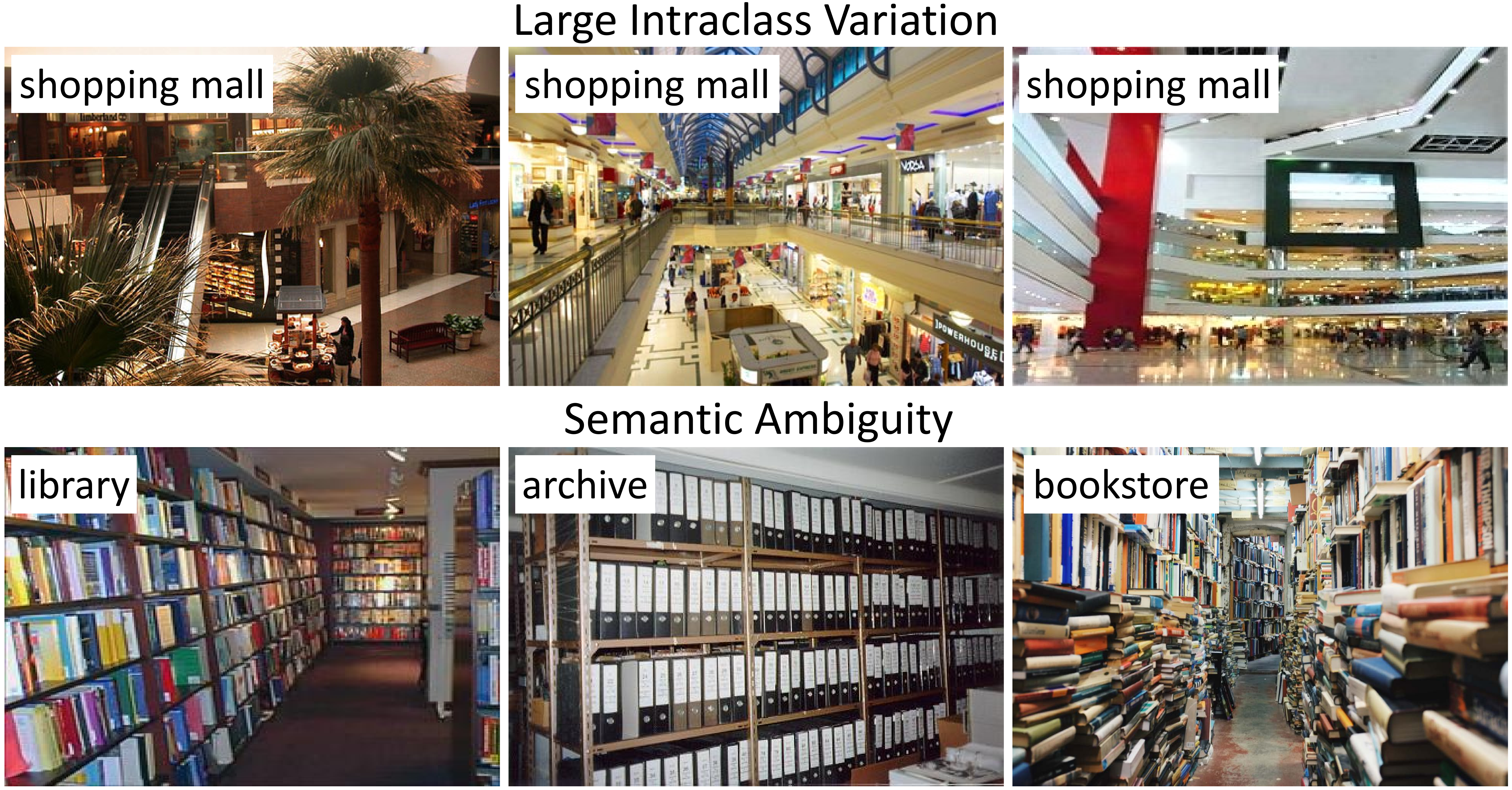}
\caption{Illustrations of \emph{large intraclass variation} and \emph{semantic ambiguity}. Top (large intraclass variation): The shopping malls are quite different caused by lighting and overall content. Below (semantic ambiguity): General layout and uniformly arranged objects are similar on archive, bookstore, and library.}
\label{fig.distance}
\end{figure}

\textbf{Large intraclass variation.} Intraclass variation mainly originates from intrinsic factors of the scene itself and imaging conditions. In terms of intrinsic factors, each scene can have many different example images, possibly varying with large variations among various objects, background, or human activities. Imaging conditions like changes in illumination, viewpoint, scale and heavy occlusion, clutter, shading, blur, motion, \emph{etc}. contribute to large intraclass variations. Further challenges may be added by digitization artifacts, noise corruption, poor resolution, and filtering distortion. For instance, three shopping malls (top row of Fig.~\ref{fig.distance}) are shown with different lighting conditions, viewing angle, and objects.

\textbf{Semantic ambiguity.} Since images of different classes may share similar objects, textures, background, \emph{etc.}, they look very similar in visual appearances, which causes ambiguity among them~\cite{boutell2004learning, lopez2020semantic}. The bottom row of Fig.~\ref{fig.distance} depicts strong visual correlation between three different indoor scenes, \ie archive, bookstore, and library. With the emerging of new scene categories, the problem of semantic ambiguity would be more serious. In addition, scene category annotation is subjective, relying on the experience of the annotators, therefore a scene image may belong to multiple semantic categories \cite{boutell2004learning, zhang2007multi}.

\textbf{Computational efficiency.} The prevalence of social media networks and mobile/wearable devices has led to increasing demands for various computer vision tasks including scene recognition. However, mobile/wearable devices have constrained computing related resources, making efficient scene recognition a pressing requirement.

\subsection{Datasets}
\label{sec:Datasets}

\begin{figure}[!htbp]
\centering
\includegraphics[width = 0.48\textwidth]{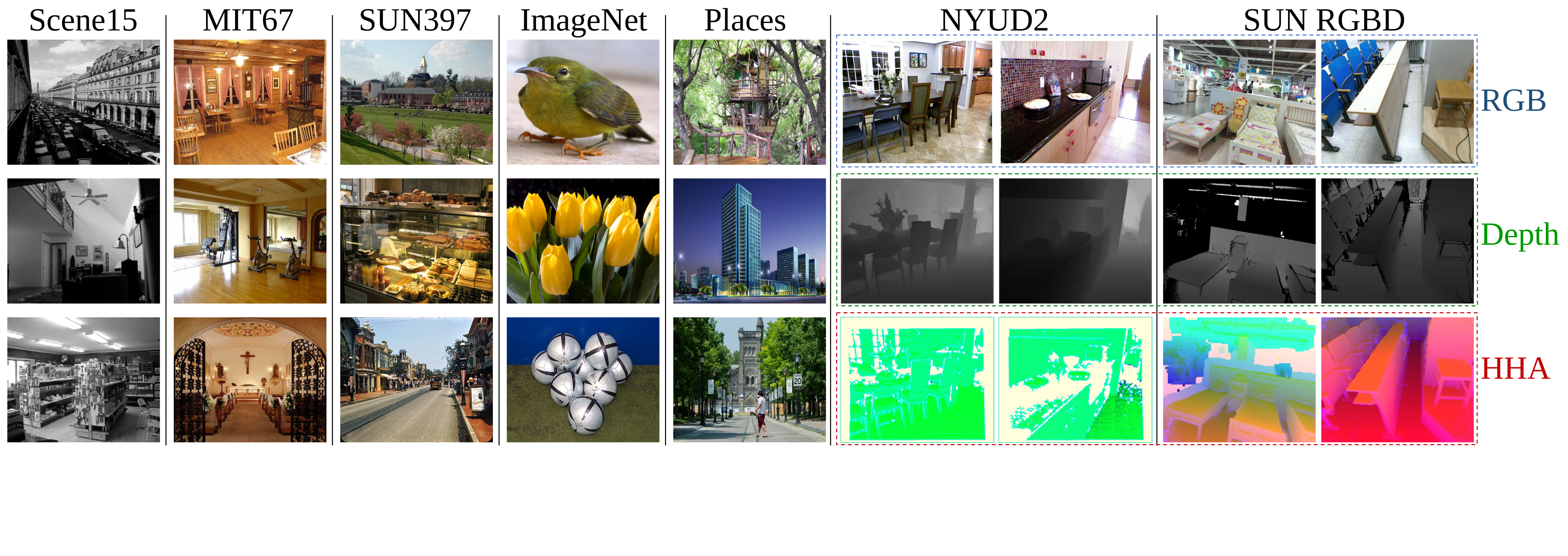}
\caption{Some example images for scene classification from benchmark datasets for a summary of these datasets. RGB-D images consist of RGB and a depth map. Moreover, Gupta \etal~\cite{gupta2014learning} proposed to convert depth image into three-channel feature maps, \ie Horizontal disparity, Height above the ground, and Angle of the surface norm (HHA). Such HHA encoding is useful for the visualization of depth data.}
\label{fig:datasets}
\end{figure}

This section reviews publicly available datasets for scene classification. The scene datasets (see Fig.~\ref{fig:datasets}) are broadly divided into two main categories based on the image type: RGB and RGB-D datasets. The datasets can further be divided into two categories in terms of their size. Small-size datasets (\eg Scene15~\cite{lazebnik2006beyond}, MIT67~\cite{quattoni2009recognizing}, SUN397~\cite{xiao2010sun}, NYUD2~\cite{silberman2012indoor}, SUN RGBD~\cite{song2015sun}) are usually used for evaluation, while large-scale datasets, \eg ImageNet~\cite{deng2009imagenet} and Places~\cite{zhou2014learning, zhou2017places}, are essential for pre-training and developing deep learning models. Table~\ref{tab:datasets} summarizes the characteristics of these datasets for scene classification.

\renewcommand\arraystretch{1.10}
\begin{table}[htbp]
  \centering
  \caption{Popular datasets for scene classification. ``\#'' denotes \emph{the number of}.}
    \setlength\tabcolsep{1pt}
    \resizebox*{9.3cm}{!}{
    \begin{tabular}{|m{3.1em}<{\centering}|m{7.3em}<{\centering}|m{5.1em}<{\centering}|m{3.1em}<{\centering}|m{5em}<{\centering}|m{10em}<{\centering}|}

    \hline
    Type  & Dataset &  \#Images & \#Class  & Resolution & Class label \\
    \hline
    \multirow{8}[6]{*}{RGB}  & Scene15 \cite{lazebnik2006beyond}  & 4,488  & 15     & $\thickapprox$ 300$\times$250 & Indoor/outdoor scene \\
\cline{2-6}    &       MIT67 \cite{quattoni2009recognizing} & 15,620  & 67    & $\ge$  200$\times$200 & Indoor scene \\
\cline{2-6}    &       SUN397 \cite{xiao2010sun} & 108,754  & 397    &  $\thickapprox$ 500$\times$300 & Indoor/outdoor scene \\
\cline{2-6}    & ImageNet \cite{krizhevsky2012imagenet} & 14 million+ & 21,841   & $\thickapprox$ 500$\times$400 & Object \\
\cline{2-6}    &       Places205 \cite{zhou2014learning} & 7,076,580  & 205    & $\ge$ 200$\times$200 & Indoor/outdoor scene \\
\cline{2-6}    &Places88 \cite{zhou2014learning} & $-$  & 88    & $\ge$ 200$\times$200 & Indoor/outdoor scene \\
\cline{2-6}    &Places365-S \cite{zhou2017places} & 1,803,460  & 365  & $\ge$ 200$\times$200 & Indoor/outdoor scene \\
\cline{2-6}    &Places365-C \cite{zhou2017places} & 8 million+  & 365   & $\ge$ 200$\times$200 & Indoor/outdoor scene \\

\hline
\multirow{2}[2]{*}{RGB-D} & NYUD2 \cite{silberman2012indoor}  & 1,449  & 10   &  $\thickapprox$ 640$\times$480 & Indoor scene \\
\cline{2-6}    & SUN RGBD \cite{song2015sun} & 10,355  & 19    & $\ge$512$\times$424 & Indoor scene \\
    \hline
    \end{tabular}%
    }
  \label{tab:datasets}%
\end{table}%

\textbf{Scene15} dataset~\cite{lazebnik2006beyond} is a small scene dataset containing 4,448 grayscale images of 15 scene categories, \ie 5 indoor scene classes (\eg office, store, and kitchen) along with 10 outdoor scene classes (like suburb, forest, and tall building). Each class contains 210$-$410 scene images, and the image size is around 300$\times$250. The dataset is divided into two splits; there are at least 100 images per class in the training set, and the rest are for testing.

\textbf{MIT Indoor 67 (MIT67)} dataset~\cite{quattoni2009recognizing} covers a wide range of indoor scenes, \eg store, public space, and leisure. MIT67 comprises 15,620 scene images from 67 indoor categories, where each category has about 100 images. Moreover, all images have a minimum resolution of 200$\times$200 pixels on the smallest axis. Because of the shared similarities among objects in this dataset, the classification of images is challenging. There are 80 and 20 images per class in the training and testing set, respectively.

\textbf{Scene UNderstanding 397 (SUN397)} dataset~\cite{xiao2010sun} consists of 397 scene categories, in which each category has more than 100 images. The dataset contains 108,754 images with an image size of about 500$\times$300 pixels. SUN397 spans over 175 indoor, 220 outdoor scene classes, and two classes with mixed indoor and outdoor images, \eg a promenade deck with a ticket booth. There are several train/test split settings with 50 images per category in the testing.

\textbf{ImageNet} dataset~\cite{deng2009imagenet} is one of the most famous large-scale image databases particularly used for visual tasks. It is organized in terms of the WordNet~\cite{miller1995wordnet} hierarchy, each node of which is depicted by hundreds and thousands of images. Up to now, there are more than 14 million images and about 20 thousand notes in the ImageNet. Usually, a subset of ImageNet dataset (about 1000 categories with a total of 1.2 million images~\cite{krizhevsky2012imagenet}) is used to pre-train the CNN for scene classification.

\textbf{Places} dataset~\cite{zhou2014learning, zhou2017places} is a large-scale scene dataset with 434 scene categories, which provides an exhaustive list of the classes of environments encountered in the real world. The Places dataset has inherited the same list of scene categories from SUN397~\cite{xiao2010sun}. Four benchmark subsets of Places are shown as follows: 1) \textbf{Places205}~\cite{zhou2014learning} has 2.5 million images from scene categories. The image number per class varies from 5,000 to 15,000. The training set has 2,448,873 images, with 100 images per category for validation and 200 images per category for testing. 2) \textbf{Places88}~\cite{zhou2014learning} contains the 88 common scene categories among the ImageNet~\cite{deng2009imagenet}, SUN397~\cite{xiao2010sun}, and Places205 datasets. Places88 includes only the images obtained in the second round of annotation from the Places. 3) \textbf{Places365-Standard}~\cite{zhou2017places} has 1,803,460 training images with the image number per class varying from 3,068 to 5,000. The validation set has 50 images/class, while the testing set has 900 images/class. 4) \textbf{Places365-Challenge} contains the same categories as the Places365-Standard, but its training set is significantly larger with a total of 8 million images. This subset was released for the Places Challenge held in conjunction with ECCV, as part of the ILSVRC 2016 Challenge.

\textbf{NYU-Depth V2 (NYUD2)} dataset~\cite{silberman2012indoor} is comprised of video sequences from a variety of indoor scenes as recorded by both the RGB and depth cameras. The dataset consists of 1,449 densely labeled pairs of aligned RGB and depth images from 27 indoor scene categories. It features 464 scenes taken from 3 cities and 407,024 unlabeled frames. With the publicly available split, NYUD2 for scene classification offers 795 images for training while 654 images for testing.

\textbf{SUN RGBD} dataset~\cite{song2015sun} consists of 10,335 RGB-D images with dense annotations in both 2D and 3D, for both objects and rooms. The dataset is collected by four different sensors at a similar scale as PASCAL VOC~\cite{everingham2010pascal}. The whole dataset is densely annotated and includes 146,617 2D polygons and 58,657 3D bounding boxes with accurate object orientations, as well as a 3D room layout and category for scenes.

\begin{figure*}[!htbp]
\centering
\includegraphics[width = 0.98\textwidth]{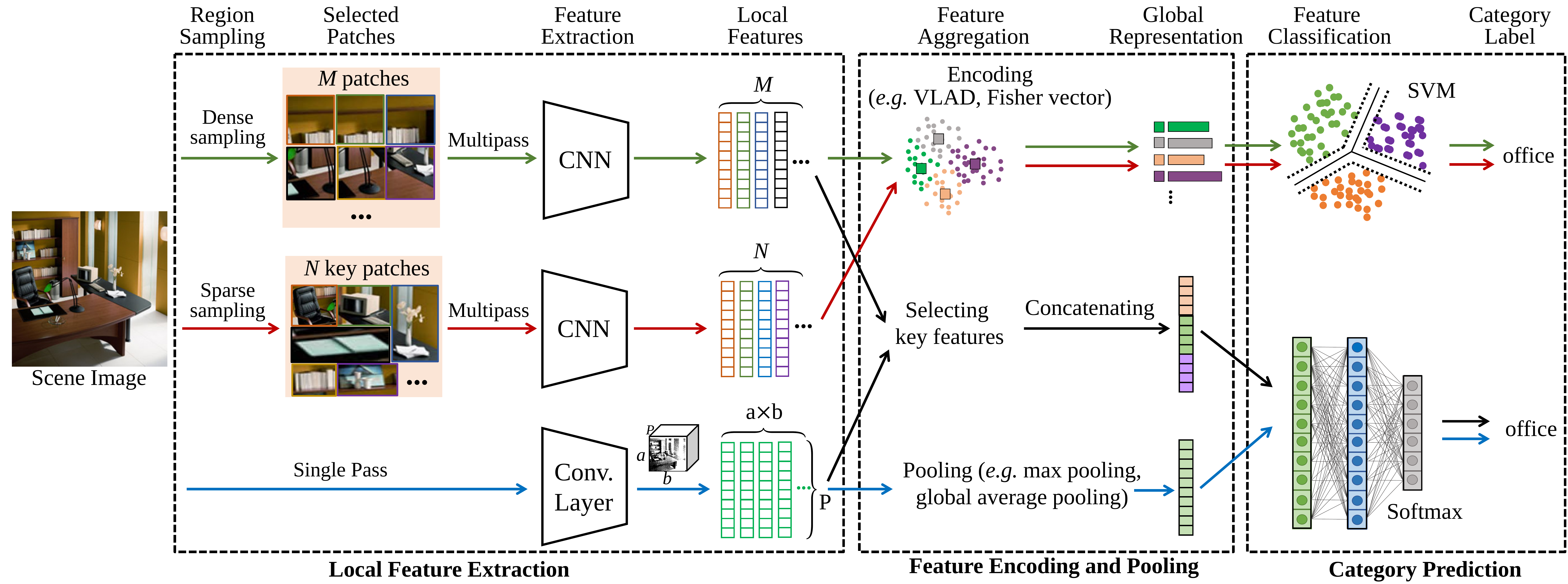}
\caption{Generic pipeline of deep learning for scene classification. An entire pipeline consists of a module in each of the three stages (local feature extraction, feature encoding and pooling, and category prediction). The common pipelines are shown with arrows in different colors, including global CNN feature based pipeline (blue arrows), spatially invariant feature based pipeline (green arrows), and semantic feature based pipeline (red arrows). Although the pipeline of some methods (like \cite{li2017deep, wang2017weakly}) are unified and trained in an end-to-end manner, they are virtually composed of these three stages.}
\label{fig:pipeline}
\end{figure*}

\section{Deep Learning based Methods}
\label{sec:deeplearning}
In this section, we present a comprehensive review of deep learning methods for scene classification. A brief introduction of deep learning is provided in the appendix due to limit space. The most common deep learning architecture is Convolutional Neural Network (CNN)~\cite{lecun2015deep}.  With CNN as feature extractor, Fig.~\ref{fig:pipeline} shows the generic pipeline of most CNN based methods for scene classification. Almost without exception, given an input scene image, the first stage is to use CNN extractors to obtain local features. Then, the second process is to aggregate these features into an image-level representation via encoding, concatenating, or pooling. Finally, with the representation as input, the classification stage is to get a predicted category.

The taxonomy, shown in Fig.~\ref{fig:overview}, covers different aspects of deep learning for scene classification. In the following investigation, we firstly study the main CNN frameworks for scene classification. Then, we review existing CNN based scene representations. Furthermore, we explore various techniques for improving the obtained representations. Finally, as a supplement, we investigate scene classification using RGB-D data.


\subsection{Main CNN Framework}
\label{ssec:framework}

Convolutional Neural Networks (CNNs) are common deep learning models to extract high quality representation. At the beginning, limited by computing resources and labeled data, scene features are extracted from pre-trained CNNs, which is usually combined to BoVW pipeline~\cite{zheng2017sift}. Then, fine-tuned CNN models are used to keep last layers more data-specific. Alternatively, specific CNN models have emerged to adapt to scene attributes.

\subsubsection{Pre-trained CNN Model}
\label{sssec:pretrainedCNN}

The network architecture plays a pivotal role in the performance of deep models. In the beginning, AlexNet~\cite{krizhevsky2012imagenet} served as the mainstream CNN model for feature representation and classification purposes. Later, Simonyan~\etal~\cite{simonyan2014very} developed VGGNet and showed that, for a given receptive field, using multiple stacked small kernels is better than using a large convolution kernel, because applying non-linearity on multiple feature maps yields more discriminative representations. On the other hand, the reduction of kernels’ receptive filed size decreases the number of parameters for bigger networks. Therefore, VGGNet has 3$\times$3 convolution kernels instead of large convolution kernels (\ie 11$\times$11, 7$\times$7, and 5$\times$5) in AlexNet. Motivated by the idea that only a handful of neurons have an effective role in feature representation, Szegedy~\etal~\cite{szegedy2015going} proposed an Inception module to make a sparse approximation of CNNs. Deeper the model, the more descriptive representations. This is the advantage of hierarchical feature extraction using CNN. However, constantly increasing CNN’s depth could result in gradient vanishing. To address this issue, He~\etal~\cite{he2016deep} included skip connection to the hierarchical structure of CNN and proposed Residual Networks (ResNets), which are easier to optimize and can gain accuracy from considerably increased depth. 

In addition to the network architecture, the performance of CNN interwinds with a sufficiently large amount of training data. However, the training data are scarce in certain applications, which results in the under-fitting of the model during the training process. To overcome this issue, pre-trained models can be employed to effectively extract feature representations of small datasets~\cite{girshick2014rich}. Training CNN on large-scale datasets, such as the ImageNet~\cite{deng2009imagenet} and the Places~\cite{zhou2014learning, zhou2017places}, makes them learn enriched visual representations. Such models can further be used as pre-trained models for other tasks. However, the effectiveness of the employment of pre-trained models largely depends on the similarity between the source and target domains. Yosinski \etal~\cite{yosinski2014transferable} documented that the transferability of pre-trained CNN models decreases as the similarity of the target task and original source task decreases. Nevertheless, pre-trained models still have better performance than random initialization of the models~\cite{yosinski2014transferable}.

Pre-trained CNNs, as fixed feature extractors, are divided into two categories: object-centric and scene-centric CNNs. Object-centric CNNs refer to the model pre-trained on object datasets, \eg the ImageNet~\cite{deng2009imagenet}, and deployed for scene classification. Since object images do not contain the diversity provided by the scene~\cite{zhou2014learning}, object-centric CNNs have limited performance for scene classification. Hence, scene-centric CNNs, pre-trained on scene images, like Places~\cite{zhou2014learning, zhou2017places}, are more effective to extract scene-related features.

\textbf{Object-centric CNNs.} Cimpoi~\etal~\cite{cimpoi2015deep} asserted that the feature representations obtained from object-centric CNNs are object descriptors since they have likely more object descriptive properties. The scene image is represented as a bag of semantics~\cite{dixit2015scene}, and object-centric CNNs are sensitive to the overall shape of objects, so many methods~\cite{gong2014multi, liu2014encoding, dixit2015scene, cimpoi2015deep, li2017deep} used object-centric CNNs to extract local features from different regions of the scene image. Another important factor in the effective deployment of object-centric CNNs is the relational size of images in the source and target datasets. Although CNNs are generally robust against size and scale, the performance of object-centric CNNs is influenced by scaling because such models are originally pre-trained on datasets to detect and/or recognize objects. Therefore, the shift to describing scenes, which have multiple objects with different scales, would drastically affect their performance~\cite{herranz2016scene}. For instance, if the image size of the target dataset is smaller than the source dataset to a certain degree, the accuracy of the model would be compromised.

\textbf{Scene-centric CNNs.} Zhou~\etal~\cite{zhou2014learning, zhou2017places} demonstrated the classification performance of scene-centric CNNs is better than object-centric CNNs since the former use the prior knowledge of the scene. Herranz~\etal~\cite{herranz2016scene} found that Places-CNNs~\cite{zhou2016learning} achieve better performance at larger scales; therefore, scene-centric CNNs generally extract the representations in the whole range of scales. Guo~\etal~\cite{guo2016locally} noticed that the CONV layers of scene-centric CNNs capture more detail information of a scene, such as local semantic regions and fine-scale objects, which is crucial to discriminate the ambiguous scenes, while the feature representations obtained from the FC layers do not convey such perceptive quality. Zhou~\etal~\cite{bolei2015object} showed that scene-centric CNNs may also perform as object detectors without explicitly being trained on object datasets.

\subsubsection{Fine-tuned CNN Model}
\label{sssec:finetunedCNN}

Pre-trained CNNs, described in Section~\ref{sssec:pretrainedCNN}, perform as feature extractor with prior knowledge of the training data~\cite{zheng2017sift, liu2019bow}. However, using only the pre-training strategy would prevent exploiting the full capability of the deep models in describing the target scenes adaptively. Hence, fine-tuning the pre-trained CNNs using the target scene dataset improves their performance by reducing the possible domain shift between two datasets~\cite{liu2019bow}. Notably, a suitable weight initialization becomes very important, because it is quite difficult to train a model with many adjustable parameters and non-convex loss functions~\cite{sutskever2013importance}. Therefore, fine-tuning the pre-trained CNN contributes to the effective training process~\cite{liu2014learning, dixit2015scene, yoo2015multi, durand2016weldon}.

For CNNs, a common fine-tuning technique is the \emph{freeze strategy}. In this method, the last FC layer of a pretrained model is replaced with a new FC layer with the same number of neurons as the classes in the target dataset (\ie MIT67, SUN397), while the previous CONV layers’ parameters are frozen, \ie they are not updated during fine-tuning. Then, this modified CNN is fine-tuned by training on the target dataset. Herein, the back-propagation is stopped after the last FC layers, which allows these layers to extract discriminative features from the previous learned layers. Through updating few parameters, training a complex model using small datasets would be affordable. Optionally, it is also possible to gradually unfreeze some layers to further enhance the learning quality as the earlier layers would adapt new representations from the target dataset. Alternatively, different learning rates could be assigned to different layers of CNN, in which the early layers of the model have very low learning rate and the last layers have higher learning rates. In this way, the early CONV layers that have more abstract representations are less affected, while the specialized FC layers are fine-tuned with higher speed.

Small-size training dataset limits the effective fine-tuning process, while \emph{data augmentation} is one alternative to deal with this issue~\cite{chatfield2014return, khan2016discriminative, liu2019novel, du2019translate}. Liu \etal~\cite{liu2019bow} indicated that deep models may not benefit from fine-tuning on a small target dataset. In addition, fine-tuning may have negative effects since the specialized FC layers are changed while inadequate training data are provided for fine-tuning. To this end, Khan~\etal~\cite{khan2016discriminative} augmented the scene image dataset with flipped, cropped, and rotated versions to increase the size of the dataset and further improve the robustness of the learned representations. Liu~\etal~\cite{liu2019novel} proposed a method to select representative image patches of the original image.

There exists a problem via data augmentation to fine-tune CNNs for scene classification. Herranz \etal~\cite{herranz2016scene} asserted that fine-tuning a CNN model have certain ``equalizing'' effect between the input patch scale and final accuracy, \ie to some extent, with too small patches as CNN inputs, the final classification accuracy is worse. This is because the small patch inputs contain insufficient image information, while the final labels indicate scene categories~\cite{wang2017weakly, song2018learning}. Moreover, the number of cropped patches is huge, so just a tiny part of these patches is used to fine-tune CNN models, rendering limited overall improvement~\cite{herranz2016scene}. On the other hand, Herranz \etal~\cite{herranz2016scene} also explored the effect of fine-tuning CNNs on different scales, \ie with different scale patches as inputs. From the practical results, there is a moderate accuracy gain in the range of scale patches where the original CNNs perform poorly, \eg in the cases of global scales for ImageNet-CNN and local scales for Places-CNN. However, there is marginal or no gain in ranges where CNN have already strong performance. For example, since Places-CNN has the best performance in the whole range of scale patches, in this case, fine-tuning on target dataset leads to negligible performance improvement.

\subsubsection{Specific CNN Model}
\label{sssec:specific}

\begin{figure}[!htbp]
\centering
\includegraphics[width = .48\textwidth]{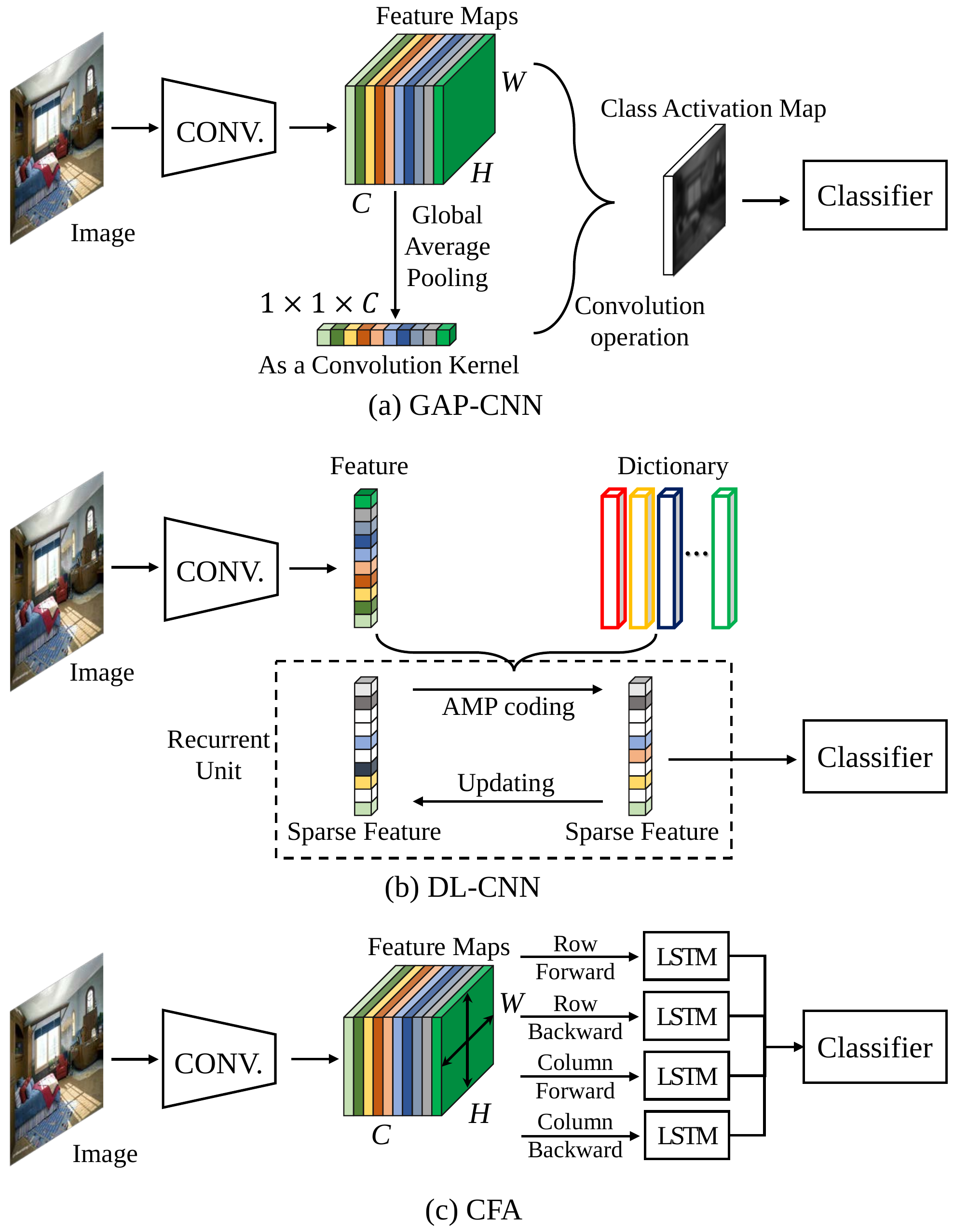}
\caption{Illustrations of three typical specific CNN models. (a) In GAP-CNN~\cite{zhou2016learning}, to reduce parameters of the standard CNN, FC layers are removed, and GAP layer is introduced to form Class Activation Maps (CAMs). (b) In DL-CNN~\cite{liu2018dictionary}, to reduce parameters of CNN model and obtain enhanced sparse features, Dictionary Learning (DL) layers are proposed to replace FC layers. (c) In CFA~\cite{sun2018fusing}, Sun \etal~bypassed four directional LSTM layers on the CONV maps to capture contextual information. 
}
\label{fig:specific}
\end{figure}

In addition to the generic CNN models, \ie pre-trained CNN models and the fine-tuned CNN models, another group of deep models are specifically designed for scene classification. These models are specifically developed to extract effective scene representations from the input by introducing new network architectures. As is shown in Fig.~\ref{fig:specific}, we only show four typical specific models~\cite{liu2018dictionary, zhou2016learning, sun2018fusing, hayat2016spatial}.

To capture discriminative information from regions of interest, Zhou~\etal~\cite{zhou2016learning} replaced the FC layers in a CNN model with a Global Average Pooling (GAP) layer~\cite{lin2013network} followed by a Softmax layer, \ie GAP-CNN. As shown in Fig.~\ref{fig:specific} (a), by a simple combination of the original GAP layer and the $1\times1$ convolution operation to form a class activation map (CAM), GAP-CNN can focus on class-specific regions and perform scene classification well. Although the GAP layer has a lower number of parameters than the FC layer~\cite{zhou2016learning, du2019translate}, the GAP-CNN can obtain comparable classification accuracy.

Hypothesizing that a certain amount of sparsity improves the discriminability of the feature representations~\cite{sun2013learning, shih2014learning, sun2015deeply}, Liu~\etal~\cite{liu2018dictionary} proposed a sparsity model named Dictionary Learning CNN (DL-CNN), seen in Fig.~\ref{fig:specific} (b). They replaced FC layers with new dictionary learning layers, which are composed of a finite number of recurrent units that correspond to iteration processes in the Approximate Message Passing~\cite{donoho2009message}. In particular, these dictionary learning layers’ parameters are updated through back-propagation in an end-to-end manner.

Since the CONV layers perform local operations on small patches of the image, they are not able to explicitly describe the contextual relation between different regions of the scene image. To address this limitation, Sun~\etal~\cite{sun2018fusing} proposed Contextual features in Appearance (CFA) based on LSTM~\cite{hochreiter1997long}. As shown in Fig.~\ref{fig:specific} (c), CONV feature maps are regarded as the input of LSTM layers, which is transformed into four directed sequences in an acyclic way. Finally, LSTM layers are used to describe spatial contextual dependencies, and the output of four LSTM modules are concatenated to describe contextual relations in appearance.

Sequential operations of CONV and FC layers in standard CNNs retain the global spatial structure of the image, which shows global features are sensitive to geometrical variations \cite{gong2014multi, li2019mapnet}, \eg object translations and rotation directly affect the obtained deep features, which drastically limits the application of these features for scene classification. To achieve geometric invariance, Hayat~\etal~\cite{hayat2016spatial} designed a spatial unstructured layer via shuffling the original position of the original feature maps by swapping adjacent diagonal image blocks.

Back-propagation algorithm is the essence of CNN training. It is the practice of fine-tuning the weights of a neural net based on the error rate (\ie loss) obtained in the previous epoch (\ie iteration). Proper tuning of the weights ensures lower error rates, making the model reliable by increasing its generalization. Therefore, many approaches \cite{arandjelovic2016netvlad, li2019mapnet, durand2016weldon, yang2015multi} developed new layers with parameters that can be updated via back-propagation.  The end-to-end system is trained via back-propagation in a holistic manner, which has been proved as a powerful training manner in various domains, and scene classification is no exception. Many methods~\cite{arandjelovic2016netvlad, durand2016weldon, li2017deep, li2019mapnet, yang2015multi} are training in an end-to-end manner. According to our investigations, theoretically, these models can learn more discriminative information through end-to-end optimization; however, the optimization results may fall into bad local optima \cite{mirowski2016learning, li2019mapnet}, so methods training in a multi-stage manner may achieve better results in some case.

\subsection{CNN based Scene Representation}
\label{ssec:representation}

Scene representation, the core of scene classification, has been the focus of this research. Hence, many methods have been put forward for effective scene representations, broadly divided into five categories: global CNN features, spatially invariant features, semantic features, multi-layer features, and combined features, \ie multi-view features.

\begin{figure*}[!htbp]
\centering
\includegraphics[width = 0.98\textwidth]{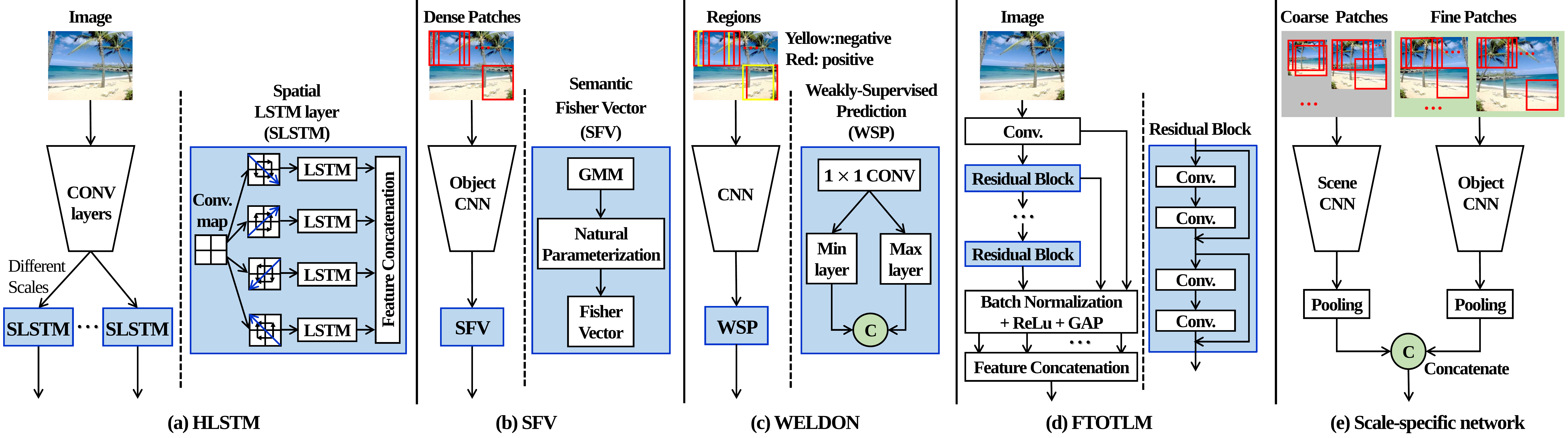}
\caption{Five typical architectures to extract CNN based scene representations (see Section~\ref{ssec:representation}), respectively. Hourglass architectures are backbone networks, such as AlexNet or VGGNet. (a) HLSTM~\cite{zuo2016learning}, a global CNN feature based method, extracts deep feature from the whole image. Spatial LSTM is used to model 2D characteristics among the spatial layout of image regions. Moreover, Zuo \etal~captured cross-scale contextual dependencies via multiple LSTM layers. (b) SFV~\cite{dixit2015scene}, a spatially invariant feature based method, extract local features from dense patches. The highlight of SFV is to add a natural parameterization to transform the semantic space into a natural parameter space. (c) WELDON~\cite{durand2016weldon}, a semantic feature based method, extracts deep features from top evidence (red) and negative instances (yellow). In WSP scheme, Durand~\etal~used the max layer and min layer to select positive and negative instances, respectively. (d) FTOTLM~\cite{liu2019novel}, a typical multi-layer feature based method, extracts deep feature from each residual block. (e) Scale-specific network~\cite{herranz2016scene}, a multi-view feature based architecture, used scene-centric CNN extract deep features from coarse versions, while object-centric CNN is used to extract features from fine patches. Two types of deep features complement each other.}
\label{fig:typicalarchitecture}
\end{figure*}

\subsubsection{Global CNN feature based Method}
\label{sssec:globalCNN}

Global CNN feature based methods directly predict the probabilities of scene categories from the whole scene image. Frequently, global CNN features are extracted from input images via generic CNN models, pre-trained on large-scale datasets (\eg ImageNet~\cite{deng2009imagenet} and Places~\cite{zhou2014learning, zhou2017places}), or then fine-tuned on target datasets (\eg SUN397~\cite{xiao2010sun} and MIT67~\cite{quattoni2009recognizing}). Owing to the available large datasets and powerful computing resources (e.g., GPUs and parallel computing clusters)~\cite{sejnowski2020unreasonable}, deep networks have been developed into deeper and more complicated, and thus global representations from these networks are able to achieve a more advanced performance on many applications including scene classification.

Except for generic CNNs, some scene-specific CNNs are designed to extract global features. For instance, as shown in Fig.~\ref{fig:typicalarchitecture} (a), Zuo~\etal~\cite{zuo2016learning} proposed Hierarchical LSTM (HLSTM) to describe the contextual relation. They treated CONV maps as an undirected graph, which is transformed into four directed acyclic graphs, and LSTM modules are used to capture spatial contextual dependencies in an acyclic way. They also explored the potential spatial dependencies among different scale CONV maps, so HLSTM features not only involve the relations within the same feature maps but also contain the contextual dependencies among different scales. In addition, Liu \etal~\cite{liu2018dictionary} proposed DL-CNN model to extract sparse global features from entire scene image. Xie \etal~\cite{xie2016interactive} presented InterActive, a novel global CNN feature extraction algorithm which integrates high-level visual context with low-level neuron responses. InterActive increases the receptive field size of low-level neurons by allowing the supervision of the high-level neurons. Hayat~\etal~\cite{hayat2016spatial} designed a spatial unstructured layer to address the challenges of large-scale spatial layout deformations and scale variations. Along this way, Xie~\etal~\cite{xie2017towards} designed a Reversal Invariant Convolution (RI-Conv) layer so that they can obtain the identical representation for an image and its left-right reversed copy. Nevertheless, global CNN feature based methods have not fully exploited the underlying geometric and appearance variability of scene images.

The performance of global CNN features is greatly affected by the content of the input image. CNN models can extract generic global feature representations once trained on a sufficiently large and rich training dataset, as opposed to handcrafted feature extraction methods. It is noteworthy that global representations obtained by scene-centric CNN models yield more enriched spatial information than those obtained using object-centric CNN models, arguably since global representations from scene-centric CNNs contain spatial correlations between objects and global scene properties~\cite{zhou2014learning, zhou2017places, herranz2016scene}. In addition, Herranz~\etal~\cite{herranz2016scene} showed that the performance of a scene recognition system depends on the entities in the scene image, \ie when the global features are extracted from images with chaotic background, the model’s performance is degraded compared to the cases that the object is isolated from the background or the image has a plain background. This suggests that the background may introduce some noise into the feature that weakens the performance. Since contour symmetry provides a perceptual advantage when human observers recognize complex real-world scenes, Rezanejad~\etal~\cite{rezanejad2019scene} studied global CNN features from the full image and only contour information and showed that the performance of the full image as input is better, because CNN captures potential information from images. Nevertheless, they still concluded that contour is an auxiliary clue to improve recognition accuracy.

\subsubsection{Spatially Invariant Feature based Method}
\label{sssec:orderless}

To alleviate the problems caused by sequential operations in the standard CNN, a body of alternatives~\cite{gong2014multi, cimpoi2015deep, guo2016locally} proposed spatially invariant feature based methods to maintain spatial robustness. The ``spatially invariant'' means that the output features are robust against the geometrical variations of the input image~\cite{li2019mapnet}.

As shown in Fig.~\ref{fig:typicalarchitecture}~(b), spatially invariant features are usually extracted from multiple local patches. The visualization of such a feature extraction process is shown in Fig.~\ref{fig:pipeline} (marked in green arrows). The entire process can be decomposed into five basic steps: 1) Local patch extraction: a given input image is divided into smaller local patches, which are used as the input to a CNN model, 2) Local feature extraction: deep features are extracted from either the CONV or FC layers of the model, 3) Codebook generation: this step is to generate a codebook with multiple codewords based on the extracted deep features from different regions of the image. The codewords usually are learned in an unsupervised way (\eg using GMM), 4) Spatially invariant feature generation: given the generated codebook, deep features are encoded into a spatially invariant representation, and 5) Class prediction: the representation input is classified into a predefined scene category.

As opposed to patch-based local feature extraction (each local feature is extracted from an original patch by independently using the CNN extractor), local features can also be extracted from the semantic CONV maps of a standard CNN~\cite{yoo2015multi, gao2015deep, xiong2019rgb, xiong2020msn}. Specifically, since each cell (deep descriptor) of the feature map corresponds to one local image patch in the input image, each cell is regarded as a local feature. In this approach, the computation time is decreased, compared to independently processing of multiple spatial patches to obtain local features. For instance, Yoo~\etal~\cite{yoo2015multi} replaced the FC layers with CONV layers to obtain large amount of local spatial features. They also used multi-scale CNN activations to achieve geometric robustness. Gao~\etal~\cite{gao2015deep} used a spatial pyramid to directly divide the activations into multi-level pyramids, which contain more discriminative spatial information.

The feature encoding technique, which aggregates the local features, is crucial in relating local features with the final feature representation, and it directly influences the accuracy and efficiency of the scene classification algorithms~\cite{liu2019bow}. Improved Fisher Vector (IFV)~\cite{sanchez2013image}, Vector of Locally Aggregated Descriptors (VLAD)~\cite{jegou2010aggregating}, and Bag-of-Visual-Word (BoVW)~\cite{csurka2004visual} are among the popular and effective encoding techniques that are used in deep learning based methods. For instance, many methods, like FV-CNN~\cite{cimpoi2015deep}, MFA-FS~\cite{dixit2016object}, and MFAFVNet~\cite{li2017deep}, apply IFV encoding to obtain the image embedding as spatially invariant representations, while MOP-CNN~\cite{gong2014multi}, SDO~\cite{cheng2018scene}, \emph{etc} utilize VLAD to cluster local features. Noteworthily, the codebook selection and encoding procedures result in disjoint training of the model. To this end, some works proposed networks that are trained in an end-to-end manner, \eg NetVLAD~\cite{arandjelovic2016netvlad}, MFAFVNet~\cite{li2017deep}, and VSAD~\cite{wang2017weakly}. 

Spatially invariant feature based methods are efficient to achieve geometric robustness. Nevertheless, the sliding windows based paradigm requires multi-resolution scanning with fixed aspect ratios, which is not suitable for arbitrary objects with variable sizes or aspect ratios in the scene image. Moreover, using dense patches may introduce noise into the final representation, which decreases the classification accuracy. Therefore, extracting semantic features from salient regions of the scene image can circumvent these drawbacks.

\subsubsection{Semantic Feature based Method}
\label{sssec:semantic}

Processing all patches of the input image requires computational cost while yields redundant information. Object detection determines whether or not any instance of the salient regions is presented in an image~\cite{liu2020deep}. Inspired by this, object detector based approaches allow identifying salient regions of the scene, which provide distinctive information about the context of the image.

Different methods have been put forward to effective saliency detection, such as selective search~\cite{uijlings2013selective}, unsupervised discovery~\cite{singh2012unsupervised}, Multi-scale Combinatorial Grouping (MCG)~\cite{arbelaez2014multiscale}, and object detection networks (\eg Fast RCNN~\cite{girshick2015fast}, Faster RCNN~\cite{ren2016faster}, SSD~\cite{liu2016ssd}, Yolo~\cite{redmon2017yolo9000, redmon2016you}). For instance, since selective search combines the strengths of exhaustive search and segmentation, Liu~\etal~\cite{liu2014learning} used it to capture all possible semantic regions, and then used a pre-trained CNN to extract the feature maps of each region followed by a spatial pyramid to reduce map dimensions. Because the common objects or characteristics in different scenes lead to the commonality of different scenes, Cheng~\etal~\cite{cheng2018scene} used a region proposal network~\cite{ren2016faster} to extract the discriminative regions while discarde non-discriminative regions. These semantic feature based methods~\cite{liu2014learning, cheng2018scene} harvest many semantic regions, so encoding technology is adapted to aggregate key features, which pipeline is shown in Fig.~\ref{fig:pipeline} (red arrows).

On the other hand, some semantic feature based methods~\cite{wu2015harvesting, durand2016weldon} are based on weakly supervised learning, which directly predicts categories by several semantic features of the scene. For instance, Wu~\etal~\cite{wu2015harvesting} generated high-quality proposal regions by using MCG~\cite{arbelaez2014multiscale}, and then used SVM on each scene category to prune outliers and redundant regions. Semantic features from different scale patches supply complementary cues, since the coarser scales deal with larger objects, while the finer levels provide smaller objects or object parts. In practice, they found two semantic features sufficient to represent the whole scene, comparable to multiple semantic features. Training a deep model only using a single salient region may result in a suboptimal performance due to the possible existence of outliers in the training set. Hence, multiple regions can be selected to train the model together~\cite{durand2016weldon}. As shown in Fig.~\ref{fig:typicalarchitecture} (c), Durand~\etal~\cite{durand2016weldon} designed a Max layer to select the attention regions to enhance the discrimination. To provide a more robust strategy, they also designed a Min layer to capture the regions with the most negative evidence to further improve the model.

Although better performance can be obtained via using more semantic local features, semantic feature based methods deeply rely on the performance of object detection. Weak supervision settings (\ie without the patch labels of scene images) make it difficult to accurately identify the scene by the key information of an image~\cite{durand2016weldon}. Moreover, the error accumulation problem and extra computation cost also limit the development of semantic feature based methods~\cite{xiong2019rgb}.

\subsubsection{Multi-layer Feature based Method}
\label{sssec:multilayer}

Global feature based methods usually extract the high-layer CNN features, and feed them into a classifier to achieve classification task. Due to the compactness of such high-layer features, it is easy to miss some important slight clues~\cite{guo2016locally, lee2015deeply}. Features from different layers are complementary~\cite{wu2015harvesting, lin2017feature}. Low-layer features generally capture small objects, while high-layer features capture big objects \cite{wu2015harvesting}. Moreover, semantic information of low-layer features is relatively less, but the object location is accurate~\cite{lin2017feature}. To take full advantage of features from different layers, many methods~\cite{tang2017g, song2017combining, liu2019novel, xie2015hybrid} used the high resolution features from the early layers along with the high semantic information of the features from the latest layers of hierarchical models (\eg CNNs).

As shown in Fig.~\ref{fig:typicalarchitecture}~(d), typical multi-layer feature formation process includes: 1) Feature extraction: the outputs (feature maps) of certain layers are extracted as deep features, 2) Feature vectorization: vectorize the extracted feature maps, 3) Multi-layer feature combination: multiple features from different layers are combined into a single feature vector, and 4) Feature classification: classify the given scene image based on the obtained combined feature.

Although using all features from different layers seems to improve the final representation, it likely increases the chance of overfitting, and thus hurts performance~\cite{yang2015multi}. Therefore, many methods~\cite{xie2015hybrid, tang2017g, liu2019novel, song2017combining, yang2015multi} only extract features from certain layers. For instance, Xie~\etal~\cite{xie2015hybrid} constructed two dictionary-based representations, Convolution Fisher Vector (CFV), and Mid-level Local Discriminative Representation (MLR) to classify subsidiarily scene images. Tang \etal~\cite{tang2017g} divided GoogLeNet layers into three parts from bottom to top and extracted final feature maps of each part. Liu~\etal~\cite{liu2019novel} captured feature maps from each residual block from ResNet independently. Song~\etal~\cite{song2017combining} selected discriminative combinations from different layers and different network branches via minimizing a weighted sum of the probability of error and the average correlation coefficient. Yang~\etal~\cite{yang2015multi} used greedily select to explore the best layer combinations.

Feature fusion in multi-layer feature based methods is another important direction. Feature fusion techniques are mainly divided into two groups~\cite{snoek2005early, gunes2005affect, dong2014performance}: 1) Early fusion: extracting multi-layer features and merging them into a comprehensive feature for scene classification, and 2) Late fusion: directly learning each multi-layer feature via a supervised learner, which enforces the features to be directly sensitive to the category label, and then merging them into a final feature. Although the performance of late fusion is better, it is more complex and time-consuming, so early fusion is more popular~\cite{xie2015hybrid, tang2017g, liu2019novel, song2017combining}. In addition, addition and product rules are usually applied to combine multiple features~\cite{tang2017g}. Since the feature spaces in different layers are disparate, product rule is better than addition rule to fusing features, and empirical experiments on \cite{tang2017g} also show this statement. Moreover, Tang~\etal~\cite{tang2017g} proposed two strategies to fuse multi-layer features, \ie `fusion with score' and `fusion with features'. Fusion with score technique has obtained a better performance over fusion with feature thanks to the end-to-end training.

\subsubsection{Multiple-view Feature based Method}
\label{sssec:multiview}

Describing a complex scene just using a single and compact feature representation is a non-trivial task. Hence, there has been extensive effort to compute a comprehensive representation of a scene by integrating multiple features generated from complementary CNN models~\cite{sun2018fusing, li2020deep, zhang2015scene, wang2015object, wang2017knowledge, wang2017weakly}.

Features generated from \emph{networks trained on different datasets} usually are complementary. As shown in Fig.~\ref{fig:typicalarchitecture} (e), Herranz \etal \cite{herranz2016scene} found the best scale response of object-centric CNNs and scene-centric CNNs, and they combine the knowledge in a scale-adaptive way via either object-centric CNNs or scene-centric CNNs. This finding is widely used~\cite{wang2015object, xia2019ws}. For instance, the authors in~\cite{wang2015object} used an object-centric CNN to carry information about object depicted in the image, while a scene-centric CNN was used to capture global scene information. Along this way, Wang~\etal~\cite{wang2017weakly} designed PatchNet, a weakly supervised learning method, which uses image-level supervision information as the supervision signal for effective extraction of the patch-level features. To enhance the recognition performance, Scene-PatchNet and Object-PatchNet jointly used to extract features for each patch.

Employing \emph{complementary CNN architectures} is essential for obtaining discriminative multi-view feature representations. Wang~\etal~\cite{wang2017knowledge} proposed a multi-resolution CNN (MR-CNN) architecture to capture visual content in multiple scale images. In their work, normal BN-Inception~\cite{ioffe2015batch} is used to extract coarse resolution features, while deeper BN-Inception is employed to extract fine resolution features. Jin~\etal~\cite{jin2018hierarchy} used global features and spatially invariant features to account for both the coarse layout of the scene and the transient objects. Sun~\etal~\cite{sun2018fusing} separately extracted three representations, \ie object semantics representation, contextual information, and global appearance, from discriminative views, which are complementarity to each other. Specifically, the object semantic features of the scene image are extracted by a CNN followed by spatial fisher vectors, while the deep feature of a multi-direction LSTM-based model represents contextual information, and the FC feature represents global appearance. Li~\etal~\cite{li2020deep} used ResNet18~\cite{he2016deep} to generate discriminative attention maps, which is used as an explicit input of CNN together with the original image. Using  global features extracted by ResNet18 and attention map features extracted from the spatial feature transformer network, the attention map features are multiplied to the global features for adaptive feature refinement so that the network focuses on the most discriminative parts. Later, a multi-modal architecture is proposed in \cite{lopez2020semantic}, composed of a deep branch and a semantic branch. The deep Branch extracts global CNN features, while semantic branch aims to extract meaningful scene objects and their relations from super pixels. 
\subsection{Strategies for Improving Scene Representation}
\label{ssec:strategy}

To obtain more discriminative representations for scene classification, a range of strategies has been proposed. Four major categories (\ie encoding strategy, attention strategy, contextual strategy, and regularization strategy) will be discussed below. 

\subsubsection{Encoding strategy}
\label{sssec:coding}

Although the current driving force has been the incorporation of CNNs, encoding technology of the first generation methods have also been adapted in deep learning based methods. Fisher Vector (FV) coding \cite{perronnin2007fisher, sanchez2013image} is an encoding technique commonly used in scene classification. Fisher vector stores the mean and the covariance deviation vectors per component of the GMM and each element of the local features together. Thanks to the covariance deviation vectors, FV encoding leads to excellent results. Moreover, it is empirically proven that Fisher vectors are complementary to global CNN features~\cite{dixit2015scene, dixit2016object, wang2016modality, xie2015hybrid, guo2016locally, li2017deep}. Therefore, this survey takes FV-based approaches as the major cue and discusses the adapted combination of encoding technology and deep learning.

\begin{figure}[!htbp]
\centering
\includegraphics[width = .499\textwidth]{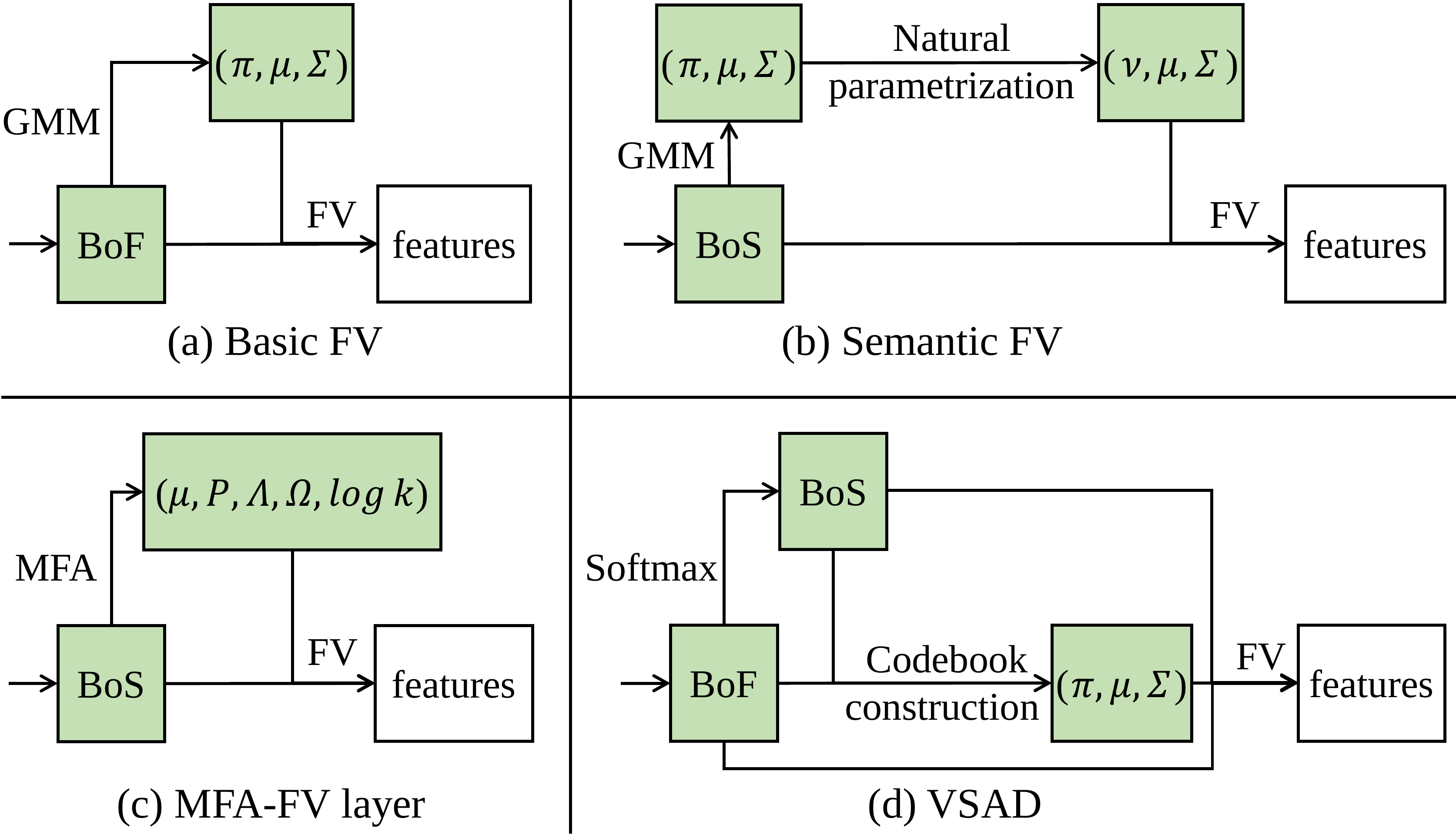}
\\
\caption{Structure comparisons of (a)~basic Fisher vector~\cite{cimpoi2015deep} and its variations. BoF denotes bag of features, while BoS represents bag of semantic probabilities. (b) In semantic FV~\cite{dixit2015scene}, natural parameterization is added to map multinomial distribution (\ie $\pi$) to its natural parameter space (\ie $\nu$). (c)~In MFAFVNet~\cite{li2017deep}, GMM is replaced with MFA to build codebook. (d) In VSAD~\cite{wang2017weakly}, codebook is constructed via exploiting semantics (\ie BoS) to aggregate local features (\ie BoF).}
\label{Fig.FV}
\end{figure}

Generally, CONV features and FC features are regarded as Bags of Features (BoF), they can be readily modeled by the Gaussian Mixture Model followed by Fisher Vector (GMM-FV)~\cite{dixit2015scene, li2017deep}. To avoid the computation of the FC layers, Cimpoi~\etal~\cite{cimpoi2015deep} utilized GMM-FV to aggregate BoF from different CONV layers, respectively. Comparing their experiment results, they asserted that the last CONV features can more effectively represent scenes. To rescue the fine-grained information of early/middle layers, Guo~\etal~\cite{guo2016locally} proposed Fisher Convolutional Vector (FCV) to encode the feature maps from multiple CONV layers. Wang~\etal~\cite{wang2016modality} extracted the feature maps from RGB, HHA, and surface normal images, and then directly encoded these maps by FV coding. In addition, through the performance comparisons of GMM-FV encoding on CONV features and FC features, respectively, Dixit \etal~\cite{dixit2015scene} asserted that the FC features is more effective for scene classification. However, since the CONV features and FC features do not derive from semantic probability space, it is likely to be both less discriminant and less abstract than the truly semantic features ~\cite{cimpoi2015deep, dixit2015scene}. The activations of Softmax layer are probability vectors, inhabiting the probability simplex, which are more abstract and semantic, but it is difficult to implement an effective invariant coding~(\eg GMM-FV)~\cite{kwitt2012scene, dixit2015scene}. To this end, Dixit~\etal~\cite{dixit2015scene} proposed an indirect FV implementation to aggregate these semantic probability features, \ie adding a step to convert semantic multinomials from probability space to the natural parameter space, as shown in Fig.~\ref{Fig.FV}~(b). Inspired by FV and VLAD, Wang~\etal~\cite{wang2017weakly} proposed Vector of Semantically Aggregated Descriptors (VSAD) to encode the probability features, as shown in Fig.~\ref{Fig.FV}~(d). Comparing the discriminant probability learned by the weakly-supervised method (PatchNet) with the generative probability from an unsupervised method (GMM), the results show that the discriminant probability is more expressive in aggregating local features. From the above discussion, representation encoding local features on probability space outperforms that on non-probability space.

Deep features usually are high dimensional ones. Therefore, more Gaussian kernels are needed to accurately model the feature space~\cite{liu2014encoding}. However, this would a lot of overhead to the computations and, hence, it is not efficient. Liu~\etal~\cite{liu2014encoding} empirically proved that the discriminative power of FV features increases slowly as the number of Gaussian kernels increases. Therefore, dimensionality reduction of the features is very important, as it directly affects the computational efficiency. A wide range of approaches~\cite{liu2014encoding, dixit2015scene, yoo2015multi, cinbis2015approximate, guo2016locally, dixit2016object, wang2017weakly} used poplular dimensionality reduction techniques, Principal Component Analysis (PCA), for pre-processing of the local features. Moreover, Liu~\etal~\cite{liu2014encoding} drew local features from Gaussian distribution with a nearly zero mean, which ensures the sparsity of the resulting FV. Wang~\etal~\cite{wang2016modality} enforced intercomponent sparsity of GMM-FV features via component regularization to discount unnecessary components.

Due to the non-linear property of deep features and a limited ability of the covariance of GMM, a large number of diagonal GMM components are required to model deep features so that the FV has very high dimensions~\cite{dixit2016object, li2017deep}. To address this issue, Dixit~\etal~\cite{dixit2016object} proposed MFA-FS, in which GMM is replaced by Mixtures of Factor Analysis (MFA)~\cite{ghahramani1996algorithm, verbeek2006learning}, \ie a set of local linear subspaces is used to capture non-linear features. MFA-FS performs well but does not support end-to-end training. However, end-to-end training is more efficient than any disjoint training process~\cite{li2017deep}. Therefore, Li~\etal~\cite{li2017deep} proposed MFAFVnet, an improved variant of MFA-FS~\cite{dixit2016object},  which is conveniently embedded into the state-of-the-art network architectures. Fig.~\ref{Fig.FV}~(c) shows the MFA-FV layer of MFAFVNet, compared with the other two structures.

In FV coding, local features are assumed to be independent and identically distributed (iid), which violates intrinsic image attributes that these patches are not iid. To this end, Cinbis~\etal~\cite{cinbis2015approximate} introduced a non-iid model via treating the model parameters as latent variables, rendering features related locally. Later, Wei~\etal~\cite{wei2017correlated} proposed a correlated topic vector, treated as an evolution oriented from Fisher kernel framework, to explore latent semantics, and consider semantic correlation.

\subsubsection{Attention strategy}
\label{sssec:attention}

As opposed to semantic feature based methods (focusing on key cues generally from original images), attention mechanism aims to capture distinguishing cues from the extracted feature space~\cite{li2019mapnet, xia2019ws, xiong2020msn, lopez2020semantic}. The attention maps are learned without any explicit training signal, rather task-related loss function alone provides the training signal for the attention weights. Generally, attention policy mainly includes channel attention and spatial attention.

\textbf{Channel attention policy.} Channel activation maps (ChAMs) generated from attention policy refers to the weighted activation maps, which highlights the class-specific discriminative regions. For instance, class activation map~\cite{zhou2016learning} is a simple ChAM, widely used in many works \cite{selvaraju2017grad, zhang2018top}. Since the same semantic cue has different roles for different types of scenes in some cases, Li~\etal~\cite{li2019mapnet} designed class-aware attentive pooling, including intra-modality attentive pooling and cross-aware attentive pooling, to learn the contributions of RGB and depth modalities, respectively. Here the attention strategies are also used to further fuse the learned discriminate semantic cues across RGB and depth modalities. Moreover, they also designed a class-agnostic attentive pooling to ignore some salient regions that may mislead classification. Inspired by the idea that specific objects are associated with a scene, Seong~\etal~\cite{seong2020fosnet} proposed correlative context gating to activate scene-specific object features.

\begin{figure}[!htbp]
\centering
\includegraphics[width = .48\textwidth]{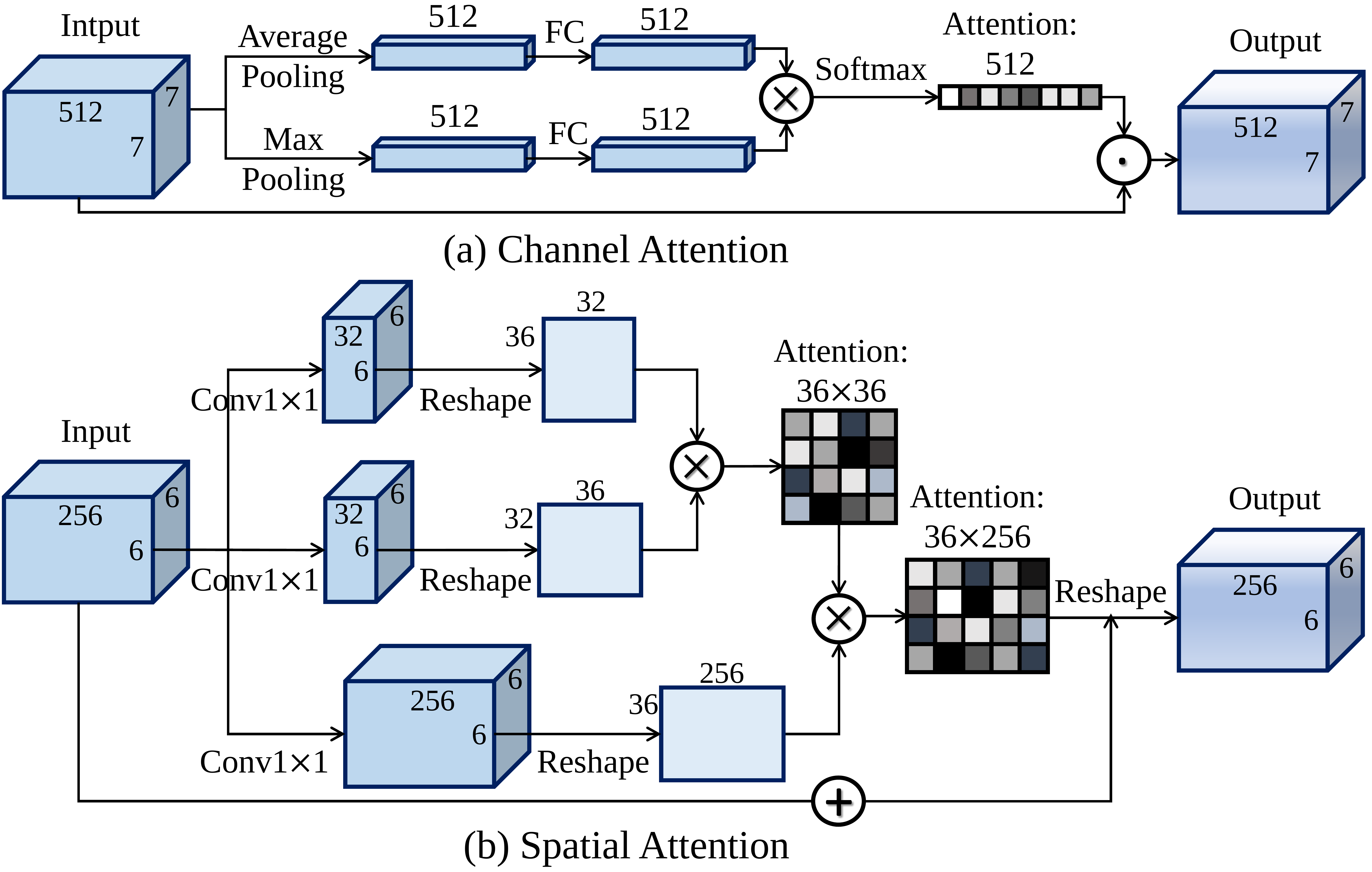}
\caption{Illustrations of two typical attentions. (a) In channel attention~\cite{lopez2020semantic}, the channel attention map is used to weight the input by a Hadamard product. (b) Spatial attention in~\cite{xiong2020msn} is used to enhance the local feature selection.}
\label{fig:attention}
\end{figure}

Channel attention maps can also be computed from different sources of information. With multiple salient regions on different scales as input, Xia~\etal~\cite{xia2019ws} designed a Weakly Supervised Attention Map (WS-AM) by proposing a gradient-weighted class activation mapping technique and privileging weakly supervised information. In another work~\cite{lopez2020semantic}, the input of semantic branch is a semantic segmentation score map, and semantic features are extracted from semantic branch via using three channel attention modules, shown in Fig.~\ref{fig:attention} (a). Moreover, semantic features are used to gate global CNN features via another attention module.

\textbf{Spatial attention policy.} Spatial attention policy infers attention maps along height and width of input feature maps, then the attention maps are combined to original maps for adaptive feature refinement. Joseph~\etal~\cite{joseph2019joint} proposed a layer-spatial attention model, including a hard attention to select a CNN layer and a soft attention to achieve spatial localization within the selected layer. Attention maps are obtained from a Conv-LSTM architecture, where the layer attention uses the previous hidden states, and spatial attention uses both the selected layer and the previous hidden states. To enhance local feature selection, Xiong~\etal~\cite{xiong2019rgb, xiong2020msn} designed a spatial attention module, shown in Fig.~\ref{fig:attention} (b), to generate attention masks. This attention masks of RGB and depth modalities are encouraged to be similar, and then learn the modal-consistent features.

\subsubsection{Contextual strategy}
\label{sssec:context}
Contextual information (the correlations among image regions, and local features), and objects/scenes may provide beneficial information in disambiguating visual words \cite{niu2012context}. However, convolution and pooling kernels are locally performed on image regions separately, and encoding technologies usually integrate multiple local features into an unstructured feature. As a result, contextual correlations among different regions have not been taken into account \cite{zuo2015convolutional}. To this end, contextual correlations have been further explored to focus on the global layout or local region coherence \cite{wang2008spatial}.

The contextual relations can broadly be grouped into two major categories: 1) spatial contextual relation: the correlations of neighboring regions, in which capturing spatial contextual relation usually encounters the problem of incomplete regions or noise caused by predefined grid patches, and 2) semantic contextual relation: the relations of salient regions. The network to extract semantic relations is often a two-stage framework (\ie detecting objects and classifying scenes). Therefore, accuracy is also influenced by object detection. Generally, there are three types of algorithms to capture contextual relations: 1) sequential model, like RNN~\cite{elman1990finding} and LSTM~\cite{hochreiter1997long}, and 2) graph-related model, such as Markov Random Field (MRF)~\cite{cross1983markov, li2009markov}, Correlated Topic Model (CTM)~\cite{rasiwasia2012holistic, wei2017correlated} and graph convolution \cite{bruna2013spectral, kipf2016semi, zhou2018graph}.

\begin{figure}[!htbp]
\centering
\subfigure[Sequential model]{{\includegraphics[width = .40\textwidth]{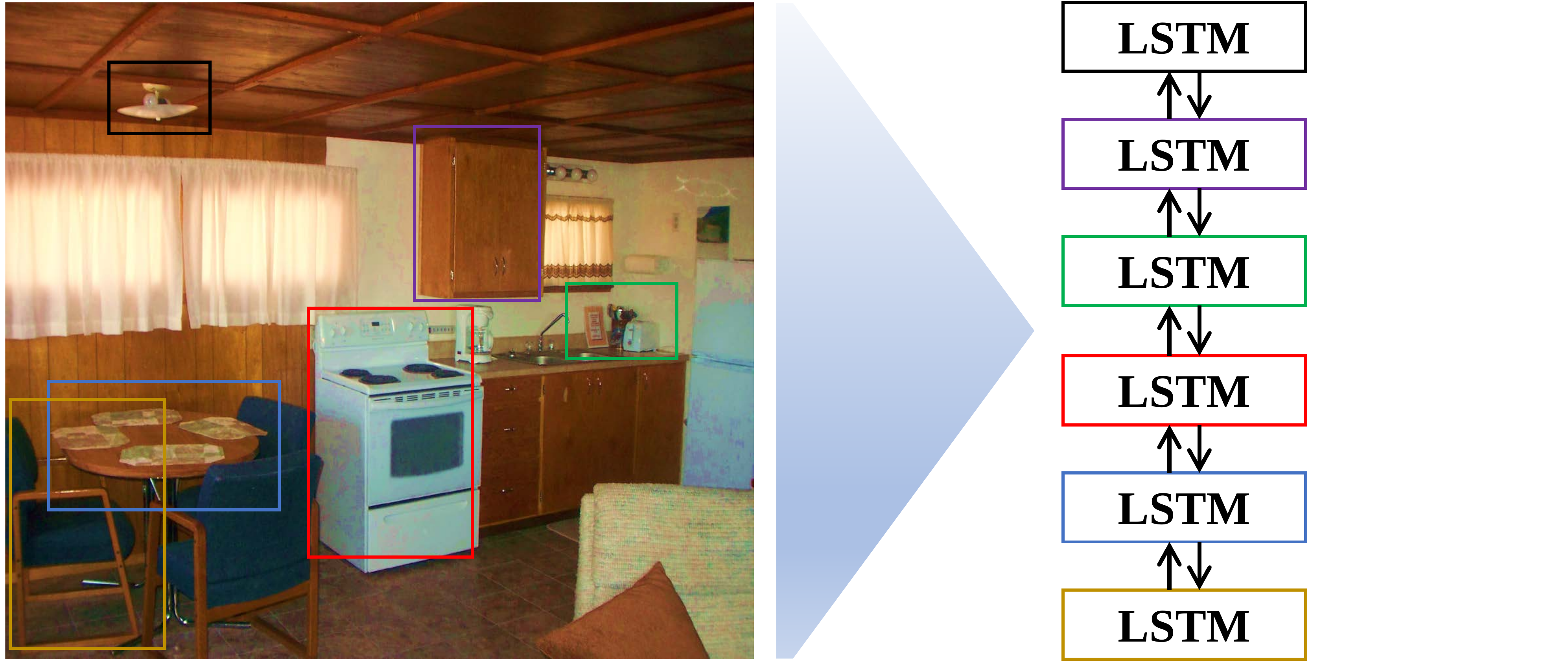}}}
\subfigure[Graph-related model]{{\includegraphics[width = .40\textwidth]{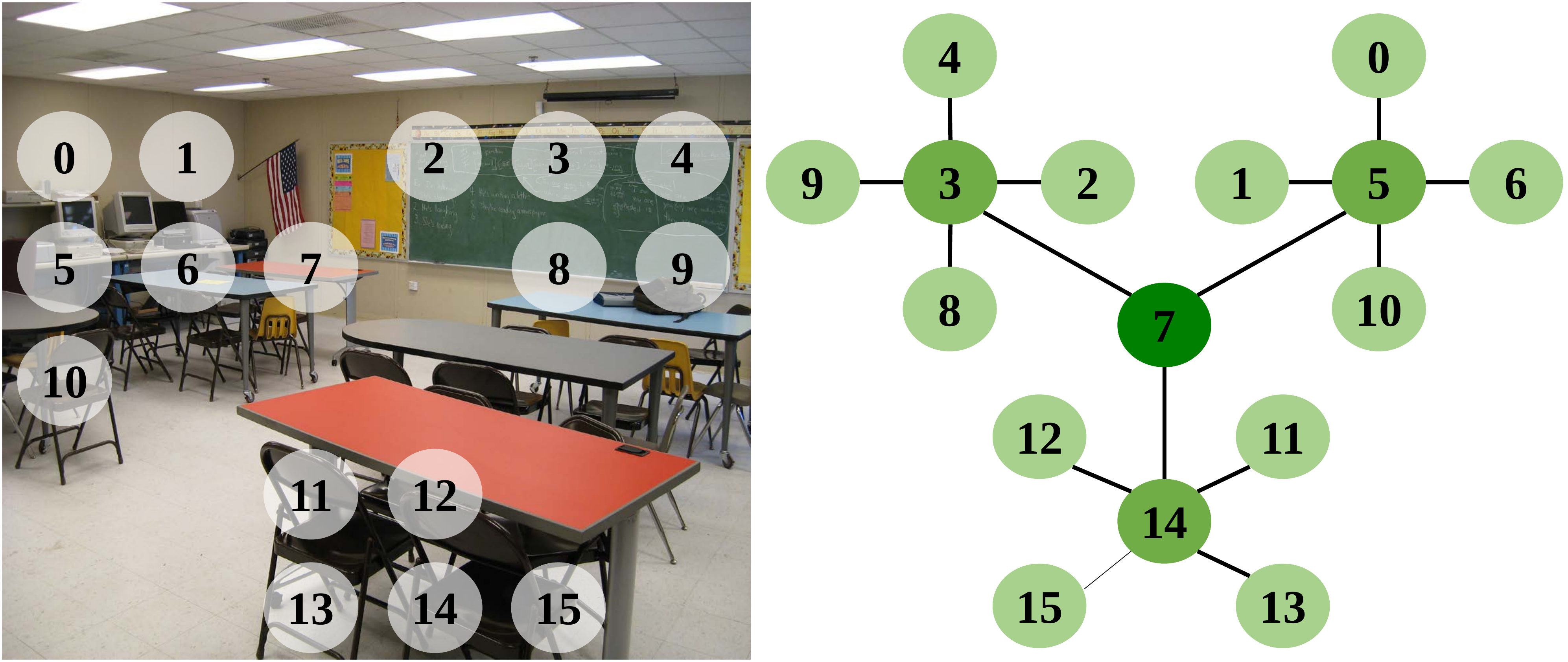}}}
\caption{Illustrations of a sequence model and a graph model to related contextual information. (a) Local feature is extracted from each salient region via a CNN, and a bidirectional LSTM is used to model synchronously many-to-many local feature relations~\cite{laranjeira2019modeling}. (b) Graph is constructed by assigning selected key features to graph nodes (including the center nodes, sub-center nodes and other nodes)~\cite{yuan2019acm}.}
\label{fig:contextual}
\end{figure}

\textbf{Sequential Model.} With the success of sequential models, such as RNN and LSTM, capturing the sequential information among local regions has shown promising performance for scene classification \cite{laranjeira2019modeling}. Spatial dependencies are captured from direct or indirect connections between each region and its surrounding neighbors. Zuo~\etal~\cite{zuo2016learning} stacked multi-directional LSTM layers on the top of CONV feature maps to encode spatial contextual information in scene images. Furthermore, a hierarchical strategy was adopted to capture cross-scale contextual information. Like the work~\cite{zuo2016learning}, Sun~\etal~\cite{sun2018fusing} bypassed two sets of multi-directional LSTM layers on the CONV feature maps. In their framework, the outputs of all LSTM layers are concatenated to form a contextual representation. In addition, the works~\cite{song2019image, laranjeira2019modeling} captured semantic contextual knowledge from variable salient regions. In~\cite{song2019image}, two types of representations, \ie COOR and SOOR, are proposed to describe object-to-object relations. Herein, COOR adapts the co-occurring frequency to represent the object-to-object relations, while SOOR is encoded with sequential model via regarding object sequences as sentences. Rooted in the work of Javed~\etal~\cite{javed2017object}, Laranjeira~\etal~\cite{laranjeira2019modeling} proposed a bidirectional LSTM to capture the contextual relations of regions of interest, as shown in Fig.~\ref{fig:contextual} (a). Their model supports variable length sequences, because the number of object parts of each image are different.

\textbf{Graph-related Model.} The sequential models often simplify the contextual relations, while graph-related models can explore more complicated structural layouts. Song~\etal~\cite{song2017multi} proposed a joint context model that uses MRFs to combine multiple scales, spatial relations, and multiple features among neighboring semantic multinomials, showing that this method can discover consistent co-occurrence patterns and filter out noisy ones. Based on CTM, Wei \etal~\cite{wei2017correlated} captured relations among latent themes as a semantic feature, \ie corrected topic vector (CTV). Later, with the development of Graph Neural Network (GNN) \cite{kipf2016semi, zhou2018graph}, graph convolution has become increasingly popular to model contextual information for scene classification. Yuan \etal~\cite{yuan2019acm} used spectral graph convolution to mine the relations among the selected local CNN features, as shown in Fig.~\ref{fig:contextual} (b). To use the complementary cues of multiple modalities, Yuan~\etal~also considered the inter-modality correlations of RGB and depth modalities through a cross-modal graph. Chen~\etal~\cite{chen2020scene} used graph convolution \cite{kipf2017semi} to model the more complex spatial structural layouts via pre-defining the features of discriminative regions as graph nodes. However, the spatial relation overlooks the semantic meanings of regions. To address this issue, Chen \etal~also defined a similarity subgraph as a complement to the spatial subgraph.

\subsubsection{Regularization strategy}
\label{sssec:regularization}

The training classifier not only requires a classification loss function, but it may also need multi-task learning with different regularization terms to reduce generalization error. The regularization strategies for scene classification mainly include sparse regularization, structured regularization, and supervised regularization.

\textbf{Sparse Regularization}. Sparse regularization is a technique to reduce the complexity of the model to prevent overfitting and even improve generalization ability. Many works~\cite{liu2014learning,zuo2014learning,guo2016locally,liu2018dictionary,jiang2019deep} include $\ell_0$, $\ell_1$, or $\ell_2$ norms to the base loss function for learning sparse features. For example, the sparse reconstruction term in~\cite{liu2014learning} encourages the learned representations to be significantly informative. The loss in~\cite{liu2018dictionary} combines the strength of the Mahalanobis and Euclidean distances to balance the accuracy and the generalization ability.

\textbf{Structured Regularization.} Minimizing the triplet loss function minimizes the distance between the anchor and positive features with the same class labels while maximizing the distance between the anchor and negative features with one different class labels. In addition, according to the maximum margin theory in learning \cite{boser1992training}, hinge distance focus on the hard training samples. Hence, many research efforts \cite{zuo2014learning, zhu2016discriminative, li2018df, xiong2019rgb, xiong2020msn} have utilized structured regularization of the triplet loss with hinge distance to learn robust feature representations. The structured regularization term is $\sum_{a,p,n}{max(d(x_a, x_p)-d(x_a, x_n)+\alpha, 0)}$, where $x_a, x_p, x_n$ are anchor, positive, negative features, and $\alpha$ is an adjustable parameter, while the function $d(x, y)$ denotes calculating a distance of $x$ and $y$. The structured regularization term promotes exemplar selection, while it also ignores noisy training examples that might overwhelm the useful discriminative patterns\cite{zuo2014learning, xiong2019rgb}.

\textbf{Supervised Regularization.} Supervised regularization uses the label information for tuning the intermediate feature maps. The supervised regularization is generally expressed in terms of $\sum_{i}{d(y_i, f(x_i))}$, where $x_i and y_i$ denote the middle-layer activated features and real label of the image $i$, respectively, and $f(x_i)$ is a predicted label. For example, Guo \etal~\cite{guo2016locally} utilized an auxiliary loss to directly propagate the label information to the CONV layers, and thus accurately captures the information of local objects and fine structures in the CONV layers. Similarly, these alternatives \cite{xiong2019rgb, yuan2019acm, xiong2020msn} used supervised regularization to learn modal-specific features.

\textbf{Others.} Extracting discriminative features by incorporating different regularization techniques has been always a mainstream topic in scene classification. For example, label consistent regularization \cite{liu2014learning} guarantees that inputs from different categories have discriminative responses. The shareable constraint in \cite{zuo2014learning} can learn a flexible number of filters to represent common patterns across different categories. Clustering loss in \cite{jin2018hierarchy} is utilized to further fine-tune confusing clusters to overcome the intra-class variation issues inherent. Since assigning soft labels to the samples cause a degree of ambiguity, which reaps high benefits when increasing the number of scene categories \cite{van2009visual}, Wang \etal~\cite{wang2017knowledge} improved generalization ability by exploiting soft labels contained in knowledge networks as a bias term of the loss function. Noteworthily, optimizing proper loss function can pick up effective patches for image classification. In fast RCNN \cite{girshick2015fast} and Faster RCNN \cite{ren2016faster}, regression loss is used to learn effective region proposals. Wu \etal~\cite{wu2015harvesting} adopted one-class SVMs \cite{scholkopf2001estimating} as discriminative models to get meta-objects. Inspired by MANTRA \cite{durand2015mantra}, the main intuition in \cite{durand2016weldon} is to equip each possible output with pairs of latent variables, \ie top positive and negative patches, via optimizing max$+$min prediction problem.

Nearly all multi-task learning approaches using regularization aim to find a trade-off among conflicting requirements, \eg accuracy, generalization robustness, and efficiency. Researchers apply completely different supervision information to a variety of auxiliary tasks in an effort to facilitate the convergence of the major scene classification task \cite{guo2016locally}.

\subsection{RGB-D Scene Classification}
\label{ssec:RGBD}

RGB modality provides the intensity of the colors and texture cues, while depth modality carries information regarding the distance of the scene surfaces from a viewpoint. The depth information is invariant to lighting and color variations, and contains geometrical and shape cues, which is useful for scene representation \cite{socher2012convolutional, wang2015mmss, cheng2016semi}. Moreover, HHA data~\cite{gupta2014learning}, an encoding result of depth image, depth information presents a certain color modality, which somewhat is similar to the RGB image. Hence, some CNNs trained on RGB images can transfer their knowledge and be used on HHA data.

The depth information of RGB-D image can further improve the performance of CNN models compared to RGB images \cite{song2018learning}. For the task of RGB-D scene classification, except for exploring suitable RGB features, described in Section~\ref{ssec:representation}, there exists another two main problems, \ie  1) how to extract depth-specific features and 2) how to properly fuse features of RGB and depth modalities.

\subsubsection{Depth-specific feature learning}

Depth information is usually scarce compared to RGB data. Therefore, it is non-trivial to train CNNs only on limited depth data to achieve depth-specific models~\cite{song2018learning}, \ie depth-CNN. Hence, different training strategies are employed to train CNNs using limited amount of depth images. 

\textbf{Fine-tuning RGB-CNNs for depth images.}
Due to the availability of RGB data, many models~\cite{cheng2017locality, zhu2016discriminative, wang2016modality} are first pre-trained on large-scale RGB datasets, such as ImageNet and Places, followed by fine-tuning on depth data. Fine-tuning only updates the last few FC layers, while the parameters of the previous layers are not adjusted. Therefore, the fine-tuned model’s layers do not fully leverage depth data~\cite{song2018learning}. However, abstract representations of early CONV layers play a crucial role in computing deep features using different modalities. Weakly-supervised learning and semi-supervised learning can enforce explicit adaptation in the previous layers.

\textbf{Weak-supervised learning with patches of depth images.}
Song~\etal~\cite{song2018learning, song2017combining} proposed to learn depth features from scratch using weakly supervised learning. Song \etal~\cite{song2018learning} pointed out that the diversity and complexity of patterns in the depth images are significantly lower than those in the RGB images. Therefore, they designed a Depth-CNN (DCNN) with fewer layers for depth features extraction. They also trained the DCNN by three strategies of freezing, fine-tuning, and training from scratch to adequately capture depth information. Nevertheless, weakly-supervised learning is sensitive to the noise in the training data. As a result, the extracted features may not have good discriminative quality for classification.

\textbf{Semi-supervised learning with unlabeled images.}
Due to the convenient collection of unlabeled RGB-D data, semi-supervised learning can also be employed in the training of CNNs with a limited number of labeled samples compared to very large size of unlabeled data~\cite{du2019translate, cheng2016semi}. Cheng~\etal~\cite{cheng2016semi} trained a CNN using a very limited number of labeled RGB-D images while a massive amount of unlabeled RGB-D images via a co-training algorithm to preserve diversity. Subsequently, Du~\etal~\cite{du2019translate} developed an encoder-decoder model to construct paired complementary-modal data of the input. In particular, the encoder is used as a modality-specific network to extract specific features for the subsequent classification task.

\subsubsection{Multiple modality fusion}

\begin{figure}[!ht]
\centering
\includegraphics[width = .48\textwidth]{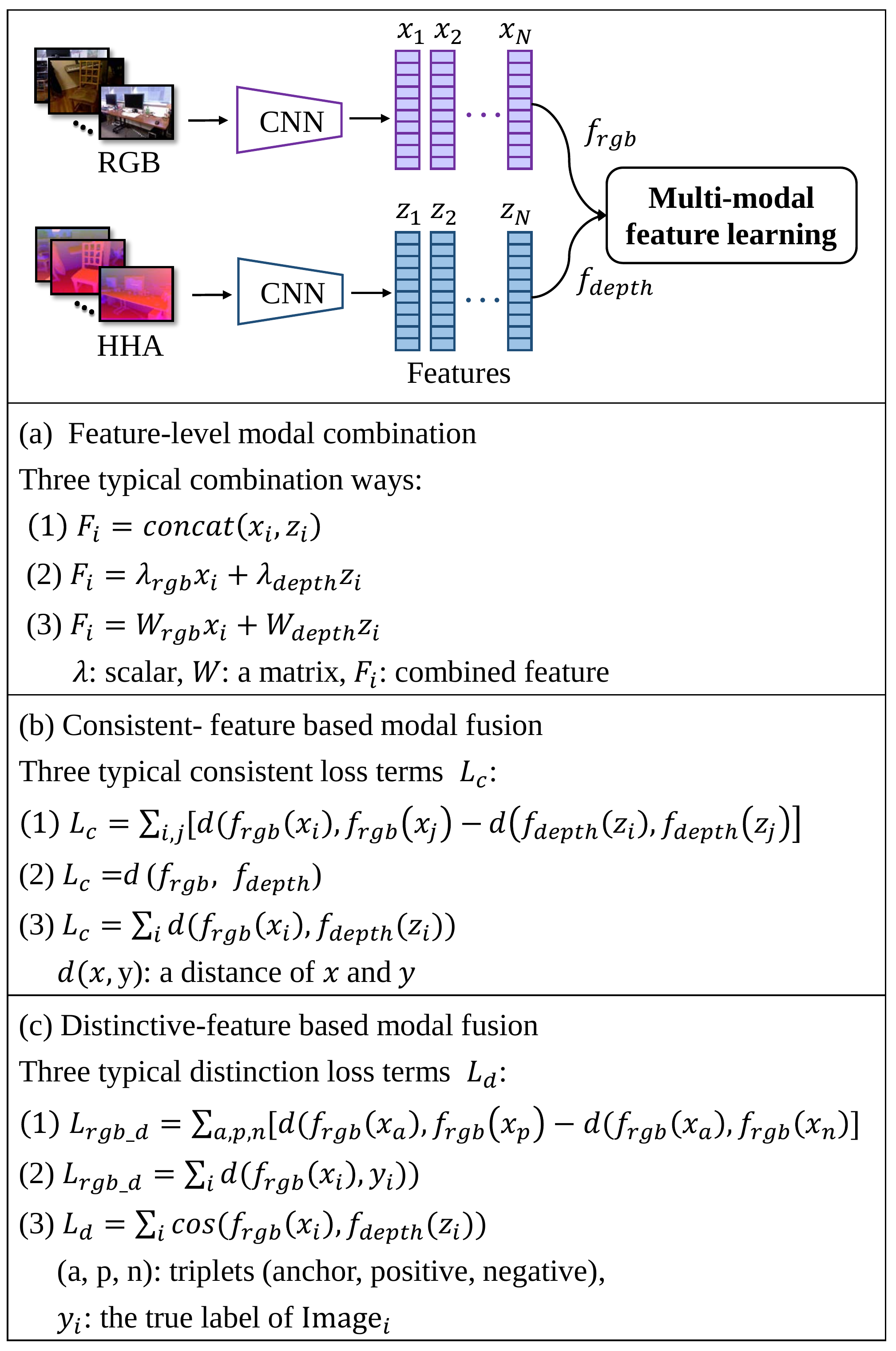}
\caption{Illustrations of multi-modal feature learning. (a) Three popular ways to achieve feature combination: directly concatenate features, combine weighted features and combine features with linear converting. (b) Three methods to achieve modal-consistent: minimize the pairwise distances between modalities \cite{zhu2016discriminative}; encourage the attention maps of modalities similar \cite{xiong2019rgb}; minimize the distances between modalities \cite{xiong2020msn, li2018df}. (c) Three strategies to achieve modal-distinctive: learn the model structure via triplet loss \cite{zhu2016discriminative, li2018df}; use label information to guide modal-specific learning \cite{xiong2019rgb, yuan2019acm}; minimize cosine similarity between modalities \cite{xiong2020msn}.}
\label{Fig.fusion}
\end{figure}

Various modality fusion methods~\cite{wang2018detecting, li2018df, xiong2020msn} have been put forward to combine the information of different modalities to further enhance the performance of the classification model. The fusion strategies are mainly divided into three categories, \ie feature-level modal combination, consistent feature based fusion, and distinctive feature based fusion. Fig.~\ref{Fig.fusion} shows illustrations of the late three categories. Despite the existence of different fusion categories, some works\cite{wang2016modality, li2018df, xiong2020msn} combine multiple fusion strategies to achieve better performance for scene classification.


\textbf{Feature-level modal combination.}
Song~\etal~\cite{song2017combining} proposed a multi-modal combination approach to select discriminative combinations of layers from different source models. They concatenated RGB and depth features for not losing the correlation between the RGB and depth data. Reducing the redundancy of features can significantly improve the performance when RGB and depth features have correlations; especially, in the case of extracting depth features merely via RGB-CNNs~\cite{song2018learning}. Because of such correlation, direct concatenation of features may result in redundancy of information. To avoid this issue, Du~\etal~\cite{du2019translate} performed global average pooling to reduce the feature dimensions after concatenating modality-specific features. Wang~\etal~\cite{wang2016modality} used the modality regularization based on exclusive group lasso to ensure feature sparsity and co-existence, while features within a modality are encouraged to compete. Li~\etal~\cite{li2019mapnet} used an attention module to discern discriminative semantic cues from intra- and cross-modalities. Moreover, Cheng~\etal~\cite{cheng2017locality} proposed a gated fusion layer to adjust the RGB and depth contributions on image pixels.

\textbf{Consistent-feature based modal fusion.}
Images may suffer from missing information or noise pollution so that multi-modal features are not consistent, hence it is essential to exploit the correlation between different modalities to exclude such issue. To drive feature consistency of different modalities, Zhu~\etal~\cite{zhu2016discriminative} introduced an inter-modality correlation term to minimize pairwise distances of two modalities from the same class, while maximize pairwise distances from different classes. Zheng~\etal~\cite{zheng2017indoor} used multi-task metric learning to learn linear transformations of RGB and depth features, making full use of inter-modal relations. Li~\etal~\cite{li2018df} learned a correlative embedding module between the RGB and depth features inspired by Canonical Correlation Analysis~\cite{thompson2005canonical, andrew2013deep}. Xiong~\etal~\cite{xiong2019rgb, xiong2020msn} proposed a learning approach to encourage two modal-specific networks to focus on features with similar spatial positions to learn more discriminative modal-consistent features.

\textbf{Distinctive-feature based modal fusion.}
In addition to constructing multimodal consistent features, features can also be processed separately to increase discriminative capability. For instance, Li~\etal~\cite{li2018df} and Zhu~\etal~\cite{zhu2016discriminative} adopted structured regularization in the triplet loss function, in which to encourage the model to learn the modal-specific features under the supervision of this regularization. Li~\etal~\cite{li2018df} designed a distinctive embedding module for each modality to learn distinctive features. Using labels for separate supervision of model-specific representation learning for each modality is also another technique of individual processing~\cite{xiong2019rgb, yuan2019acm, xiong2020msn}.  Moreover, by minimizing the feature correlation, Xiong \etal~\cite{xiong2020msn} learned the modal distinctive features as the RGB and depth modalities have different characteristics.

\section{Performance Comparison}
\label{sec:performance}

\subsection{Performance on RGB scene data}

\renewcommand\arraystretch{1.18}
\begin{table*}[htbp]
  \centering
  \setlength\tabcolsep{1pt}
  \caption{Performance (\%) summarization of some representative methods on popular benchmark datasets. All scores are quoted directly from the original papers. For each dataset, the highest three classification scores are highlighted. Some abbreviations. Column ``Scheme'': Whole Image (WI), Dense Patches (DP), Regional Proposals (RP); Column ``Init.''(Initialization): ImageNet (IN), Places205 (PL1), Places365 (PL2).  }
  \resizebox*{15cm}{!}{
    \begin{tabular}{|c|m{11em}<{\centering}|m{4.2em}<{\centering}|m{4.2em}<{\centering}|m{5em}<{\centering}|m{5.5em}<{\centering}|m{5.5em}<{\centering}|m{5em}<{\centering}|m{5.5em}<{\centering}|m{3.8em}<{\centering}|m{3.8em}<{\centering}|m{3.8em}<{\centering}|}
    \hline
    \multirow{2}{*}{Group} & \multirow{2}{*}{Method} & \multicolumn{3}{c|}{Input information} & \multicolumn{2}{c|}{Feature Information} & \multicolumn{2}{c|}{Architecture} & \multicolumn{3}{c|}{Results (RGB)} \\
\cline{3-12}          &  & Scale & Scheme & Data Aug. & Aggregation & Dimension & Init. & Backbone & Scene15 & MIT67 & SUN397 \\
    \hline
    \multicolumn{1}{|c|}{\multirow{4}[16]{*}{\shortstack{Global\\ CNN\\ features\\ based\\ methods}}} & ImageNet-CNN \cite{krizhevsky2012imagenet} & Single & WI    & ${\times}$ & pooling & 4,096  & IN    & AlexNet & 84.2 & 56.8 & 42.6 \\
\cline{2-12}          & PL1-CNN \cite{zhou2014learning} & Single & WI    & ${\times}$ & pooling & 4,096  & PL1   & VGGNet   & 91.6 & 79.8 & 62.0 \\
\cline{2-12}          & PL2-CNN \cite{zhou2017places} & Single & WI    & ${\times}$ & pooling & 4,096  & PL2   & VGGNet   & 92.0 & 76.5 & 63.2 \\
\cline{2-12}          & S2ICA \cite{hayat2016spatial} & Multi & DP    & ${\surd}$ & pooling & 8,192  & IN    & AlexNet & 93.1 & 71.2 & $-$ \\
\cline{2-12}          & GAP-CNN \cite{zhou2016learning} & Single & WI    & ${\surd}$ & GAP   & 4,096  & PL1   & GoogLeNet & 88.3 & 66.6 & 51.3 \\
\cline{2-12}          & InterActive \cite{xie2016interactive} & Single & WI    & ${\times}$ & pooling & 4,096  & IN    & VGG19 & $-$   & 78.7 & 63.0 \\
\cline{2-12}          & C-HLSTM \cite{zuo2016learning} & Multi & WI    & ${\surd}$ & LSTM  & 4,096  & PL1 & AlexNet & $-$   & 75.7 & 60.3 \\
\cline{2-12}          & DL-CNN \cite{liu2018dictionary} & Single & WI    & ${\times}$ & DL    & $-$ & PL1   & VGGNet   & 96.0 & 86.4 & 70.1 \\
    \hline
    \multicolumn{1}{|c|}{\multirow{15}[9]{*}{\shortstack{Spatially\\invariant\\ features\\ based\\ methods}}} & SCFVC \cite{liu2014encoding} & Single & DP    & ${\times}$ & FV    & 200,000  & IN    & AlexNet & $-$   & 68.2 & $-$ \\
\cline{2-12}          & MOP-CNN \cite{gong2014multi} & Multi & DP    & ${\times}$ & VLAD  & 12,288  & IN    & AlexNet & $-$   & 68.9 & 52.0 \\
\cline{2-12}          & DSP \cite{gao2015deep} & Multi & WI    & ${\times}$ & FV    & 12,288  & IN    & VGGNet   & 91.8 & 78.3 & 59.8 \\
\cline{2-12}          & MPP-CNN \cite{yoo2015multi} & Multi & RP    & ${\times}$ & FV    & 65,536  & IN    & AlexNet & $-$   & 80.8 & $-$ \\
\cline{2-12}          & SFV \cite{dixit2015scene} & Multi & DP, WI & ${\times}$ & FV    & 9,216  & IN    & AlexNet & $-$   & 72.8 & 54.4 \\
\cline{2-12}          & FV-CNN \cite{cimpoi2015deep} & Multi & RP, WI & ${\times}$ & FV    & 4,096  & IN    & VGGNet   & $-$   & 81.0 & $-$ \\
\cline{2-12}          & LatMoG \cite{cinbis2015approximate} & Multi & RP    & ${\times}$ & FV    & $-$ & IN    & AlexNet & $-$   & 69.1 & $-$ \\
\cline{2-12}          & DUCA \cite{khan2016discriminative} & Single & DP    & ${\surd}$ & CSMC  & 4,096  & IN    & AlexNet & 94.5 & 78.8 & $-$ \\
\cline{2-12}          & D3 \cite{wu2016representing} & Single & DP    & ${\times}$ & D3, FV & 1,048,576  & IN    & VGG16 & 92.8 & 77.1 & 61.5 \\
\cline{2-12}          & MFA-FS \cite{dixit2016object} & Multi & DP    & ${\times}$ & MFA-FV    & 5,000  & IN    & VGGNet   & $-$   & 81.4 & 63.3 \\
\cline{2-12}          & CTV \cite{wei2017correlated} & Multi & WI    & ${\times}$ & CTV   & $-$ & PL1   & AlexNet & $-$   & 73.9 & 58.4 \\
\cline{2-12}          & MFAFVNet \cite{li2017deep} & Multi & DP    & ${\times}$ & MFA-FV & 500,000  & IN    & VGG19 & $-$   & 82.7 & 64.6  \\
\cline{2-12}          & EMFS \cite{song2017multi} & Multi & DP    & ${\times}$ & SM    & 4,096  & IN, PL2 & VGGNet   & $-$   & 86.5 & 72.6 \\
\cline{2-12}          & VSAD \cite{wang2017weakly} & Multi & DP    & ${\surd}$ & VSAD  & 25,600  & IN, PL1 & VGG16 & $-$   & 86.1 & 72.0 \\
\cline{2-12}          & LLC \cite{jiang2019deep} & Single & DP    & ${\times}$ & SSE   & 3,072  & IN, PL2 & VGG16 & $-$   & 79.6 & 57.5 \\
    \hline
    \multicolumn{1}{|c|}{\multirow{8}[3]{*}{\shortstack{Semantic\\ features\\ based\\ methods}}} & URDL \cite{liu2014learning} & Multi & SS    & ${\surd}$ & pooling & 4,096+ & IN    & VGG16 & 91.2 & 71.9  & $-$ \\
\cline{2-12}          & MetaObject-CNN \cite{wu2015harvesting} & Multi & RP    & ${\surd}$ & LSAQ  & 4,096  & PL1   & AlexNet & $-$   & 78.9 & 58.1 \\
\cline{2-12}          & SOAL \cite{bappy2016online} & Multi & RP    & ${\times}$ & CRF   & 1,024  & PL1   & VGGNet   & $-$   & 82.5 & 75.5  \\
\cline{2-12}          & WELDON \cite{durand2016weldon} & Single & WI    & ${\times}$ & pooling & 4,096  & IN    & VGG16 & 94.3 & 78.0 & $-$ \\
\cline{2-12}          & Adi-Red \cite{zhao2018volcano} & Multi & DisNet & ${\times}$ & GAP   & 12,288  & IN, PL1-2 & ResNet & $-$   & $-$   & 73.6 \\
\cline{2-12}          & \textbf{SDO} \cite{cheng2018scene} & Multi & OMD   & ${\times}$ & VLAD  & 8,192  & PL1   & VGGNet   & \textbf{\emph{95.9}} & 86.8 & 73.4 \\
\cline{2-12}          & \textbf{M2M BiLSTM}~\cite{laranjeira2019modeling} & Single & SS    & ${\times}$ & LSTM  & $-$ & IN    & ResNet & \textbf{\emph{96.3}} & 88.3 & 71.8 \\
\cline{2-12}          & LGN \cite{chen2020scene} & Single & WI    & ${\times}$ & LGN   & 8,192  & PL2   & ResNet & $-$   & 88.1 & 74.1 \\
    \hline
    \multicolumn{1}{|c|}{\multirow{3}[8]{*}{\shortstack{Multilayer\\ features\\ based\\ methods}}} & Deep19-DAG \cite{yang2015multi} & Single & WI    & ${\times}$ & pooling & 6,144  & IN    & VGG19 & 92.9 & 77.5 & 56.2 \\
\cline{2-12}          & Hybrid CNNs \cite{xie2015hybrid} & Multi & SS    & ${\times}$ & FV    & 12,288+ & IN, PL1 & VGGNet & $-$   & 82.3 & 64.5 \\
\cline{2-12}          & G-MS2F \cite{tang2017g} & Single & WI    & ${\surd}$ & pooling & 3,072  & IN, PL1 & GoogLeNet & 93.2 & 80.0 & 65.1 \\
\cline{2-12}          & FTOTLM \cite{liu2019novel} & Single & WI    & ${\times}$ & GAP   & 3,968  & IN, PL2 & ResNet & 94.0   & 74.6 & $65.5$ \\
\cline{2-12}          & \textbf{FTOTLM Aug.} \cite{liu2019novel} & Single & WI    & ${\surd}$ & GAP   & 3,968  & IN, PL2 & ResNet & \textbf{\emph{97.4$^*$}}   & \textbf{\emph{94.1$^*$}} & \textbf{\emph{85.2$^*$}} \\
    \hline
    \multicolumn{1}{|c|}{\multirow{7}[3]{*}{\shortstack{Multiview\\ features\\ based\\ methods}}} & \cite{koskela2014convolutional} & Multi & WI    & ${\surd}$ & pooling & 8,192  & IN, aratio & AlexNet & 92.1 & 70.1 & 54.7 \\
\cline{2-12}          & Scale-specific CNNs \cite{herranz2016scene} & Multi & Crops & ${\times}$ & pooling & 4,096  & IN, PL1 & VGGNet   & 95.2 & 86.0 & 70.2 \\
\cline{2-12}          & LS-DHM \cite{guo2016locally} & Single & WI, DP & ${\times}$ & FV    & 40,960  & IN    & VGGNet   & $-$   & 83.8 & 67.6 \\
\cline{2-12}          &  \cite{nascimento2017robust} & Multi & WI, DP & ${\times}$ & SC    & 6,096  & IN, PL1 & VGG16 & 95.7 & 87.2 & 71.1 \\
\cline{2-12}          & MR CNN \cite{wang2017knowledge} & Multi & WI    & ${\surd}$ & pooling & $-$ & Places401 & Inception 2 & $-$   & 86.7 & 72.0 \\
\cline{2-12}          & \textbf{SOSF+CFA+GAF} \cite{sun2018fusing} & Single & WI, DP & ${\surd}$ & SFV   & 12,288  & IN    & VGG16 & $-$   & \textbf{\emph{89.5}} & \textbf{\emph{78.9}} \\
\cline{2-12}          & \textbf{FOSNet} \cite{seong2020fosnet} & Single & WI    & ${\surd}$ & GAP   & 4,096  & PL2 & ResNet & $-$   & \textbf{\emph{90.3}} & \textbf{\emph{77.3}} \\
\cline{2-12}          & \textbf{ChAM} \cite{lopez2020semantic} & Single & WI    & ${\surd}$ & pooling   & 512  & PL2 & ResNet & $-$   & \textbf{\emph{87.1}} & \textbf{\emph{74.0}} \\
    \hline
    \end{tabular}%
    }
  \label{tab:RGBresults}
\end{table*}

\begin{table*}[htbp]
  \setlength\tabcolsep{3pt}
  \centering
  \caption{Performance (\%) comparison of related methods with/without concatenating global CNN feature on benchmark scene datasets.}
  \resizebox*{15cm}{!}{
    \begin{tabular}{|m{4em}<{\centering}|m{8.2em}<{\centering}|m{5.2em}<{\centering}|m{5.2em}<{\centering}|m{6.5em}<{\centering}|m{7.5em}<{\centering}|m{5.2em}<{\centering}|m{7.5em}<{\centering}|m{5.2em}<{\centering}|m{5.2em}<{\centering}|}
    \hline
          &       & DSFL~\cite{zuo2014learning} & SFV~\cite{dixit2015scene} & MFA$-$FS~\cite{dixit2016object} & MFAFVNet~\cite{li2017deep} & VSAD~\cite{wang2017weakly} & SOSF$+$CFA~\cite{sun2018fusing} & SDO~\cite{cheng2018scene} & LGN~\cite{chen2020scene} \\
    \hline
    \multirow{2}[2]{*}{MIT67} & Baseline & 52.2 & 72.8  & 81.4  & 82.6  & 84.9  & 84.1  & 68.1  & 85.2 \\
          & $+$Global feature &76.2 ($\uparrow24$) & 79 ($\uparrow6.2$) & 87.2 ($\uparrow5.8$) & 87.9 ($\uparrow5.3$) & 85.3 ($\uparrow0.4$) & 89.5 ($\uparrow5.4$) & 84 ($\uparrow15.9$) &85.4 ($\uparrow0.2$) \\
    \hline
    \multirow{2}[2]{*}{SUN397} & Baseline & $-$   & 54.4  & 63.3  & 64.6  & 71.7  & 66.5  & 54.8  & $-$ \\
          & $+$Global feature & $-$   & 61.7 ($\uparrow7.3$) & 71.1 ($\uparrow7.8$) & 72 ($\uparrow7.4$) & 72.5 ($\uparrow0.8$) & 78.9 ($\uparrow12.4$) & 67 ($\uparrow12.2$) & $-$ \\
    \hline
    \end{tabular}%
    }
  \label{tab:withdeepfeature}%
\end{table*}%

In contrast, CNN-based methods have quickly demonstrated their strengths in scene classification. Table~\ref{tab:RGBresults} compares the performance of deep models for scene classification on RGB datasets. To gain insight into the performance of the presented methods, we also provided input information, feature information, and architecture of each method. The results show that a simple deep model (\ie AlexNet), which is trained on ImageNet, achieves 84.23\%, 56.79\%, and 42.61\% accuracy on Scene15, MIT67, and SUN397 datasets, respectively. This accuracy is comparable with the best non-deep learning methods. Starting from the generic deep models~\cite{krizhevsky2012imagenet, zhou2014learning, zhou2017places}, CNN-based methods improve steadily when more effective strategies have been introduced. As a result, nearly all the approaches yielded an accuracy of 90\% on the Scene15 dataset. Moreover, FTOTLM~\cite{liu2019novel} combined with a novel data augmentation outperforms other state-of-the-art models and achieves the best accuracy on three benchmark datasets so far.

Extracting global CNN features, which are computed using a pre-trained model, is faster than other deep feature representation techniques, but their quality is not good when there are large differences between the source and target datasets. Comparing these performances~\cite{krizhevsky2012imagenet, zhou2014learning, zhou2017places} demonstrates that the expressive power of global CNN features is improved as richer scene datasets appear. In GAP-CNN~\cite{zhou2016learning} and DL-CNN~\cite{liu2018dictionary}, new layers with a small number of parameters substitute for FC layers, but they can still achieve considerable results comparing with benchmark CNNs~\cite{krizhevsky2012imagenet, zhou2014learning}. 

Spatially invariant feature based methods are usually time-consuming, especially the computational time of sampling local patches, extracting individually local features, and building codebook. However, these methods are robust against geometrical variance, and thus improve the accuracy of benchmark CNNs, like SFV~\cite{dixit2015scene} vs. ImageNet-CNN~\cite{krizhevsky2012imagenet}, and MFA-FS~\cite{dixit2016object} vs. PL1-CNN~\cite{zhou2014learning}. Encoding technologies generally include more complicated training procedure, so some architectures (\eg MFAFVNet~\cite{li2017deep} and VSAD~\cite{wang2017weakly}) are designed in a unified pipeline to reduce the operation complexity.

Semantic feature based methods~\cite{zhao2018volcano, cheng2018scene, laranjeira2019modeling, chen2020scene} demonstrate very competitive performance, due to the discriminative information laying on the salient regions, compared to global CNN feature based and spatially invariant feature based methods. Salient regions generally are generated by region selection algorithms, which may cause a two-stage training procedure and require more time and computations~\cite{xie2020scene}. In addition, even though the contextual analysis demands more computational power, methods~\cite{laranjeira2019modeling, chen2020scene}, exploring the contextual relations among salient regions, can significantly improve the classification accuracy.

Multi-layer feature based methods employ the complementary features from different layers to improve performance. It is a simple way to use more feature cues, while it also does not require to add any other layers. However, these methods are structurally complicated and have high-dimensional features, which make training models difficult and prone to overfitting~\cite{yang2015multi}. Nevertheless, owing to a novel data augmentation, FTOTLM~\cite{liu2019novel} yields a gain of 19.5\% and 19.7\% on MIT67 and SUN397, respectively, and has achieved the best results so far.

Multi-view feature based methods take full advantage of features extracted from various CNNs to achieve high classification accuracy. For instance, Table~\ref{tab:withdeepfeature} shows that combining global features with other baselines significantly improves their original classification accuracy, \eg a baseline model ``SFV''\cite{dixit2015scene} achieves 72.8\% on MIT67, while ``SFV$+$global feature'' yields 79\%. Moreover, there are two aspects to emphasize: 1) Herranz~\etal~\cite{herranz2016scene} empirically proved that combining too much invalid features is marginally helpful and significantly increases calculation and introduces noise into the final feature. 2) It is essential to improve the expression ability of each view feature, and thus enhance the entire ability of multi-view features. 

In summary, the scene classification performance can be boosted by adopting more sophisticated deep models~\cite{simonyan2014very, he2016deep} and large-scale datasets~\cite{zhou2014learning, zhou2017places}. Meanwhile, deep learning based methods can obtain relatively satisfied accuracy on public datasets via combining multiple features~\cite{sun2018fusing}, focusing on semantic regions~\cite{cheng2018scene}, augmenting data~\cite{liu2019novel}, and training in a unified pipeline~\cite{seong2020fosnet}. In addition, many methods also improve their accuracy via adopting different strategies, \ie improved encoding~\cite{wang2017weakly, li2017deep}, contextual modeling~\cite{laranjeira2019modeling, chen2020scene}, attention policy~\cite{zhou2016learning, lopez2020semantic}, and multi-task learning~\cite{liu2018dictionary, guo2016locally}.

\renewcommand\arraystretch{1.18}
\begin{table*}[htbp]
  \centering
  \setlength\tabcolsep{1pt}
  \caption{Performance (\%) comparison of representative methods on benchmark RGB-D scene datasets. For each dataset, the top three scores are highlighted.}
  \resizebox*{15cm}{!}{
    \begin{tabular}{|m{5em}<{\centering}|m{8.5em}<{\centering}|m{6em}<{\centering}|m{6em}<{\centering}|m{5em}<{\centering}|m{16em}<{\centering}|m{4.5em}<{\centering}|m{5.5em}<{\centering}|m{5.5em}<{\centering}|}
    \hline
    \multirow{2}{*}{Group} & \multirow{2}{*}{Method} & \multicolumn{2}{c|}{Architecture} & \multicolumn{3}{c|}{Detailed Information} & \multicolumn{2}{c|}{Results} \\
\cline{3-9}          &  & RGB-CNN & Depth-CNN & Dimension & Modal Fusion & Classifier & NYUD2 & SUN RGBD \\
    \hline
    \multirow{1}{*}{Dataset} & SUN RGBD\cite{song2015sun} & \multicolumn{2}{m{12em}<{\centering}|}{PL1-AlexNet} & 8,192 & Feature-level concatenation & SVM   & $-$ &$39$ \\
    \hline
    \multirow{7}{*}{\shortstack{Feature\\learning}} & SS-CNN \cite{liao2016understand} & \multicolumn{2}{m{12em}<{\centering}|}{PL1-ASPP} & 4,096 & Image-level stacking & Softmax & \multicolumn{1}{m{4.19em}<{\centering}|}{$-$} & $41.3$ \\
\cline{2-9}          & MMML \cite{zheng2017indoor} & \multicolumn{2}{m{12em}<{\centering}|}{IN-DeCAF} & 256 & Feature-level concatenation & SVM   & $61.4$ & \multicolumn{1}{m{4.19em}<{\centering}|}{$-$} \\
\cline{2-9}          & MSMM \cite{song2017combining} & \multicolumn{2}{m{12em}<{\centering}|}{PL1-AlexNet} & 12,288+ & Feature-level concatenation & wSVM  & $66.7$  & $52.3$ \\
\cline{2-9}          & \textbf{MAPNet} \cite{li2019mapnet} & \multicolumn{2}{m{12em}<{\centering}|}{PL1-AlexNet} & 5,120 & Local and semantic feature concatenation & Softmax & $67.7$  & \textbf{\emph{56.2}} \\
\cline{2-9}          & SOOR\cite{song2019image} & PL1-AlexNet & PL1-DCNN & 512 & Local and global feature concatenation & SVM   & $67.4$  & $55.5$ \\
\cline{2-9}          & ACM \cite{yuan2019acm} & \multicolumn{2}{m{12em}<{\centering}|}{PL2-AlexNet} & 8192+ & Feature-level concatenation & Softmax & $67.2$  & $55.1$ \\
\cline{2-9}          & LM-CNN \cite{cai2019rgb} & \multicolumn{2}{m{12em}<{\centering}|}{IN-AlexNet} & 8,192 & Local feature concatenation & Softmax & \multicolumn{1}{m{4.19em}<{\centering}|}{$-$} & $48.7$ \\
    \hline
    \multirow{2}{*}{\shortstack{Depth\\ feature \\learning}} & DCNN \cite{song2018learning} & PL1-RCNN & PL1-DCNN & 4,608 & Feature-level concatenation & wSVM  & $67.5$  & $53.8$ \\
\cline{2-9}          & \textbf{TRecgNet} \cite{du2019translate} & \multicolumn{2}{m{12em}<{\centering}|}{SUN RGBD \& PL2 ResNet18} & 1024 & Feature-level concatenation & Softmax & \textbf{\emph{69.2}} & \textbf{\emph{56.7}} \\
    \hline
    \multirow{6}{*}{\shortstack{Multiple\\modal\\ fusion}} & DMMF \cite{zhu2016discriminative} & \multicolumn{2}{m{12em}<{\centering}|}{PL1-AlexNet} & 4096 & Inter- and intra- modal correlation and distinction  & L-SVM & \multicolumn{1}{m{4.19em}<{\centering}|}{$-$} & $41.5$ \\
\cline{2-9}          & MaCAFF \cite{wang2016modality} & \multicolumn{2}{m{12em}<{\centering}|}{PL1-AlexNet} & $-$  & Local and global feature concatenation & L-SVM & $63.9$  & $48.1$ \\
\cline{2-9}          & DF2Net \cite{li2018df} & \multicolumn{2}{m{12em}<{\centering}|}{PL1-AlexNet} & 512 & Modal correlation  and distinction & Softmax & $65.4$  & $54.6$ \\
\cline{2-9}          & \textbf{KFS} \cite{xiong2019rgb} & \multicolumn{2}{m{12em}<{\centering}|}{PL1-AlexNet} & 9,216+ & Modal correlation  and distinction & Softmax & $67.8$  & 55.9 \\
\cline{2-9}          & \textbf{CBSC} \cite{ayub2019centroid} & \multicolumn{2}{m{12em}<{\centering}|}{PL2-VGG16} & $-$  & Feature-level concatenation & Softmax & \textbf{\emph{69.7$^*$}}  &  \textbf{\emph{57.8$^*$}} \\
\cline{2-9}          & \textbf{MSN} \cite{xiong2020msn} & \multicolumn{2}{m{12em}<{\centering}|}{PL1-AlexNet} & 9,216+ & Modal correlation  and distinction & Softmax & \textbf{\emph{68.1}}  & \textbf{\emph{56.2}}  \\
    \hline
    \end{tabular}%
    }
  \label{tab:RGBDresults}%
\end{table*}%

\begin{table*}[htbp]
  \setlength\tabcolsep{0pt}
  \centering
  \caption{Ablation study on benchmark datasets to validate the performance (\%) of depth information.}
  \resizebox*{15cm}{!}{
    \begin{tabular}{|m{5.8em}<{\centering}|m{4.8em}<{\centering}|m{7em}<{\centering}|m{6em}<{\centering}|m{6em}<{\centering}|m{6.3em}<{\centering}|m{7.1em}<{\centering}|m{5.8em}<{\centering}|m{6em}<{\centering}|m{5.8em}<{\centering}|m{5.68em}<{\centering}|}
    \hline
          &       & MaCAFF\cite{wang2016modality} & MSMM\cite{song2017combining} & DCNN\cite{song2018learning} & DF2Net\cite{li2018df} & TRecgNet\cite{du2019translate} & ACM\cite{yuan2019acm} & CBCL\cite{ayub2019centroid} & KFS\cite{xiong2019rgb} & MSN\cite{xiong2020msn} \\
    \hline
    \multirow{2}[2]{*}{NYUD2} & Baseline & 53.5  & $-$ & 53.4  & 61.1  & 53.7  & 55.4  & 66.4  & 53.5  & 53.5 \\
          & $+$Depth & 63.9~($\uparrow10.4$) & $-$ & 67.5~($\uparrow14.1$) & 65.4~($\uparrow4.3$) & 67.5~($\uparrow13.8$) & 67.4~($\uparrow12$) & 69.7~($\uparrow3.3$) & 67.8~($\uparrow14.3$) & 68.1~($\uparrow14.6$) \\
    \hline
    \multirow{2}[2]{*}{SUN RGBD} & Baseline & 40.4  & 41.5  & 44.3  & 46.3  & 42.6  & 45.7  & 48.8  & 36.1  & $-$ \\
          & $+$Depth & 48.1~($\uparrow7.7$) & 52.3~($\uparrow10.8$) & 53.8~($\uparrow9.5$) & 54.6~($\uparrow8.3$) & 53.3~($\uparrow10.7$) & 55.1~($\uparrow9.4$) & 57.8~($\uparrow9.0$) & 41.3~($\uparrow5.2$) & $-$ \\
    \hline
    \end{tabular}%
    }
  \label{tab:withdepth}%
\end{table*}%

\subsection{Performance on RGB-D scene datasets}

The accuracy of different methods on RGB-D datasets is summarized in Table~\ref{tab:RGBDresults}. By adding depth information with different fusion strategies, accuracy results (see Table~\ref{tab:withdepth}) are improved over 10.8\% and 8.8\% on average on NYUD2 and SUN RGBD datasets, respectively. Since depth data provide extra information to train classification model, this observation is within expectation. Noteworthily, it is more difficult to improve the effect on a large dataset (SUN RGBD) than a small dataset (NYUD2). Moreover, the best results on NYUD2 and SUN RGBD datasets achieved by CBSC~\cite{ayub2019centroid} are as high as 69.7\% and 57.84\%, respectively. 

RGB-D scene data for training are relatively limited, while the dimension of scene features is high. Hence, Support Vector Machines (SVMs) are commonly used in RGB-D scene classification~\cite{song2015sun, zheng2017indoor, song2017combining, song2018learning} in the early stages. Thanks to data augmentation and back-propagation, Softmax classifier becomes progressively popular, and it is an important reason to yield a comparable performance~\cite{li2019mapnet, du2019translate, ayub2019centroid, xiong2020msn}.

Many methods, such as~\cite{wang2016modality, zhu2016discriminative}, fine-tune RGB-CNNs to extract deep features of depth modality, where the training process is simple, and the computational cost is low. To adapt to depth data, Song~\etal~\cite{song2017depth, song2018learning} used weakly-supervised learning to train depth-specific models from scratch, which achieves a gain of 3.5\% accuracy, compared to the fine-tuned RGB-CNN. TRecgNet~\cite{du2019translate}, which is based on semi-supervised learning, requires complicated training process and high computational cost, but it obtains comparable results (69.2\% on NYUD2 and 56.7\% on SUN RGBD).

Feature-level fusion based methods are commonly used due to their high cost-effectiveness, \eg \cite{ayub2019centroid, li2019mapnet, yuan2019acm}. Along this way, consistent-feature based, and distinctive-feature based modal fusion use complex fusion layer with high cost, like inference speed and training complexity, but they generally yield more effective features~\cite{li2018df, xiong2019rgb, xiong2020msn}. 

We can observe that the field of RGB-D scene classification has constantly been improving. Weakly-supervised and semi-supervised learning are useful to learn depth-specific features~\cite{song2017combining, song2018learning, du2019translate}. Moreover, multi-modal feature fusion is a major issue to improve performance on public datasets~\cite{li2018df, ayub2019centroid, xiong2020msn}. In addition, effective strategies (like contextual strategy~\cite{song2019image, yuan2019acm} and attention mechanism~\cite{li2019mapnet, xiong2020msn}) are also popular for RGB-D scene classification. Nevertheless, the accuracy achieved by current methods is far from expectation and there remains significant rooms for future improvement.

\section{Conclusion and Outlook}
\label{sec:conclusion}

As a contemporary survey for scene classification using deep learning, this paper has highlighted the recent achievements, provided some structural taxonomy for various methods according to their roles in scene representation for scene classification, analyzed their advantages and limitations, summarized existing popular scene datasets, and discussed performance for the most representative approaches. Despite great progress, there are still many unsolved problems. Thus in this section, we will point out these problems and introduce some promising trends for future research. We hope that this survey not only provides a better understanding of scene classification for researchers but also stimulates future research activities.

\textbf{Develop more advanced network frameworks.} With the development of deep CNN architectures, from generic CNNs~\cite{krizhevsky2012imagenet, simonyan2014very, szegedy2015going, he2016deep} to scene-specific CNNs~\cite{zhou2016learning, liu2018dictionary}, the accuracy of scene classification is getting increasingly comparable. Nevertheless, there still exists lots of works to be explored on the theoretical research of deep learning~\cite{najafabadi2015deep}. It is a further direction to solidify the theoretical basis so as to get more advanced network frameworks. In particular, it is essential to design specific frameworks for scene classification, such as using automated Neural Architecture Search (NAS)~\cite{zoph2017neural, elsken2019neural}, or according to scene attributes.

\textbf{Release rich scene datasets.} Deep learning based models require enormous amounts of data to initialize their parameters so that they can learn the scene knowledge well~\cite{zhou2014learning, zhou2017places}. However, compared to scenes of real world, the publicly available datasets are not large or rich enough, so it is essential to release datasets that encompass richness and high-diversity of environmental scenes~\cite{xiao2016sun}, especially large-scale RGB-D scene datasets. As opposed to object/texture datasets, scene appearance may be changed dramatically as time goes by, and there emerges new functional scenes as humans develop activity places. Therefore, it requires updating the original scene data and releasing new scene datasets regularly.

\textbf{Reduce the dependence of labeled scene images.}
The success of deep learning heavily relies on gargantuan amounts of labeled images. However, the labeled training images are always very limited, so supervised learning is not scalable in the absence of fully labeled training data and its generalization ability to classify scenes frequently deteriorates. Therefore, it is desirable to reduce dependence on large amounts of labeled data. To alleviate this difficulty, if with large numbers of unlabel data, it is necessary to further study semi-supervised learning~\cite{chapelle2009semi}, unsupervised learning~\cite{barlow1989unsupervised}, or self-supervised learning~\cite{kolesnikov2019revisiting}. Even more constrained, without any unlabel training data, the ability to learn from only few labeled images, small-sample learning~\cite{wang2016learning}, is also appealing.

\textbf{Few shot scene classification.}
The success of generic CNNs for scene classification relies heavily on gargantuan amounts of labeled training data~\cite{liu2020deep}. Due to the large intra-variation among scenes, scene datasets cannot cover various classes so that the performance of CNNs frequently deteriorates and fails to generalize well. In contrast, humans can learn a visual concept quickly from very few given examples and often generalize well~\cite{fei2006one, lake2015human}. Inspired by this, domain adaptation approaches utilize the knowledge of labeled data in task-relevant domains to execute new tasks in target domain~\cite{long2015learning, wang2018deep}. Furthermore, domain generalization methods aim at learning generic representation from multiple task-irrelevant domains to generalize unseen scenarios \cite{peng2018zero, li2018domain}.

\textbf{Robust scene classification.}
Once scene classification in the laboratory environment is deployed in the real application scenario, there will still be a variety of unacceptable phenomena, that is, the robustness in open environments is a bottleneck to restrict pattern recognition technology. The main reasons why the pattern recognition systems are not robust are basic assumptions, \eg closed world assumption, independent identical distribution and big data assumption~\cite{zhang2020towards}, which are main differences between machine learning and human intelligence; hence, it is a fundamental challenge to improve the robustness by breaking these assumptions. It is a nature consider via utilizing adversarial training and optimization~\cite{shaham2018understanding, qin2019adversarial, shafahi2019adversarial}, which have been applied to pattern recognition~\cite{ganin2016domain, athalye2018synthesizing}.

\textbf{Realtime scene classification.}
Many methods for scene classification, trained in a multiple-stage manner, are computationally expensive for current mobile/wearable devices, which have limited storage and computational capability, therefore researchers have begun to develop convenient and efficient unified networks (encapsulating all computation in a one-stage network)~\cite{li2017deep, liu2018dictionary, xiong2020msn}. Moreover, it is also a challenge to keep the model scalable and efficient well when big data from smart wearables and mobile applications is growing rapidly in size temporally or spatially~\cite{dargazany2018wearabledl}.

\textbf{Imbalanced scene classification.}
The Places365 challenge dataset~\cite{zhou2017places} has more than 8M training images, and the numbers of images in different classes range from 4,000 to 30,000 per class. It shows that scene categories are imbalanced, \ie some categories are abundant while others have scarce examples. Generally, the minority class samples are poorly predicted by the learned model~\cite{shen2016relay, thabtah2020data}. Therefore, learning a model which respects both type of categories and equally performs well on frequent and infrequent ones remains a great challenge and needs further exploration~\cite{buda2018systematic, johnson2019survey, thabtah2020data}.

\textbf{Continuous scene classification.}
The ultimate goal is to develop methods capable of accurately and efficiently classifying samples in thousands or more unseen scene categories in open environments~\cite{liu2020deep}. The classic deep learning paradigm learns in isolation, \ie it needs many training examples and is only suitable for well-defined tasks in closed environments~\cite{lecun2015deep, guo2016deep}. In contrast, ``human learning'' is a continuous learning and adapting to new environments: humans accumulate the knowledge gained in the past and use this knowledge to help future learning and problem solving with possible adaptations~\cite{chen2018lifelong}. Ideally, it should also be capable to discover unknown scenarios and learn new works in a self-supervised manner. Inspired by this, it is necessary to do lifelong machine learning via developing versatile systems that continually accumulate and refine their knowledge over time~\cite{thrun1995lifelong, hassabis2017neuroscience}. Such lifelong machine learning has represented a long-standing challenge for deep learning and, consequently, artificial intelligence systems.

\textbf{Multi-label scene classification.}
Many scenes are semantic multiplicity \cite{boutell2004learning, zhang2007multi}, \ie a scene image may belong to multiple semantic classes. Such a problem poses a challenge to the classic pattern recognition paradigm and requires developing multi-label learning methods~\cite{zhang2007multi, zhang2007ml}. Moreover, when constructing scene datasets, most researchers either avoid labeling multi-label images or use the most obvious class (single label) to annotate subjectively each image~\cite{boutell2004learning}. Hence, it is hard to improve the generalization ability of the model trained on single-label datasets, which also brings problems to classification task.

\textbf{Other-modal scene classification.}
RGB images provide key features such as color, texture, and spectrum of objects. Nevertheless, the scenes reproduced by RGB images may have uneven lighting, target occlusion, \emph{etc}. Therefore, the robustness of RGB scene classification is poor, and it is difficult to accurately extract key information such as target contours and spatial positions. In contrast, the rapid development of sensors has made the acquisition of other modal data easier and easier, such as RGB-D~\cite{song2015sun}, video~\cite{feichtenhofer2017temporal}, 3D point clouds~\cite{dai2017scannet}. Recently, research on recognizing and understanding various modalities has attracted an increasing attention~\cite{song2018learning, tran2015learning, qi2017pointnet}.

\appendices

\section{A Road Map of Scene Classification in 20 years}
\label{sec:twodecades}

\begin{figure*}[!htbp]
\centering
\includegraphics[width =0.98\textwidth]{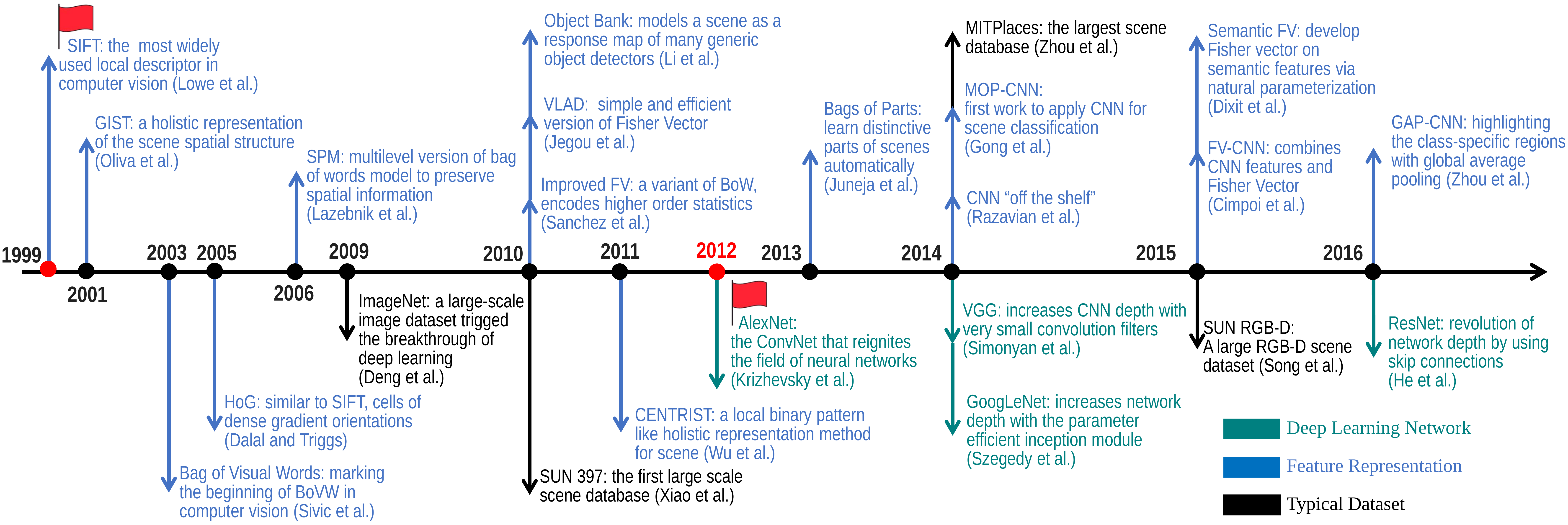}
\caption{Milestones of scene classification.
Handcrafted features gained tremendous popularity, starting from SIFT~\cite{lowe1999object} and GIST~\cite{oliva2001modeling}. Then, HoG~\cite{dalal2005histograms} and CENTRIST~\cite{wu2010centrist} were proposed by Dalal \etal~and Wu \etal, respectively, further promoting the development of scene classification.
In 2003, Sivic \etal~\cite{sivic2003video} proposed BoVW model, marking the beginning of codebook learning. Along this way, more effective BoVW based methods, SPM~\cite{lazebnik2006beyond}, IFV~\cite{sanchez2013image} and VLAD~\cite{jegou2010aggregating}, also emerged to deal with larger-scale tasks.
In 2010, Object Bank~\cite{li2010object, li2010objects} represents the scene as object attributes, marking the beginning of more semantic representations. Then, Juneja \etal~\cite{juneja2013blocks} proposed Bags of Part to learn distinctive parts of scenes automatically.
In 2012, AlexNet~\cite{krizhevsky2012imagenet} reignites the field of artificial neural networks. Since then, CNN-based methods, VGGNet~\cite{simonyan2014very}, GoogLeNet~\cite{szegedy2015going} and ResNet~\cite{he2016deep}, have begun to take over handcrafted methods. Additionally, Razavian \etal~\cite{sharif2014cnn} highlights the effectiveness and generality of CNN representations for different tasks. Along this way, in 2014, Gong \etal~\cite{gong2014multi} proposed MOP-CNN, the first deep learning methods for scene classification. Later, FV-CNN~\cite{cimpoi2015deep}, Semantic FV~\cite{dixit2015scene} and GAP-CNN~\cite{zhou2016learning} are proposed one after another to learn more effective representations.
For datasets, ImageNet \cite{deng2009imagenet} triggers the breakthrough of deep learning. Then, Xiao \etal~\cite{xiao2010sun} proposed SUN database to evaluate numerous algorithms for scene classification. Later, Places~\cite{zhou2014learning, zhou2017places}, the largest scene database currently, emerged to satisfy the need of deep learning training. Additionally, SUN RGBD~\cite{song2015sun} has been introduced, marking the beginning of deep learning for RGB-D scene classification.
}
\label{fig:milestones}
\end{figure*}

Scene representation or scene feature extraction, the process of converting a scene image into feature vectors, plays the critical role in scene classification, and thus is the focus of research in this field. In the past two decades, remarkable progress has been witnessed in scene representation, which mainly consists of two important generations: handcrafted feature engineering, and deep learning (feature learning). The milestones of scene classification in the past two decades are presented in Fig.~\ref{fig:milestones}, in which two main stages (SIFT vs. DNN) are highlighted.

\textbf{Handcrafted feature engineering era.} From 1995 to 2012, the field was dominated by the Bag of Visual Word (BoVW) model~\cite{vasconcelos2000probabilistic, csurka2004visual, sivic2003video, wallraven2003recognition} borrowed from document classification which represents a document as a vector of word occurrence counts over a global word vocabulary. In the image domain, BoVW firstly probes an image with local feature descriptors such as Scale Invariant Feature Transform (SIFT)~\cite{lowe1999object, lowe2004distinctive}, and then represents an image statistically as an orderless histogram over a pre-trained visual vocabulary, in a similar form to a document. Some important variants of BoVW such as Bag of Semantics~\cite{li2010object, kwitt2012scene, rasiwasia2012holistic, juneja2013blocks} and Improved Fisher Vector (IFV) \cite{sanchez2013image}, have also been proposed.

Local invariant feature descriptors play an important role in BoVW because they are discriminative, yet less sensitive to image variations such as illumination, scale, rotation, viewpoint \emph{etc}, and thus have been widely studied. Representative local descriptors for scene classification have started from SIFT~\cite{lowe1999object, lowe2004distinctive} and Global Information Systems Technology (GIST)~\cite{oliva2001modeling, oliva2006building}. Other local descriptors, such as Local Binary Patterns (LBP)~\cite{ojala2002multiresolution}, Deformable Part Model (DPM)~\cite{felzenszwalb2008discriminatively, felzenszwalb2009object, pandey2011scene}, CENsus TRansform hISTogram (CENTRIST)~\cite{wu2010centrist}, also contribute to the development of scene classification. To improve the performance, research focus shifts to feature encoding and aggregation, mainly including Bag-of-Visual-Words (BoVW)~\cite{csurka2004visual}, Latent Dirichlet Allocation (LDA)~\cite{blei2003latent}, Histogram of Gradients (HoG)~\cite{dalal2005histograms}, Spatial Pyramid Matching (SPM)~\cite{grauman2005pyramid, lazebnik2006beyond}, Vector of Locally Aggregated Descriptors (VLAD)~\cite{jegou2010aggregating}, Fisher kernel coding~\cite{perronnin2007fisher, sanchez2013image}, Multi-Resolution BoVW (MR-BoVW)~\cite{zhou2013scene}, and Orientational Pyramid Matching (OPM) \cite{xie2014orientational}. The quality of the learned codebook has a great impact on the coding procedure. The generic codebooks mainly include Fisher kernels~\cite{perronnin2007fisher, sanchez2013image}, sparse codebook~\cite{yang2009linear, gao2010local}, Locality-constrained Linear Codes (LLC)~\cite{wang2010locality}, Histogram Intersection Kernels (HIK)~\cite{wu2009beyond}, contextual visual words~\cite{qin2010scene}, Efficient Match Kernels (EMK)~\cite{bo2009efficient} and Supervised Kernel Descriptors (SKDES)~\cite{wang2013supervised}. Particularly, semantic codebooks generate from salient regions, like Object Bank~\cite{li2010objects, li2010object, li2014object}, object-to-class \cite{zhang2014learning}, Latent Pyramidal Regions (LPR)~\cite{sadeghi2012latent}, Bags of Parts (BoP)~\cite{juneja2013blocks} and Pairwise Constraints based Multiview Subspace Learning (PC-MSL)~\cite{yu2013pairwise}, capturing more discriminative features for scene classification.

\textbf{Deep learning era.} In 2012, Krizhevsky~\etal~\cite{krizhevsky2012imagenet} introduced a DNN, commonly referred to as ``AlexNet'', for the object classification task, and achieved breakthrough performance surpassing the best result of hand-engineered features by a large margin, and thus triggered the recent revolution in AI. Since then, deep learning has started to dominate various tasks (like computer vision~\cite{girshick2014rich, taigman2014deepface, liu2020deep, guo2020deep}, speech recognition~\cite{hinton2012deep}, autonomous driving~\cite{chen2015deepdriving}, cancer detection~\cite{esteva2017dermatologist, mckinney2020international}, machine translation~\cite{wu2016google}, playing complex games~\cite{silver2016mastering, silver2017mastering, silver2018general, vinyals2019grandmaster}, earthquake forecasting~\cite{devries2018deep}, medicine discovery~\cite{mit2020breakthrough, stokes2020deep}), and scene classification is no exception, leading to a new generation of scene representation methods with remarkable performance improvements. Such substantial progress can be mainly attributed to advances in deep models including VGGNet \cite{simonyan2014very}, GoogLeNet \cite{szegedy2015going}, ResNet \cite{he2016deep}, \emph{etc.}, the availability of large-scale image datasets like ImageNet~\cite{deng2009imagenet} and Places~\cite{zhou2014learning, zhou2017places} and more powerful computational resources.

Deep learning networks have gradually replaced the local feature descriptors of the first generation methods and are certainly the engine for scene classification. Although the major driving force of progress in scene classification has been the incorporation of deep learning networks, the general pipelines like BoVW, feature encoding and aggregation methods like Fisher Vector, VLAD of the first generation methods have also been adapted in current deep learning based scene methods, \eg MOP-CNN~\cite{gong2014multi}, SCFVC~\cite{liu2014encoding}, MPP-CNN~\cite{yoo2015multi}, DSP~\cite{gao2015deep}, Semantic FV~\cite{dixit2015scene}, LatMoG~\cite{cinbis2015approximate}, MFA-FS~\cite{dixit2016object} and DUCA~\cite{khan2016discriminative}. To take fully advantage of back-propagation, scene representations are extracted from end-to-end trainable CNNs, like DAG-CNN~\cite{yang2015multi}, MFAFVNet \cite{li2017deep}, VSAD \cite{wang2017weakly}, G-MS2F~\cite{tang2017g}, and DL-CNN~\cite{liu2018dictionary}. To focus on main content of the scene, object detection is used to capture salient regions, such as MetaObject-CNN \cite{wu2015harvesting}, WELDON \cite{durand2016weldon}, SDO \cite{cheng2018scene}, and BiLSTM \cite{laranjeira2019modeling}. Since features from multiple CNN layers or multiple views are complementary, many literatures~\cite{herranz2016scene, guo2016locally, wang2017knowledge, sun2018fusing, liu2019novel} also explored their complementarity to improve performance. In addition, there exists many strategies (like attention mechanism, contextual modeling, multi-task learning with regularization terms) to enhance representation ability, such as  CFA \cite{sun2018fusing}, BiLSTM \cite{laranjeira2019modeling}, MAPNet~\cite{li2019mapnet}, MSN~\cite{xiong2020msn},  and LGN~\cite{chen2020scene}. For datasets, because depth images from RGB-D cameras are not vulnerable to illumination changes, since 2015, researchers have started to explore RGB-D scene recognition. Some works~\cite{song2017combining, song2018learning, du2019translate} focus on depth-specific feature learning, while other alternatives, like DMMF~\cite{zhu2016discriminative}, ACM~\cite{yuan2019acm}, and MSN~\cite{xiong2020msn} focus on multi-modal feature fusion.

\section{A Brief Introduction to Deep Learning}
\label{sec:brief}

\begin{figure*}[!htbp]
\centering
\subfigure[VGG]{{\includegraphics[width = .6\textwidth]{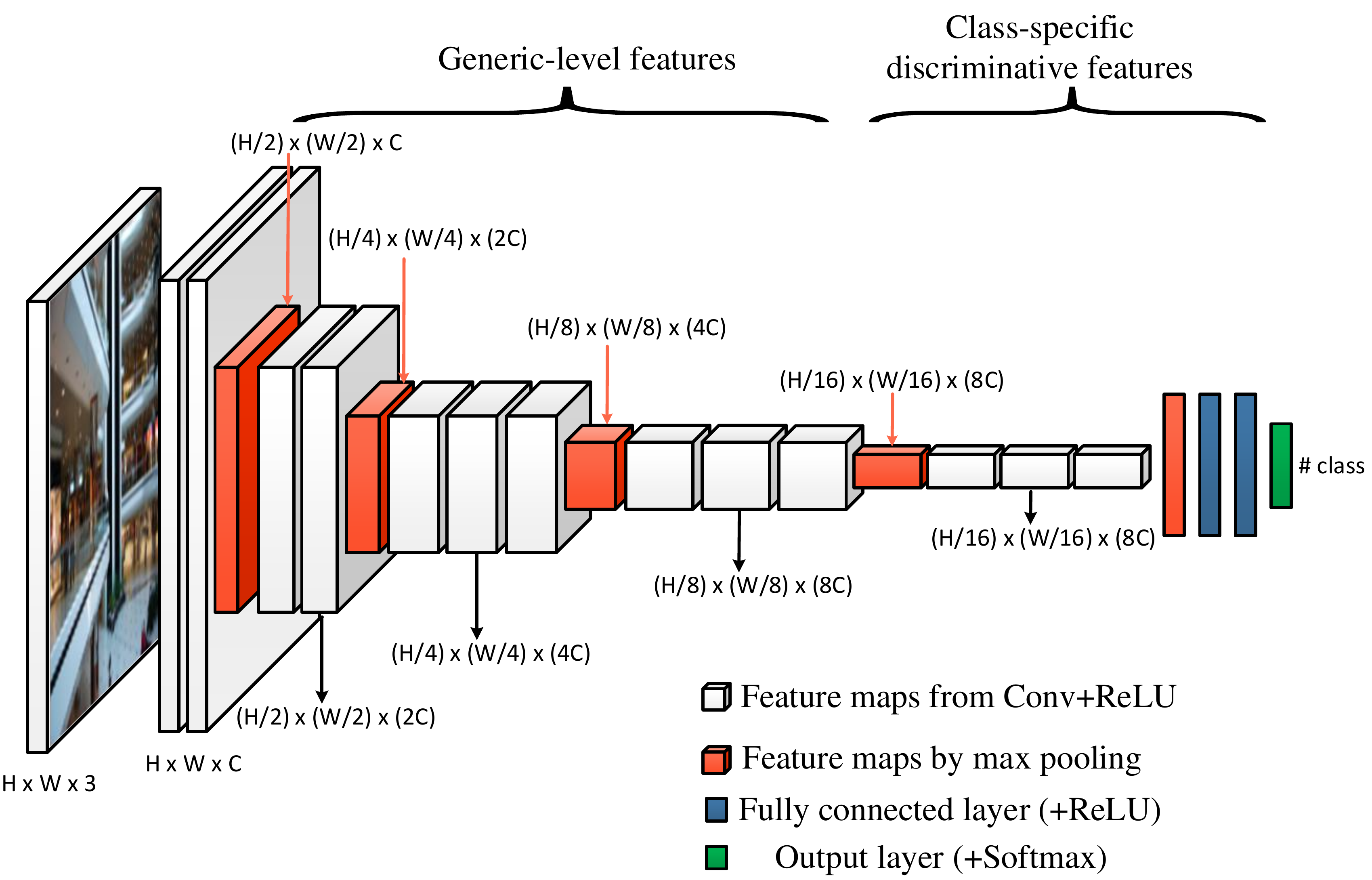}}}
\subfigure[Convolution operation]{{\includegraphics[width = .35\textwidth]{CONVexplain.pdf}}}
\caption{\textcolor{black}{(a) A typical CNN architecture VGGNet \cite{simonyan2014very} with 16 weight layers. The network has 13 convolutional layers, 5 max pooling layers (The last one is global max pooling) and 3 fully-connected layers (The last one uses Softmax function as nonlinear activation function). The whole network can be learned from labeled training data by optimizing a loss function (\eg, cross-entropy loss). 
(b) Illustration of basic operations (\ie, convolution, nonlinearity and downsampling) that are repeatedly applied for a typical CNN. (b1) The outputs (called the feature maps) of each layer (horizontally) of a typical CNN applied to a scene image. Each feature map in the second row corresponds to the output for one of the learned 3D filters (see (b2)).}}
\label{fig:CNNandConv}
\end{figure*}

\textcolor{black}{In 2012, the breakthrough object classification result on the large scale ImageNet dataset~\cite{deng2009imagenet} achieved by a deep learning network~\cite{krizhevsky2012imagenet} is arguably what reignited the field of Artificial Neural Networks (ANNs) and triggered the recent revolution in AI. Since then, deep learning, or Deep Neural Networks (DNNs)~\cite{schmidhuber2015deep}, shines in a broad range of areas, including computer vision~\cite{krizhevsky2012imagenet, girshick2014rich, liu2020deep, guo2020deep, taigman2014deepface}, speech recognition~\cite{hinton2012deep}, autonomous driving~\cite{chen2015deepdriving}, cancer detection~\cite{esteva2017dermatologist, mckinney2020international}, machine translation~\cite{wu2016google}, playing complex games~\cite{silver2016mastering, silver2017mastering, silver2018general, vinyals2019grandmaster}, earthquake forecasting~\cite{devries2018deep}, medicine discovery~\cite{mit2020breakthrough, stokes2020deep}, \emph{etc.} In many of these domains, DNNs have reached breakthrough levels of performance, often approaching and sometimes exceeding the abilities of human experts. Thanks to the growth of big data and more powerful computational resources, deep learning and AI for scientific research are evolving quickly, with new developments appearing continually for analyzing datasets, discovering patterns, and predicting behaviors in almost all fields of science and technology~\cite{sejnowski2020unreasonable}.}

 \textcolor{black}{In computer vision, the most commonly used type of DNNs is Convolutional Neural Networks (CNNs)~\cite{lecun2015deep}.}  As is illustrated in Figure~\ref{fig:CNNandConv} (a), the frontend of a typical CNN is a stack of CONV layers and pooling layers to learn generic-level features, and these features are further transformed into class-specific discriminative representations via training multiple layers on target datasets. As we slide a convolution filter over the width and height of the input of 3 color channels, we will produce a two-dimensional (2-D) activation map, as shown in Figure~\ref{fig:CNNandConv}~(b1), giving the responses of that filter at every spatial position. As shown in Figure~\ref{fig:CNNandConv}~(b2), a 2-D convolution operates $x^{l-1}*w^{l}$, describing as a weighted average of an input map $x^{l-1}$ from previous layer $l-1$, where the corresponding weighting is given by $w^{l}$; since every filter extends through the full depth of the input maps with $C$ channels, so we calculate the sum of 2-D convolution as the result of a 3-D convolution, \ie
\begin{equation}
\sum_{i=1}^{C}{x_{i}^{l-1}*w_{i}^{l}}
\end{equation}
During the forward pass, the $j$ neuron in $l$ CONV layer operates a 3-D convolution with $N^{l-1}$ channels between input maps $x^{l-1}$ and the corresponding filter $w_{j}^{l}$, plus a bias term $b_{j}^{l}$, and passes the above result to a nonlinear function $\sigma(\cdot)$ to obtain the final output $x_{j}^{l}$, \ie
\begin{equation}
x_{j}^{l}=\sigma(\sum_{i=1}^{N^{l-1}}{x_{i}^{l-1}}*w_{i, j}^{l}+b_{j}^{l})
\end{equation}
The nonlinear function $\sigma(\cdot)$ is typically a rectified linear unit (ReLU) for an element $x$, \ie $\sigma(x)=\mathrm{max}(x, 0)$.

\textcolor{black}{It is common to periodically insert a pooling (\ie downsampling) in-between successive convolutional layers in a CNN architecture. Its function is to progressively reduce the spatial size of the input maps, as shown in Figure~\ref{fig:CNNandConv}~(b1), to reduce the number of parameters and computation in the network, and hence to also control overfitting.} Finally, the output layer expresses a differentiable score function: from the discriminative representations $x$ on one end to class scores $y$ at the other.

Supervised by scene labels, Softmax classifier uses cross-entropy loss function to estimate model parameters, and formulas are as follow:
\begin{equation}
L_{softmax}=\sum_{i}^{N}L_{CE}(x_{i})
\end{equation}
where $L_{CE}(x_{i})$ denotes the cross-entropy loss function of each image and its formula is as follows:
\begin{equation}
L_{CE}(x_{i})=-\sum_{k=1}^{K}(y_{k}\text{log}f_{k}(x_{i}))
\end{equation}
where $x_{i}$ denotes a discriminative feature of scene image $I_{i}$; $K$ denotes the count of scene categories; $y_{k}$ denotes the real label of scene image $I_{i}$, when $y_{k}=1$ represents that the real label of $I_{i}$ is category $k$, otherwise $y_{k}=0$.

Generally, the score function $f(\cdot)$ is Softmax function, denoting the probability of estimating scene image $I_{i}$ in class $k$.

so the $k$ component of output layer is
\begin{equation}
f_{k}(x)=e^{w_{k}x_{k}}/\sum_{j=1}^{K}{e^{w_{j}x_{j}}}
\end{equation}

In addition to the CNN architectures and datasets, some other training strategies or tricks have been proposed to achieve better performance. Overfitting happens when CNNs learn the detail and noise in the training data to the extent that it negatively impacts the generalization ability of models. To this end, some strategies have been proposed to reduce the overfitting tendency of CNNs, such as data augmentation, early stopping, dropout \cite{krizhevsky2012imagenet}, smaller convolution kernel size \cite{simonyan2014very, zeiler2014visualizing}, and multi-scale cropping/warping \cite{simonyan2014very}. In addition, some optimization techniques have been proposed to  overcome the difficulties encountered in the training of deep CNNs, such as batch normalization \cite{ioffe2015batch, ioffe2017batch} and relay back propagation \cite{shen2016relay}. \textcolor{black}{Comprehensive review of deep learning is out the scope of this survey, and we refer readers to~\cite{lecun2015deep, goodfellow2016deep, litjens2017survey, gu2018recent, pouyanfar2018survey} for more details.}

\ifCLASSOPTIONcompsoc
  \section*{Acknowledgments}
\else
  \section*{Acknowledgment}
\fi

The authors would like to thank the pioneer researchers in scene classification and other related fields. This work was supported in part by grants from National Science Foundation of China (61872379, 91846301, 61571005), the Academy of Finland (331883), the fundamental research program of Guangdong, China (2020B1515310023), the Hunan Science and Technology Plan Project (2019GK2131), the China Scholarship Council (201806155037), the Science and Technology Research Program of Guangzhou, China (201804010429).

\ifCLASSOPTIONcaptionsoff
  \newpage
\fi

\bibliographystyle{IEEEtran}
\bibliography{bibrefs}

\begin{thebibliography}{100}
\providecommand{\url}[1]{#1}
\csname url@samestyle\endcsname
\providecommand{\newblock}{\relax}
\providecommand{\bibinfo}[2]{#2}
\providecommand{\BIBentrySTDinterwordspacing}{\spaceskip=0pt\relax}
\providecommand{\BIBentryALTinterwordstretchfactor}{4}
\providecommand{\BIBentryALTinterwordspacing}{\spaceskip=\fontdimen2\font plus
\BIBentryALTinterwordstretchfactor\fontdimen3\font minus
  \fontdimen4\font\relax}
\providecommand{\BIBforeignlanguage}[2]{{%
\expandafter\ifx\csname l@#1\endcsname\relax
\typeout{** WARNING: IEEEtran.bst: No hyphenation pattern has been}%
\typeout{** loaded for the language `#1'. Using the pattern for}%
\typeout{** the default language instead.}%
\else
\language=\csname l@#1\endcsname
\fi
#2}}
\providecommand{\BIBdecl}{\relax}
\BIBdecl

\bibitem{henderson1999high}
J.~M. Henderson and A.~Hollingworth, ``High-level scene perception,''
  \emph{Annual review of psychology}, vol.~50, no.~1, pp. 243--271, 1999.

\bibitem{epstein2005cortical}
R.~Epstein, ``The cortical basis of visual scene processing,'' \emph{Visual
  Cognition}, vol.~12, no.~6, pp. 954--978, 2005.

\bibitem{greene2009briefest}
M.~R. Greene and A.~Oliva, ``The briefest of glances: {T}he time course of
  natural scene understanding,'' \emph{Psychological Science}, vol.~20, no.~4,
  pp. 464--472, 2009.

\bibitem{walther2011simple}
D.~B. Walther, B.~Chai, E.~Caddigan, D.~M. Beck, and L.~Fei-Fei, ``Simple line
  drawings suffice for functional {MRI} decoding of natural scene categories,''
  \emph{PNAS}, vol. 108, no.~23, pp. 9661--9666, 2011.

\bibitem{vogel2007semantic}
J.~Vogel and B.~Schiele, ``Semantic modeling of natural scenes for
  content-based image retrieval,'' \emph{IJCV}, vol.~72, no.~2, pp. 133--157,
  2007.

\bibitem{zheng2017sift}
L.~Zheng, Y.~Yang, and Q.~Tian, ``{SIFT} meets {CNN}: {A} decade survey of
  instance retrieval,'' \emph{IEEE TPAMI}, vol.~40, no.~5, pp. 1224--1244,
  2017.

\bibitem{zhang2017learning}
W.~Zhang, X.~Yu, and X.~He, ``Learning bidirectional temporal cues for
  video-based person re-identification,'' \emph{IEEE TCSVT}, vol.~28, no.~10,
  pp. 2768--2776, 2017.

\bibitem{hou2019deep}
J.~Hou, H.~Zeng, J.~Zhu, J.~Hou, J.~Chen, and K.-K. Ma, ``Deep quadruplet
  appearance learning for vehicle re-identification,'' \emph{IEEE TVT},
  vol.~68, no.~9, pp. 8512--8522, 2019.

\bibitem{zhang2012mining}
T.~Zhang, S.~Liu, C.~Xu, and H.~Lu, ``Mining semantic context information for
  intelligent video surveillance of traffic scenes,'' \emph{IEEE TII}, vol.~9,
  no.~1, pp. 149--160, 2012.

\bibitem{sreenu2019intelligent}
G.~Sreenu and M.~S. Durai, ``Intelligent video surveillance: {A} review through
  deep learning techniques for crowd analysis,'' \emph{Journal of Big Data},
  vol.~6, no.~1, p.~48, 2019.

\bibitem{behzadan2011integrated}
A.~H. Behzadan and V.~R. Kamat, ``Integrated information modeling and visual
  simulation of engineering operations using dynamic augmented reality scene
  graphs,'' \emph{ITcon}, vol.~16, no.~17, pp. 259--278, 2011.

\bibitem{nee2012augmented}
A.~Y. Nee, S.~Ong, G.~Chryssolouris, and D.~Mourtzis, ``Augmented reality
  applications in design and manufacturing,'' \emph{CIRP annals}, vol.~61,
  no.~2, pp. 657--679, 2012.

\bibitem{muhammad2018early}
K.~Muhammad, J.~Ahmad, and S.~W. Baik, ``Early fire detection using
  convolutional neural networks during surveillance for effective disaster
  management,'' \emph{Neurocomputing}, vol. 288, pp. 30--42, 2018.

\bibitem{lazebnik2006beyond}
S.~Lazebnik, C.~Schmid, and J.~Ponce, ``Beyond bags of features: {S}patial
  pyramid matching for recognizing natural scene categories,'' in \emph{CVPR},
  vol.~2, 2006, pp. 2169--2178,
  \url{https://figshare.com/articles/15-Scene_Image_Dataset/7007177}.

\bibitem{li2010object}
L.-J. Li, H.~Su, L.~Fei-Fei, and E.~P. Xing, ``Object bank: {A} high-level
  image representation for scene classification \& semantic feature
  sparsification,'' in \emph{NeurIPS}, 2010, pp. 1378--1386.

\bibitem{margolin2014otc}
R.~Margolin, L.~Zelnik-Manor, and A.~Tal, ``{OTC}: {A} novel local descriptor
  for scene classification,'' in \emph{ECCV}, 2014, pp. 377--391.

\bibitem{krizhevsky2012imagenet}
A.~Krizhevsky, I.~Sutskever, and G.~E. Hinton, ``Imagenet classification with
  deep convolutional neural networks,'' in \emph{NeurIPS}, 2012, pp.
  1097--1105.

\bibitem{zhou2014learning}
B.~Zhou, A.~Lapedriza, J.~Xiao, A.~Torralba, and A.~Oliva, ``Learning deep
  features for scene recognition using places database,'' in \emph{NeurIPS},
  2014, pp. 487--495, \url{http://places.csail.mit.edu/downloadData.html}.

\bibitem{herranz2016scene}
L.~Herranz, S.~Jiang, and X.~Li, ``Scene recognition with {CNNs}: {O}bjects,
  scales and dataset bias,'' in \emph{CVPR}, 2016, pp. 571--579.

\bibitem{khan2016discriminative}
S.~H. Khan, M.~Hayat, M.~Bennamoun, R.~Togneri, and F.~A. Sohel, ``A
  discriminative representation of convolutional features for indoor scene
  recognition,'' \emph{IEEE TIP}, vol.~25, no.~7, pp. 3372--3383, 2016.

\bibitem{liu2019novel}
S.~Liu, G.~Tian, and Y.~Xu, ``A novel scene classification model combining
  {ResNet} based transfer learning and data augmentation with a filter,''
  \emph{Neurocomputing}, vol. 338, pp. 191--206, 2019.

\bibitem{liu2018dictionary}
Y.~Liu, Q.~Chen, W.~Chen, and I.~Wassell, ``Dictionary learning inspired deep
  network for scene recognition,'' in \emph{AAAI}, 2018.

\bibitem{zhou2016learning}
B.~Zhou, A.~Khosla, A.~Lapedriza, A.~Oliva, and A.~Torralba, ``Learning deep
  features for discriminative localization,'' in \emph{CVPR}, 2016, pp.
  2921--2929.

\bibitem{sun2018fusing}
N.~Sun, W.~Li, J.~Liu, G.~Han, and C.~Wu, ``Fusing object semantics and deep
  appearance features for scene recognition,'' \emph{IEEE TCSVT}, vol.~29,
  no.~6, pp. 1715--1728, 2018.

\bibitem{zhou2017places}
B.~Zhou, A.~Lapedriza, A.~Khosla, A.~Oliva, and A.~Torralba, ``Places: {A} 10
  million image database for scene recognition,'' \emph{IEEE TPAMI}, vol.~40,
  no.~6, pp. 1452--1464, 2017,
  \url{http://places2.csail.mit.edu/download.html}.

\bibitem{hayat2016spatial}
M.~Hayat, S.~H. Khan, M.~Bennamoun, and S.~An, ``A spatial layout and scale
  invariant feature representation for indoor scene classification,''
  \emph{IEEE TIP}, vol.~25, no.~10, pp. 4829--4841, 2016.

\bibitem{zuo2016learning}
Z.~Zuo, B.~Shuai, G.~Wang, X.~Liu, X.~Wang, B.~Wang, and Y.~Chen, ``Learning
  contextual dependence with convolutional hierarchical recurrent neural
  networks,'' \emph{IEEE TIP}, vol.~25, no.~7, pp. 2983--2996, 2016.

\bibitem{gong2014multi}
Y.~Gong, L.~Wang, R.~Guo, and S.~Lazebnik, ``Multi-scale orderless pooling of
  deep convolutional activation features,'' in \emph{ECCV}, 2014, pp. 392--407.

\bibitem{yoo2015multi}
D.~Yoo, S.~Park, J.-Y. Lee, and I.~So~Kweon, ``Multi-scale pyramid pooling for
  deep convolutional representation,'' in \emph{CVPRW}, 2015, pp. 71--80.

\bibitem{dixit2015scene}
M.~Dixit, S.~Chen, D.~Gao, N.~Rasiwasia, and N.~Vasconcelos, ``Scene
  classification with semantic fisher vectors,'' in \emph{CVPR}, 2015, pp.
  2974--2983.

\bibitem{li2017deep}
Y.~Li, M.~Dixit, and N.~Vasconcelos, ``Deep scene image classification with the
  {MFAFVNet},'' in \emph{ICCV}, 2017, pp. 5746--5754.

\bibitem{wang2017weakly}
Z.~Wang, L.~Wang, Y.~Wang, B.~Zhang, and Y.~Qiao, ``Weakly supervised
  patchnets: {D}escribing and aggregating local patches for scene
  recognition,'' \emph{IEEE TIP}, vol.~26, no.~4, pp. 2028--2041, 2017.

\bibitem{wu2015harvesting}
R.~Wu, B.~Wang, W.~Wang, and Y.~Yu, ``Harvesting discriminative meta objects
  with deep {CNN} features for scene classification,'' in \emph{ICCV}, 2015,
  pp. 1287--1295.

\bibitem{durand2016weldon}
T.~Durand, N.~Thome, and M.~Cord, ``{WELDON}: {W}eakly supervised learning of
  deep convolutional neural networks,'' in \emph{CVPR}, 2016, pp. 4743--4752.

\bibitem{cheng2018scene}
X.~Cheng, J.~Lu, J.~Feng, B.~Yuan, and J.~Zhou, ``Scene recognition with
  objectness,'' \emph{Pattern Recognition}, vol.~74, pp. 474--487, 2018.

\bibitem{laranjeira2019modeling}
C.~Laranjeira, A.~Lacerda, and E.~R. Nascimento, ``On modeling context from
  objects with a long short-term memory for indoor scene recognition,'' in
  \emph{SIBGRAPI}, 2019, pp. 249--256.

\bibitem{yang2015multi}
S.~Yang and D.~Ramanan, ``Multi-scale recognition with {DAG-CNNs},'' in
  \emph{ICCV}, 2015, pp. 1215--1223.

\bibitem{xie2015hybrid}
G.-S. Xie, X.-Y. Zhang, S.~Yan, and C.-L. Liu, ``Hybrid {CNN} and
  dictionary-based models for scene recognition and domain adaptation,''
  \emph{IEEE TCSVT}, vol.~27, no.~6, pp. 1263--1274, 2015.

\bibitem{tang2017g}
P.~Tang, H.~Wang, and S.~Kwong, ``{G-MS2F}: {G}ooglenet based multi-stage
  feature fusion of deep {CNN} for scene recognition,'' \emph{Neurocomputing},
  vol. 225, pp. 188--197, 2017.

\bibitem{guo2016locally}
S.~Guo, W.~Huang, L.~Wang, and Y.~Qiao, ``Locally supervised deep hybrid model
  for scene recognition,'' \emph{IEEE TIP}, vol.~26, no.~2, pp. 808--820, 2016.

\bibitem{wang2017knowledge}
L.~Wang, S.~Guo, W.~Huang, Y.~Xiong, and Y.~Qiao, ``Knowledge guided
  disambiguation for large-scale scene classification with multi-resolution
  {CNNs},'' \emph{IEEE TIP}, vol.~26, no.~4, pp. 2055--2068, 2017.

\bibitem{dixit2016object}
M.~D. Dixit and N.~Vasconcelos, ``Object based scene representations using
  fisher scores of local subspace projections,'' in \emph{NeurIPS}, 2016, pp.
  2811--2819.

\bibitem{lopez2020semantic}
A.~L{\'o}pez-Cifuentes, M.~Escudero-Vi{\~n}olo, J.~Besc{\'o}s, and
  {\'A}.~Garc{\'\i}a-Mart{\'\i}n, ``Semantic-aware scene recognition,''
  \emph{Pattern Recognition}, vol. 102, p. 107256, 2020.

\bibitem{xiong2020msn}
Z.~Xiong, Y.~Yuan, and Q.~Wang, ``{MSN}: {M}odality separation networks for
  {RGB-D} scene recognition,'' \emph{Neurocomputing}, vol. 373, pp. 81--89,
  2020.

\bibitem{chen2020scene}
G.~Chen, X.~Song, H.~Zeng, and S.~Jiang, ``Scene recognition with
  prototype-agnostic scene layout,'' \emph{IEEE TIP}, vol.~29, pp. 5877--5888,
  2020.

\bibitem{wang2016modality}
A.~Wang, J.~Cai, J.~Lu, and T.-J. Cham, ``Modality and component aware feature
  fusion for {RGB-D} scene classification,'' in \emph{CVPR}, 2016, pp.
  5995--6004.

\bibitem{song2017combining}
X.~Song, S.~Jiang, and L.~Herranz, ``Combining models from multiple sources for
  {RGB-D} scene recognition.'' in \emph{IJCAI}, 2017, pp. 4523--4529.

\bibitem{du2019translate}
D.~Du, L.~Wang, H.~Wang, K.~Zhao, and G.~Wu, ``Translate-to-recognize networks
  for {RGB-D} scene recognition,'' in \emph{CVPR}, 2019, pp. 11\,836--11\,845.

\bibitem{cheng2017locality}
Y.~Cheng, R.~Cai, Z.~Li, X.~Zhao, and K.~Huang, ``Locality-sensitive
  deconvolution networks with gated fusion for {RGB-D} indoor semantic
  segmentation,'' in \emph{CVPR}, 2017, pp. 3029--3037.

\bibitem{zhu2016discriminative}
H.~Zhu, J.-B. Weibel, and S.~Lu, ``Discriminative multi-modal feature fusion
  for {RGB-D} indoor scene recognition,'' in \emph{CVPR}, 2016, pp. 2969--2976.

\bibitem{girshick2015fast}
R.~Girshick, ``Fast {R-CNN},'' in \emph{ICCV}, 2015, pp. 1440--1448.

\bibitem{ren2016faster}
S.~Ren, K.~He, R.~Girshick, and J.~Sun, ``Faster {R-CNN}: {T}owards real-time
  object detection with region proposal networks,'' \emph{IEEE TPAMI}, vol.~39,
  no.~6, pp. 1137--1149, 2016.

\bibitem{badrinarayanan2017segnet}
V.~Badrinarayanan, A.~Kendall, and R.~Cipolla, ``{SegNet}: {A} deep
  convolutional encoder-decoder architecture for image segmentation,''
  \emph{IEEE TPAMI}, vol.~39, no.~12, pp. 2481--2495, 2017.

\bibitem{chen2017deeplab}
L.-C. Chen, G.~Papandreou, I.~Kokkinos, K.~Murphy, and A.~L. Yuille, ``Deeplab:
  {S}emantic image segmentation with deep convolutional nets, atrous
  convolution, and fully connected crfs,'' \emph{IEEE TPAMI}, vol.~40, no.~4,
  pp. 834--848, 2017.

\bibitem{cai2018menet}
S.~Cai, J.~Huang, D.~Zeng, X.~Ding, and J.~Paisley, ``{ME}net: {A} metric
  expression network for salient object segmentation,'' in \emph{IJCAI}, 2018,
  pp. 598--605.

\bibitem{deng2009imagenet}
J.~Deng, W.~Dong, R.~Socher, L.-J. Li, K.~Li, and L.~Fei-Fei, ``Imagenet: {A}
  large-scale hierarchical image database,'' in \emph{CVPR}, 2009, pp.
  248--255, \url{http://image-net.org/download}.

\bibitem{xiao2010sun}
J.~Xiao, J.~Hays, K.~A. Ehinger, A.~Oliva, and A.~Torralba, ``Sun database:
  {L}arge-scale scene recognition from abbey to zoo,'' in \emph{CVPR}, 2010,
  pp. 3485--3492, \url{http://places2.csail.mit.edu/download.html}.

\bibitem{cinbis2015approximate}
R.~G. Cinbis, J.~Verbeek, and C.~Schmid, ``Approximate fisher kernels of
  non-iid image models for image categorization,'' \emph{IEEE TPAMI}, vol.~38,
  no.~6, pp. 1084--1098, 2015.

\bibitem{wei2016visual}
X.~Wei, S.~L. Phung, and A.~Bouzerdoum, ``Visual descriptors for scene
  categorization: {E}xperimental evaluation,'' \emph{AI Review}, vol.~45,
  no.~3, pp. 333--368, 2016.

\bibitem{cheng2017remote}
G.~Cheng, J.~Han, and X.~Lu, ``Remote sensing image scene classification:
  {B}enchmark and state of the art,'' \emph{Proceedings of the IEEE}, vol. 105,
  no.~10, pp. 1865--1883, 2017.

\bibitem{xie2020scene}
L.~Xie, F.~Lee, L.~Liu, K.~Kotani, and Q.~Chen, ``Scene recognition: {A}
  comprehensive survey,'' \emph{Pattern Recognition}, vol. 102, p. 107205,
  2020.

\bibitem{nogueira2017towards}
K.~Nogueira, O.~A. Penatti, and J.~A. Dos~Santos, ``Towards better exploiting
  convolutional neural networks for remote sensing scene classification,''
  \emph{Pattern Recognition}, vol.~61, pp. 539--556, 2017.

\bibitem{hu2015transferring}
F.~Hu, G.-S. Xia, J.~Hu, and L.~Zhang, ``Transferring deep convolutional neural
  networks for the scene classification of high-resolution remote sensing
  imagery,'' \emph{Remote Sensing}, vol.~7, no.~11, pp. 14\,680--14\,707, 2015.

\bibitem{mesaros2016tut}
A.~Mesaros, T.~Heittola, and T.~Virtanen, ``Tut database for acoustic scene
  classification and sound event detection,'' in \emph{EUSIPCO}, 2016, pp.
  1128--1132.

\bibitem{ren2018deep}
Z.~Ren, K.~Qian, Y.~Wang, Z.~Zhang, V.~Pandit, A.~Baird, and B.~Schuller,
  ``Deep scalogram representations for acoustic scene classification,''
  \emph{JAS}, vol.~5, no.~3, pp. 662--669, 2018.

\bibitem{lowry2015visual}
S.~Lowry, N.~S{\"u}nderhauf, P.~Newman, J.~J. Leonard, D.~Cox, P.~Corke, and
  M.~J. Milford, ``Visual place recognition: A survey,'' \emph{IEEE T-RO},
  vol.~32, no.~1, pp. 1--19, 2015.

\bibitem{arandjelovic2016netvlad}
R.~Arandjelovic, P.~Gronat, A.~Torii, T.~Pajdla, and J.~Sivic, ``{NetVLAD}:
  {CNN} architecture for weakly supervised place recognition,'' in \emph{CVPR},
  2016, pp. 5297--5307.

\bibitem{boutell2004learning}
M.~R. Boutell, J.~Luo, X.~Shen, and C.~M. Brown, ``Learning multi-label scene
  classification,'' \emph{Pattern Recognition}, vol.~37, no.~9, pp. 1757--1771,
  2004.

\bibitem{zhang2007multi}
M.-L. Zhang and Z.-H. Zhou, ``Multi-label learning by instance
  differentiation,'' in \emph{AAAI}, vol.~7, 2007, pp. 669--674.

\bibitem{quattoni2009recognizing}
A.~Quattoni and A.~Torralba, ``Recognizing indoor scenes,'' in \emph{CVPR},
  2009, pp. 413--420, \url{http://web.mit.edu/torralba/www/indoor.html}.

\bibitem{silberman2012indoor}
N.~Silberman, D.~Hoiem, P.~Kohli, and R.~Fergus, ``Indoor segmentation and
  support inference from {RGB-D} images,'' in \emph{ECCV}, 2012, pp. 746--760,
  \url{https://cs.nyu.edu/~silberman/datasets/nyu_depth_v2.html}.

\bibitem{song2015sun}
S.~Song, S.~P. Lichtenberg, and J.~Xiao, ``{SUN RGB-D}: {A RGB-D} scene
  understanding benchmark suite,'' in \emph{CVPR}, 2015, pp. 567--576,
  \url{https://github.com/ankurhanda/sunrgbd-meta-data}.

\bibitem{gupta2014learning}
S.~Gupta, R.~Girshick, P.~Arbel{\'a}ez, and J.~Malik, ``Learning rich features
  from {RGB-D} images for object detection and segmentation,'' in \emph{ECCV},
  2014, pp. 345--360.

\bibitem{miller1995wordnet}
G.~A. Miller, ``Wordnet: {A} lexical database for english,''
  \emph{Communications of the ACM}, vol.~38, no.~11, pp. 39--41, 1995.

\bibitem{everingham2010pascal}
M.~Everingham, L.~Van~Gool, C.~K. Williams, J.~Winn, and A.~Zisserman, ``The
  pascal visual object classes (voc) challenge,'' \emph{IJCV}, vol.~88, no.~2,
  pp. 303--338, 2010.

\bibitem{lecun2015deep}
Y.~LeCun, Y.~Bengio, and G.~Hinton, ``Deep learning,'' \emph{Nature}, vol. 521,
  no. 7553, pp. 436--444, 2015.

\bibitem{simonyan2014very}
K.~Simonyan and A.~Zisserman, ``Very deep convolutional networks for
  large-scale image recognition,'' \emph{ICLR}, 2015.

\bibitem{szegedy2015going}
C.~Szegedy, W.~Liu, Y.~Jia, P.~Sermanet, S.~Reed, D.~Anguelov, D.~Erhan,
  V.~Vanhoucke, and A.~Rabinovich, ``Going deeper with convolutions,'' in
  \emph{CVPR}, 2015, pp. 1--9.

\bibitem{he2016deep}
K.~He, X.~Zhang, S.~Ren, and J.~Sun, ``Deep residual learning for image
  recognition,'' in \emph{CVPR}, 2016, pp. 770--778.

\bibitem{girshick2014rich}
R.~Girshick, J.~Donahue, T.~Darrell, and J.~Malik, ``Rich feature hierarchies
  for accurate object detection and semantic segmentation,'' in \emph{CVPR},
  2014, pp. 580--587.

\bibitem{yosinski2014transferable}
J.~Yosinski, J.~Clune, Y.~Bengio, and H.~Lipson, ``How transferable are
  features in deep neural networks?'' in \emph{NeurIPS}, 2014, pp. 3320--3328.

\bibitem{cimpoi2015deep}
M.~Cimpoi, S.~Maji, and A.~Vedaldi, ``Deep filter banks for texture recognition
  and segmentation,'' in \emph{CVPR}, 2015, pp. 3828--3836.

\bibitem{liu2014encoding}
L.~Liu, C.~Shen, L.~Wang, A.~Van Den~Hengel, and C.~Wang, ``Encoding high
  dimensional local features by sparse coding based fisher vectors,'' in
  \emph{NeurIPS}, 2014, pp. 1143--1151.

\bibitem{bolei2015object}
Z.~Bolei, A.~Khosla, A.~Lapedriza, A.~Oliva, and A.~Torralba, ``Object
  detectors emerge in deep scene {CNNs},'' \emph{ICLR}, 2015.

\bibitem{liu2019bow}
L.~Liu, J.~Chen, P.~Fieguth, G.~Zhao, R.~Chellappa, and M.~Pietik{\"a}inen,
  ``From {BoW} to {CNN}: {T}wo decades of texture representation for texture
  classification,'' \emph{IJCV}, vol. 127, no.~1, pp. 74--109, 2019.

\bibitem{sutskever2013importance}
I.~Sutskever, J.~Martens, G.~Dahl, and G.~Hinton, ``On the importance of
  initialization and momentum in deep learning,'' in \emph{ICML}, 2013, pp.
  1139--1147.

\bibitem{liu2014learning}
B.~Liu, J.~Liu, J.~Wang, and H.~Lu, ``Learning a representative and
  discriminative part model with deep convolutional features for scene
  recognition,'' in \emph{ACCV}, 2014, pp. 643--658.

\bibitem{chatfield2014return}
K.~Chatfield, K.~Simonyan, A.~Vedaldi, and A.~Zisserman, ``Return of the devil
  in the details: {D}elving deep into convolutional nets,''
  \emph{arXiv:1405.3531}, 2014.

\bibitem{song2018learning}
X.~Song, S.~Jiang, L.~Herranz, and C.~Chen, ``Learning effective {RGB-D}
  representations for scene recognition,'' \emph{IEEE TIP}, vol.~28, no.~2, pp.
  980--993, 2018.

\bibitem{lin2013network}
M.~Lin, Q.~Chen, and S.~Yan, ``Network in network,'' \emph{ICLR}, 2014.

\bibitem{sun2013learning}
J.~Sun and J.~Ponce, ``Learning discriminative part detectors for image
  classification and cosegmentation,'' in \emph{ICCV}, 2013, pp. 3400--3407.

\bibitem{shih2014learning}
K.~J. Shih, I.~Endres, and D.~Hoiem, ``Learning discriminative collections of
  part detectors for object recognition,'' \emph{IEEE TPAMI}, vol.~37, no.~8,
  pp. 1571--1584, 2014.

\bibitem{sun2015deeply}
Y.~Sun, X.~Wang, and X.~Tang, ``Deeply learned face representations are sparse,
  selective, and robust,'' in \emph{CVPR}, 2015, pp. 2892--2900.

\bibitem{donoho2009message}
D.~L. Donoho, A.~Maleki, and A.~Montanari, ``Message-passing algorithms for
  compressed sensing,'' \emph{PNAS}, vol. 106, no.~45, pp. 18\,914--18\,919,
  2009.

\bibitem{hochreiter1997long}
S.~Hochreiter and J.~Schmidhuber, ``Long short-term memory,'' \emph{Neural
  Computation}, vol.~9, no.~8, pp. 1735--1780, 1997.

\bibitem{li2019mapnet}
Y.~Li, Z.~Zhang, Y.~Cheng, L.~Wang, and T.~Tan, ``{MAPNet}: {M}ulti-modal
  attentive pooling network for {RGB-D} indoor scene classification,''
  \emph{Pattern Recognition}, vol.~90, pp. 436--449, 2019.

\bibitem{mirowski2016learning}
P.~Mirowski, R.~Pascanu, F.~Viola, H.~Soyer, A.~J. Ballard, A.~Banino,
  M.~Denil, R.~Goroshin, L.~Sifre, K.~Kavukcuoglu \emph{et~al.}, ``Learning to
  navigate in complex environments,'' \emph{ICLR}, 2016.

\bibitem{sejnowski2020unreasonable}
T.~J. Sejnowski, ``The unreasonable effectiveness of deep learning in
  artificial intelligence,'' \emph{PNAS}, 2020.

\bibitem{xie2016interactive}
L.~Xie, L.~Zheng, J.~Wang, A.~L. Yuille, and Q.~Tian, ``Interactive:
  {I}nter-layer activeness propagation,'' in \emph{CVPR}, 2016, pp. 270--279.

\bibitem{xie2017towards}
L.~Xie, J.~Wang, W.~Lin, B.~Zhang, and Q.~Tian, ``Towards reversal-invariant
  image representation,'' \emph{IJCV}, vol. 123, no.~2, pp. 226--250, 2017.

\bibitem{rezanejad2019scene}
M.~Rezanejad, G.~Downs, J.~Wilder, D.~B. Walther, A.~Jepson, S.~Dickinson, and
  K.~Siddiqi, ``Scene categorization from contours: {M}edial axis based
  salience measures,'' in \emph{CVPR}, 2019, pp. 4116--4124.

\bibitem{gao2015deep}
B.-B. Gao, X.-S. Wei, J.~Wu, and W.~Lin, ``Deep spatial pyramid: {T}he devil is
  once again in the details,'' \emph{arXiv:1504.05277}, 2015.

\bibitem{xiong2019rgb}
Z.~Xiong, Y.~Yuan, and Q.~Wang, ``{RGB-D} scene recognition via spatial-related
  multi-modal feature learning,'' \emph{IEEE Access}, vol.~7, pp.
  106\,739--106\,747, 2019.

\bibitem{sanchez2013image}
J.~S{\'a}nchez, F.~Perronnin, T.~Mensink, and J.~Verbeek, ``Image
  classification with the {F}isher vector: {T}heory and practice,''
  \emph{IJCV}, vol. 105, no.~3, pp. 222--245, 2013.

\bibitem{jegou2010aggregating}
H.~J{\'e}gou, M.~Douze, C.~Schmid, and P.~P{\'e}rez, ``Aggregating local
  descriptors into a compact image representation,'' in \emph{CVPR}, 2010, pp.
  3304--3311.

\bibitem{csurka2004visual}
G.~Csurka, C.~Dance, L.~Fan, J.~Willamowski, and C.~Bray, ``Visual
  categorization with bags of keypoints,'' in \emph{ECCVW}, 2004, pp. 1--2.

\bibitem{liu2020deep}
L.~Liu, W.~Ouyang, X.~Wang, P.~Fieguth, J.~Chen, X.~Liu, and
  M.~Pietik{\"a}inen, ``Deep learning for generic object detection: {A}
  survey,'' \emph{IJCV}, vol. 128, no.~2, pp. 261--318, 2020.

\bibitem{uijlings2013selective}
J.~R. Uijlings, K.~E. Van De~Sande, T.~Gevers, and A.~W. Smeulders, ``Selective
  search for object recognition,'' \emph{IJCV}, vol. 104, no.~2, pp. 154--171,
  2013.

\bibitem{singh2012unsupervised}
S.~Singh, A.~Gupta, and A.~A. Efros, ``Unsupervised discovery of mid-level
  discriminative patches,'' in \emph{ECCV}, 2012, pp. 73--86.

\bibitem{arbelaez2014multiscale}
P.~Arbel{\'a}ez, J.~Pont-Tuset, J.~T. Barron, F.~Marques, and J.~Malik,
  ``Multiscale combinatorial grouping,'' in \emph{CVPR}, 2014, pp. 328--335.

\bibitem{liu2016ssd}
W.~Liu, D.~Anguelov, D.~Erhan, C.~Szegedy, S.~Reed, C.-Y. Fu, and A.~C. Berg,
  ``{SSD}: {S}ingle shot multibox detector,'' in \emph{ECCV}, 2016, pp. 21--37.

\bibitem{redmon2017yolo9000}
J.~Redmon and A.~Farhadi, ``Yolo9000: {B}etter, faster, stronger,'' in
  \emph{CVPR}, 2017, pp. 7263--7271.

\bibitem{redmon2016you}
J.~Redmon, S.~Divvala, R.~Girshick, and A.~Farhadi, ``You only look once:
  {U}nified, real-time object detection,'' in \emph{CVPR}, 2016, pp. 779--788.

\bibitem{lee2015deeply}
C.-Y. Lee, S.~Xie, P.~Gallagher, Z.~Zhang, and Z.~Tu, ``Deeply-supervised
  nets,'' in \emph{AISTATS}, 2015, pp. 562--570.

\bibitem{lin2017feature}
T.-Y. Lin, P.~Doll{\'a}r, R.~Girshick, K.~He, B.~Hariharan, and S.~Belongie,
  ``Feature pyramid networks for object detection,'' in \emph{CVPR}, 2017, pp.
  2117--2125.

\bibitem{snoek2005early}
C.~G. Snoek, M.~Worring, and A.~W. Smeulders, ``Early versus late fusion in
  semantic video analysis,'' in \emph{Proceedings of the 13th annual ACM
  international conference on Multimedia}, 2005, pp. 399--402.

\bibitem{gunes2005affect}
H.~Gunes and M.~Piccardi, ``Affect recognition from face and body: {E}arly
  fusion vs. late fusion,'' in \emph{International conference on systems, man
  and cybernetics}, vol.~4, 2005, pp. 3437--3443.

\bibitem{dong2014performance}
Y.~Dong, S.~Gao, K.~Tao, J.~Liu, and H.~Wang, ``Performance evaluation of early
  and late fusion methods for generic semantics indexing,'' \emph{Pattern
  Analysis and Applications}, vol.~17, no.~1, pp. 37--50, 2014.

\bibitem{li2020deep}
J.~Li, D.~Lin, Y.~Wang, G.~Xu, Y.~Zhang, C.~Ding, and Y.~Zhou, ``Deep
  discriminative representation learning with attention map for scene
  classification,'' \emph{Remote Sensing}, vol.~12, no.~9, p. 1366, 2020.

\bibitem{zhang2015scene}
F.~Zhang, B.~Du, and L.~Zhang, ``Scene classification via a gradient boosting
  random convolutional network framework,'' \emph{IEEE TGRS}, vol.~54, no.~3,
  pp. 1793--1802, 2015.

\bibitem{wang2015object}
L.~Wang, Z.~Wang, W.~Du, and Y.~Qiao, ``Object-scene convolutional neural
  networks for event recognition in images,'' in \emph{CVPRW}, 2015, pp.
  30--35.

\bibitem{xia2019ws}
S.~Xia, J.~Zeng, L.~Leng, and X.~Fu, ``{WS-AM}: {W}eakly supervised attention
  map for scene recognition,'' \emph{Electronics}, vol.~8, no.~10, p. 1072,
  2019.

\bibitem{ioffe2015batch}
S.~Ioffe and C.~Szegedy, ``Batch normalization: {A}ccelerating deep network
  training by reducing internal covariate shift,'' \emph{ICML}, 2015.

\bibitem{jin2018hierarchy}
H.~Jin~Kim and J.-M. Frahm, ``Hierarchy of alternating specialists for scene
  recognition,'' in \emph{ECCV}, 2018, pp. 451--467.

\bibitem{perronnin2007fisher}
F.~Perronnin and C.~Dance, ``Fisher kernels on visual vocabularies for image
  categorization,'' in \emph{CVPR}, 2007, pp. 1--8.

\bibitem{kwitt2012scene}
R.~Kwitt, N.~Vasconcelos, and N.~Rasiwasia, ``Scene recognition on the semantic
  manifold,'' in \emph{ECCV}, 2012, pp. 359--372.

\bibitem{ghahramani1996algorithm}
Z.~Ghahramani, G.~E. Hinton \emph{et~al.}, ``The em algorithm for mixtures of
  factor analyzers,'' University of Toronto, Tech. Rep., 1996.

\bibitem{verbeek2006learning}
J.~Verbeek, ``Learning nonlinear image manifolds by global alignment of local
  linear models,'' \emph{IEEE TPAMI}, vol.~28, no.~8, pp. 1236--1250, 2006.

\bibitem{wei2017correlated}
P.~Wei, F.~Qin, F.~Wan, Y.~Zhu, J.~Jiao, and Q.~Ye, ``Correlated topic vector
  for scene classification,'' \emph{IEEE TIP}, vol.~26, no.~7, pp. 3221--3234,
  2017.

\bibitem{selvaraju2017grad}
R.~R. Selvaraju, M.~Cogswell, A.~Das, R.~Vedantam, D.~Parikh, and D.~Batra,
  ``Grad-{CAM}: {V}isual explanations from deep networks via gradient-based
  localization,'' in \emph{ICCV}, 2017, pp. 618--626.

\bibitem{zhang2018top}
J.~Zhang, S.~A. Bargal, Z.~Lin, J.~Brandt, X.~Shen, and S.~Sclaroff, ``Top-down
  neural attention by excitation backprop,'' \emph{IJCV}, vol. 126, no.~10, pp.
  1084--1102, 2018.

\bibitem{seong2020fosnet}
H.~Seong, J.~Hyun, and E.~Kim, ``Fosnet: {A}n end-to-end trainable deep neural
  network for scene recognition,'' \emph{IEEE Access}, vol.~8, pp.
  82\,066--82\,077, 2020.

\bibitem{joseph2019joint}
T.~Joseph, K.~G. Derpanis, and F.~Z. Qureshi, ``Joint spatial and layer
  attention for convolutional networks,'' \emph{arXiv:1901.05376}, 2019.

\bibitem{niu2012context}
Z.~Niu, G.~Hua, X.~Gao, and Q.~Tian, ``Context aware topic model for scene
  recognition,'' in \emph{CVPR}, 2012, pp. 2743--2750.

\bibitem{zuo2015convolutional}
Z.~Zuo, B.~Shuai, G.~Wang, X.~Liu, X.~Wang, B.~Wang, and Y.~Chen,
  ``Convolutional recurrent neural networks: {L}earning spatial dependencies
  for image representation,'' in \emph{CVPRW}, 2015, pp. 18--26.

\bibitem{wang2008spatial}
X.~Wang and E.~Grimson, ``Spatial latent dirichlet allocation,'' in
  \emph{NeurIPS}, 2008, pp. 1577--1584.

\bibitem{elman1990finding}
J.~L. Elman, ``Finding structure in time,'' \emph{Cognitive science}, vol.~14,
  no.~2, pp. 179--211, 1990.

\bibitem{cross1983markov}
G.~R. Cross and A.~K. Jain, ``Markov random field texture models,'' \emph{IEEE
  TPAMI}, no.~1, pp. 25--39, 1983.

\bibitem{li2009markov}
S.~Z. Li, \emph{Markov random field modeling in image analysis}.\hskip 1em plus
  0.5em minus 0.4em\relax Springer, 2009.

\bibitem{rasiwasia2012holistic}
N.~Rasiwasia and N.~Vasconcelos, ``Holistic context models for visual
  recognition,'' \emph{IEEE TPAMI}, vol.~34, no.~5, pp. 902--917, 2012.

\bibitem{bruna2013spectral}
J.~Bruna, W.~Zaremba, A.~Szlam, and Y.~LeCun, ``Spectral networks and locally
  connected networks on graphs,'' \emph{arXiv:1312.6203}, 2013.

\bibitem{kipf2016semi}
T.~N. Kipf and M.~Welling, ``Semi-supervised classification with graph
  convolutional networks,'' \emph{ICLR}, 2016.

\bibitem{zhou2018graph}
J.~Zhou, G.~Cui, Z.~Zhang, C.~Yang, Z.~Liu, L.~Wang, C.~Li, and M.~Sun, ``Graph
  neural networks: {A} review of methods and applications,''
  \emph{arXiv:1812.08434}, 2018.

\bibitem{yuan2019acm}
Y.~Yuan, Z.~Xiong, and Q.~Wang, ``Acm: {A}daptive cross-modal graph
  convolutional neural networks for {RGB-D} scene recognition,'' in
  \emph{AAAI}, vol.~33, 2019, pp. 9176--9184.

\bibitem{song2019image}
X.~Song, S.~Jiang, B.~Wang, C.~Chen, and G.~Chen, ``Image representations with
  spatial object-to-object relations for {RGB-D} scene recognition,''
  \emph{IEEE TIP}, vol.~29, pp. 525--537, 2019.

\bibitem{javed2017object}
S.~A. Javed and A.~K. Nelakanti, ``Object-level context modeling for scene
  classification with context-{CNN},'' \emph{arXiv:1705.04358}, 2017.

\bibitem{song2017multi}
X.~Song, S.~Jiang, and L.~Herranz, ``Multi-scale multi-feature context modeling
  for scene recognition in the semantic manifold,'' \emph{IEEE TIP}, vol.~26,
  no.~6, pp. 2721--2735, 2017.

\bibitem{kipf2017semi}
T.~N. Kipf and M.~Welling, ``Semi-supervised classification with graph
  convolutional networks,'' \emph{ICLR}, 2017.

\bibitem{zuo2014learning}
Z.~Zuo, G.~Wang, B.~Shuai, L.~Zhao, Q.~Yang, and X.~Jiang, ``Learning
  discriminative and shareable features for scene classification,'' in
  \emph{ECCV}, 2014, pp. 552--568.

\bibitem{jiang2019deep}
S.~Jiang, G.~Chen, X.~Song, and L.~Liu, ``Deep patch representations with
  shared codebook for scene classification,'' \emph{ACM TOMM}, vol.~15, no.~1s,
  pp. 1--17, 2019.

\bibitem{boser1992training}
B.~E. Boser, I.~M. Guyon, and V.~N. Vapnik, ``A training algorithm for optimal
  margin classifiers,'' in \emph{Annual workshop on Computational learning
  theory}, 1992, pp. 144--152.

\bibitem{li2018df}
Y.~Li, J.~Zhang, Y.~Cheng, K.~Huang, and T.~Tan, ``Df2net: {D}iscriminative
  feature learning and fusion network for {RGB-D} indoor scene
  classification,'' in \emph{AAAI}, 2018.

\bibitem{van2009visual}
J.~C. Van~Gemert, C.~J. Veenman, A.~W. Smeulders, and J.-M. Geusebroek,
  ``Visual word ambiguity,'' \emph{IEEE TPAMI}, vol.~32, no.~7, pp. 1271--1283,
  2009.

\bibitem{scholkopf2001estimating}
B.~Sch{\"o}lkopf, J.~C. Platt, J.~Shawe-Taylor, A.~J. Smola, and R.~C.
  Williamson, ``Estimating the support of a high-dimensional distribution,''
  \emph{Neural computation}, vol.~13, no.~7, pp. 1443--1471, 2001.

\bibitem{durand2015mantra}
T.~Durand, N.~Thome, and M.~Cord, ``Mantra: {M}inimum maximum latent structural
  svm for image classification and ranking,'' in \emph{ICCV}, 2015, pp.
  2713--2721.

\bibitem{socher2012convolutional}
R.~Socher, B.~Huval, B.~Bath, C.~D. Manning, and A.~Y. Ng,
  ``Convolutional-recursive deep learning for 3d object classification,'' in
  \emph{NeurIPS}, 2012, pp. 656--664.

\bibitem{wang2015mmss}
A.~Wang, J.~Cai, J.~Lu, and T.-J. Cham, ``{MMSS}: {M}ulti-modal sharable and
  specific feature learning for {RGB-D} object recognition,'' in \emph{ICCV},
  2015, pp. 1125--1133.

\bibitem{cheng2016semi}
Y.~Cheng, X.~Zhao, R.~Cai, Z.~Li, K.~Huang, Y.~Rui \emph{et~al.},
  ``Semi-supervised multimodal deep learning for {RGB-D} object recognition,''
  \emph{IJCAI}, pp. 3346--3351, 2016.

\bibitem{wang2018detecting}
Q.~Wang, M.~Chen, F.~Nie, and X.~Li, ``Detecting coherent groups in crowd
  scenes by multiview clustering,'' \emph{IEEE TPAMI}, vol.~42, no.~1, pp.
  46--58, 2018.

\bibitem{zheng2017indoor}
Y.~Zheng and X.~Gao, ``Indoor scene recognition via multi-task metric
  multi-kernel learning from {RGB-D} images,'' \emph{Multimedia Tools and
  Applications}, vol.~76, no.~3, pp. 4427--4443, 2017.

\bibitem{thompson2005canonical}
B.~Thompson, ``Canonical correlation analysis,'' \emph{Encyclopedia of
  statistics in behavioral science}, 2005.

\bibitem{andrew2013deep}
G.~Andrew, R.~Arora, J.~Bilmes, and K.~Livescu, ``Deep canonical correlation
  analysis,'' in \emph{ICML}, 2013, pp. 1247--1255.

\bibitem{wu2016representing}
J.~Wu, B.-B. Gao, and G.~Liu, ``Representing sets of instances for visual
  recognition,'' in \emph{AAAI}, 2016, pp. 2237--2243.

\bibitem{bappy2016online}
J.~H. Bappy, S.~Paul, and A.~K. Roy-Chowdhury, ``Online adaptation for joint
  scene and object classification,'' in \emph{ECCV}, 2016, pp. 227--243.

\bibitem{zhao2018volcano}
Z.~Zhao and M.~Larson, ``From volcano to toyshop: {A}daptive discriminative
  region discovery for scene recognition,'' in \emph{ACM MM}, 2018, pp.
  1760--1768.

\bibitem{koskela2014convolutional}
M.~Koskela and J.~Laaksonen, ``Convolutional network features for scene
  recognition,'' in \emph{ACM MM}, 2014, pp. 1169--1172.

\bibitem{nascimento2017robust}
G.~Nascimento, C.~Laranjeira, V.~Braz, A.~Lacerda, and E.~R. Nascimento, ``A
  robust indoor scene recognition method based on sparse representation,'' in
  \emph{CIARP}, 2017, pp. 408--415.

\bibitem{liao2016understand}
Y.~Liao, S.~Kodagoda, Y.~Wang, L.~Shi, and Y.~Liu, ``Understand scene
  categories by objects: {A} semantic regularized scene classifier using
  convolutional neural networks,'' in \emph{CRA}, 2016, pp. 2318--2325.

\bibitem{cai2019rgb}
Z.~Cai and L.~Shao, ``{RGB-D} scene classification via multi-modal feature
  learning,'' \emph{Cognitive Computation}, vol.~11, no.~6, pp. 825--840, 2019.

\bibitem{ayub2019centroid}
A.~Ayub and A.~Wagner, ``Cbcl: {B}rain-inspired model for {RGB-D} indoor scene
  classification,'' \emph{arXiv:1911.00155}, 2019.

\bibitem{song2017depth}
X.~Song, L.~Herranz, and S.~Jiang, ``Depth {CNNs} for {RGB-D} scene
  recognition: {L}earning from scratch better than transferring from
  {RGB-CNNs},'' in \emph{AAAI}, vol.~31, no.~1, 2017.

\bibitem{najafabadi2015deep}
M.~M. Najafabadi, F.~Villanustre, T.~M. Khoshgoftaar, N.~Seliya \emph{et~al.},
  ``Deep learning applications and challenges in big data analytics,''
  \emph{Journal of Big Data}, vol.~2, no.~1, p.~1, 2015.

\bibitem{zoph2017neural}
B.~Zoph and Q.~V. Le, ``Neural architecture search with reinforcement
  learning,'' \emph{ICLR}, 2017.

\bibitem{elsken2019neural}
T.~Elsken, J.~H. Metzen, F.~Hutter \emph{et~al.}, ``Neural architecture search:
  {A} survey.'' \emph{JMLR}, vol.~20, no.~55, pp. 1--21, 2019.

\bibitem{xiao2016sun}
J.~Xiao, K.~A. Ehinger, J.~Hays, A.~Torralba, and A.~Oliva, ``{SUN} database:
  {E}xploring a large collection of scene categories,'' \emph{IJCV}, vol. 119,
  no.~1, pp. 3--22, 2016.

\bibitem{chapelle2009semi}
O.~Chapelle, B.~Scholkopf, and A.~Zien, ``Semi-supervised learning,''
  \emph{IEEE TNN}, vol.~20, no.~3, pp. 542--542, 2009.

\bibitem{barlow1989unsupervised}
H.~B. Barlow, ``Unsupervised learning,'' \emph{Neural computation}, vol.~1,
  no.~3, pp. 295--311, 1989.

\bibitem{kolesnikov2019revisiting}
A.~Kolesnikov, X.~Zhai, and L.~Beyer, ``Revisiting self-supervised visual
  representation learning,'' in \emph{CVPR}, 2019, pp. 1920--1929.

\bibitem{wang2016learning}
Y.-X. Wang and M.~Hebert, ``Learning to learn: Model regression networks for
  easy small sample learning,'' in \emph{ECCV}, 2016, pp. 616--634.

\bibitem{fei2006one}
L.~Fei-Fei, R.~Fergus, and P.~Perona, ``One-shot learning of object
  categories,'' \emph{IEEE TPAMI}, vol.~28, no.~4, pp. 594--611, 2006.

\bibitem{lake2015human}
B.~M. Lake, R.~Salakhutdinov, and J.~B. Tenenbaum, ``Human-level concept
  learning through probabilistic program induction,'' \emph{Science}, vol. 350,
  no. 6266, pp. 1332--1338, 2015.

\bibitem{long2015learning}
M.~Long, Y.~Cao, J.~Wang, and M.~Jordan, ``Learning transferable features with
  deep adaptation networks,'' in \emph{ICML}, 2015, pp. 97--105.

\bibitem{wang2018deep}
M.~Wang and W.~Deng, ``Deep visual domain adaptation: {A} survey,''
  \emph{Neurocomputing}, vol. 312, pp. 135--153, 2018.

\bibitem{peng2018zero}
K.-C. Peng, Z.~Wu, and J.~Ernst, ``Zero-shot deep domain adaptation,'' in
  \emph{ECCV}, 2018, pp. 764--781.

\bibitem{li2018domain}
H.~Li, S.~Jialin~Pan, S.~Wang, and A.~C. Kot, ``Domain generalization with
  adversarial feature learning,'' in \emph{CVPR}, 2018, pp. 5400--5409.

\bibitem{zhang2020towards}
X.-Y. Zhang, C.-L. Liu, and C.~Y. Suen, ``Towards robust pattern recognition:
  {A} review,'' \emph{Proceedings of the IEEE}, vol. 108, no.~6, pp. 894--922,
  2020.

\bibitem{shaham2018understanding}
U.~Shaham, Y.~Yamada, and S.~Negahban, ``Understanding adversarial training:
  {I}ncreasing local stability of supervised models through robust
  optimization,'' \emph{Neurocomputing}, vol. 307, pp. 195--204, 2018.

\bibitem{qin2019adversarial}
C.~Qin, J.~Martens, S.~Gowal, D.~Krishnan, K.~Dvijotham, A.~Fawzi, S.~De,
  R.~Stanforth, and P.~Kohli, ``Adversarial robustness through local
  linearization,'' in \emph{NeurIPS}, 2019, pp. 13\,847--13\,856.

\bibitem{shafahi2019adversarial}
A.~Shafahi, M.~Najibi, M.~A. Ghiasi, Z.~Xu, J.~Dickerson, C.~Studer, L.~S.
  Davis, G.~Taylor, and T.~Goldstein, ``Adversarial training for free!'' in
  \emph{NeurIPS}, 2019, pp. 3358--3369.

\bibitem{ganin2016domain}
Y.~Ganin, E.~Ustinova, H.~Ajakan, P.~Germain, H.~Larochelle, F.~Laviolette,
  M.~Marchand, and V.~Lempitsky, ``Domain-adversarial training of neural
  networks,'' \emph{JMLR}, vol.~17, no.~1, pp. 2096--2030, 2016.

\bibitem{athalye2018synthesizing}
A.~Athalye, L.~Engstrom, A.~Ilyas, and K.~Kwok, ``Synthesizing robust
  adversarial examples,'' in \emph{ICML}, 2018, pp. 284--293.

\bibitem{dargazany2018wearabledl}
A.~R. Dargazany, P.~Stegagno, and K.~Mankodiya, ``{WearableDL: Wearable
  internet-of-things and deep learning for big data analytics—concept,
  literature, and future},'' \emph{Mobile Information Systems}, vol. 2018,
  2018.

\bibitem{shen2016relay}
L.~Shen, Z.~Lin, and Q.~Huang, ``Relay backpropagation for effective learning
  of deep convolutional neural networks,'' in \emph{ECCV}, 2016, pp. 467--482.

\bibitem{thabtah2020data}
F.~Thabtah, S.~Hammoud, F.~Kamalov, and A.~Gonsalves, ``Data imbalance in
  classification: {E}xperimental evaluation,'' \emph{Information Sciences},
  vol. 513, pp. 429--441, 2020.

\bibitem{buda2018systematic}
M.~Buda, A.~Maki, and M.~A. Mazurowski, ``A systematic study of the class
  imbalance problem in convolutional neural networks,'' \emph{Neural Networks},
  vol. 106, pp. 249--259, 2018.

\bibitem{johnson2019survey}
J.~M. Johnson and T.~M. Khoshgoftaar, ``Survey on deep learning with class
  imbalance,'' \emph{Journal of Big Data}, vol.~6, no.~1, p.~27, 2019.

\bibitem{guo2016deep}
Y.~Guo, Y.~Liu, A.~Oerlemans, S.~Lao, S.~Wu, and M.~S. Lew, ``Deep learning for
  visual understanding: {A} review,'' \emph{Neurocomputing}, vol. 187, pp.
  27--48, 2016.

\bibitem{chen2018lifelong}
Z.~Chen and B.~Liu, ``Lifelong machine learning,'' \emph{Synthesis Lectures on
  Artificial Intelligence and Machine Learning}, vol.~12, no.~3, pp. 1--207,
  2018.

\bibitem{thrun1995lifelong}
S.~Thrun and T.~M. Mitchell, ``Lifelong robot learning,'' \emph{Robotics and
  autonomous systems}, vol.~15, no. 1-2, pp. 25--46, 1995.

\bibitem{hassabis2017neuroscience}
D.~Hassabis, D.~Kumaran, C.~Summerfield, and M.~Botvinick,
  ``Neuroscience-inspired artificial intelligence,'' \emph{Neuron}, vol.~95,
  no.~2, pp. 245--258, 2017.

\bibitem{zhang2007ml}
M.-L. Zhang and Z.-H. Zhou, ``Ml-knn: A lazy learning approach to multi-label
  learning,'' \emph{Pattern Recognition}, vol.~40, no.~7, pp. 2038--2048, 2007.

\bibitem{feichtenhofer2017temporal}
C.~Feichtenhofer, A.~Pinz, and R.~P. Wildes, ``Temporal residual networks for
  dynamic scene recognition,'' in \emph{CVPR}, 2017, pp. 4728--4737.

\bibitem{dai2017scannet}
A.~Dai, A.~X. Chang, M.~Savva, M.~Halber, T.~Funkhouser, and M.~Nie{\ss}ner,
  ``Scannet: {R}ichly-annotated 3d reconstructions of indoor scenes,'' in
  \emph{CVPR}, 2017, pp. 5828--5839.

\bibitem{tran2015learning}
D.~Tran, L.~Bourdev, R.~Fergus, L.~Torresani, and M.~Paluri, ``Learning
  spatiotemporal features with 3d convolutional networks,'' in \emph{ICCV},
  2015, pp. 4489--4497.

\bibitem{qi2017pointnet}
C.~R. Qi, H.~Su, K.~Mo, and L.~J. Guibas, ``Pointnet: {D}eep learning on point
  sets for {3D} classification and segmentation,'' in \emph{CVPR}, 2018, pp.
  652--660.

\bibitem{lowe1999object}
D.~G. Lowe, ``Object recognition from local scale-invariant features,'' in
  \emph{IJCV}, vol.~2, 1999, pp. 1150--1157.

\bibitem{oliva2001modeling}
A.~Oliva and A.~Torralba, ``Modeling the shape of the scene: {A} holistic
  representation of the spatial envelope,'' \emph{IJCV}, vol.~42, no.~3, pp.
  145--175, 2001.

\bibitem{dalal2005histograms}
N.~Dalal and B.~Triggs, ``Histograms of oriented gradients for human
  detection,'' in \emph{CVPR}, vol.~1, 2005, pp. 886--893.

\bibitem{wu2010centrist}
J.~Wu and J.~M. Rehg, ``{CENTRIST}: {A} visual descriptor for scene
  categorization,'' \emph{IEEE TPAMI}, vol.~33, no.~8, pp. 1489--1501, 2010.

\bibitem{sivic2003video}
J.~Sivic and A.~Zisserman, ``Video google: {A} text retrieval approach to
  object matching in videos,'' in \emph{ICCV}, vol.~2, 2003, p. 1470–1477.

\bibitem{li2010objects}
L.-J. Li, H.~Su, Y.~Lim, and L.~Fei-Fei, ``Objects as attributes for scene
  classification,'' in \emph{ECCV}, 2010, pp. 57--69.

\bibitem{juneja2013blocks}
M.~Juneja, A.~Vedaldi, C.~Jawahar, and A.~Zisserman, ``Blocks that shout:
  {D}istinctive parts for scene classification,'' in \emph{CVPR}, 2013, pp.
  923--930.

\bibitem{sharif2014cnn}
A.~Sharif~Razavian, H.~Azizpour, J.~Sullivan, and S.~Carlsson, ``{CNN} features
  off-the-shelf: {A}n astounding baseline for recognition,'' in \emph{CVPRW},
  2014, pp. 806--813.

\bibitem{vasconcelos2000probabilistic}
N.~Vasconcelos and A.~Lippman, ``A probabilistic architecture for content-based
  image retrieval,'' in \emph{CVPR}, 2000, pp. 216--221.

\bibitem{wallraven2003recognition}
C.~Wallraven, B.~Caputo, and A.~Graf, ``Recognition with local features: {T}he
  kernel recipe,'' in \emph{ICCV}, 2003, pp. 257--264.

\bibitem{lowe2004distinctive}
D.~G. Lowe, ``Distinctive image features from scale-invariant keypoints,''
  \emph{IJCV}, vol.~60, no.~2, pp. 91--110, 2004.

\bibitem{oliva2006building}
A.~Oliva and A.~Torralba, ``Building the gist of a scene: {T}he role of global
  image features in recognition,'' \emph{Progress in brain research}, vol. 155,
  pp. 23--36, 2006.

\bibitem{ojala2002multiresolution}
T.~Ojala, M.~Pietikainen, and T.~Maenpaa, ``Multiresolution gray-scale and
  rotation invariant texture classification with local binary patterns,''
  \emph{IEEE TPAMI}, vol.~24, no.~7, pp. 971--987, 2002.

\bibitem{felzenszwalb2008discriminatively}
P.~Felzenszwalb, D.~McAllester, and D.~Ramanan, ``A discriminatively trained,
  multiscale, deformable part model,'' in \emph{CVPR}, 2008, pp. 1--8.

\bibitem{felzenszwalb2009object}
P.~F. Felzenszwalb, R.~B. Girshick, D.~McAllester, and D.~Ramanan, ``Object
  detection with discriminatively trained part-based models,'' \emph{IEEE
  TPAMI}, vol.~32, no.~9, pp. 1627--1645, 2009.

\bibitem{pandey2011scene}
M.~Pandey and S.~Lazebnik, ``Scene recognition and weakly supervised object
  localization with deformable part-based models,'' in \emph{ICCV}, 2011, pp.
  1307--1314.

\bibitem{blei2003latent}
D.~M. Blei, A.~Y. Ng, and M.~I. Jordan, ``Latent dirichlet allocation,''
  \emph{JMLR}, vol.~3, no.~1, pp. 993--1022, 2003.

\bibitem{grauman2005pyramid}
K.~Grauman and T.~Darrell, ``The pyramid match kernel: {D}iscriminative
  classification with sets of image features,'' in \emph{ICCV}, vol.~2, 2005,
  pp. 1458--1465.

\bibitem{zhou2013scene}
L.~Zhou, Z.~Zhou, and D.~Hu, ``Scene classification using a multi-resolution
  bag-of-features model,'' \emph{Pattern Recognition}, vol.~46, no.~1, pp.
  424--433, 2013.

\bibitem{xie2014orientational}
L.~Xie, J.~Wang, B.~Guo, B.~Zhang, and Q.~Tian, ``Orientational pyramid
  matching for recognizing indoor scenes,'' in \emph{CVPR}, 2014, pp.
  3734--3741.

\bibitem{yang2009linear}
J.~Yang, K.~Yu, Y.~Gong, and T.~Huang, ``Linear spatial pyramid matching using
  sparse coding for image classification,'' in \emph{CVPR}, 2009, pp.
  1794--1801.

\bibitem{gao2010local}
S.~Gao, I.~W.-H. Tsang, L.-T. Chia, and P.~Zhao, ``Local features are not
  lonely--laplacian sparse coding for image classification,'' in \emph{CVPR},
  2010, pp. 3555--3561.

\bibitem{wang2010locality}
J.~Wang, J.~Yang, K.~Yu, F.~Lv, T.~Huang, and Y.~Gong, ``Locality-constrained
  linear coding for image classification,'' in \emph{CVPR}, 2010, pp.
  3360--3367.

\bibitem{wu2009beyond}
J.~Wu and J.~M. Rehg, ``Beyond the euclidean distance: {C}reating effective
  visual codebooks using the histogram intersection kernel,'' in \emph{ICCV},
  2009, pp. 630--637.

\bibitem{qin2010scene}
J.~Qin and N.~H. Yung, ``Scene categorization via contextual visual words,''
  \emph{Pattern Recognition}, vol.~43, no.~5, pp. 1874--1888, 2010.

\bibitem{bo2009efficient}
L.~Bo and C.~Sminchisescu, ``Efficient match kernel between sets of features
  for visual recognition,'' in \emph{NeurIPS}, 2009, pp. 135--143.

\bibitem{wang2013supervised}
P.~Wang, J.~Wang, G.~Zeng, W.~Xu, H.~Zha, and S.~Li, ``Supervised kernel
  descriptors for visual recognition,'' in \emph{CVPR}, 2013, pp. 2858--2865.

\bibitem{li2014object}
L.-J. Li, H.~Su, Y.~Lim, and L.~Fei-Fei, ``Object bank: {A}n object-level image
  representation for high-level visual recognition,'' \emph{IJCV}, vol. 107,
  no.~1, pp. 20--39, 2014.

\bibitem{zhang2014learning}
L.~Zhang, X.~Zhen, and L.~Shao, ``Learning object-to-class kernels for scene
  classification,'' \emph{IEEE TIP}, vol.~23, no.~8, pp. 3241--3253, 2014.

\bibitem{sadeghi2012latent}
F.~Sadeghi and M.~F. Tappen, ``Latent pyramidal regions for recognizing
  scenes,'' in \emph{ECCV}, 2012, pp. 228--241.

\bibitem{yu2013pairwise}
J.~Yu, D.~Tao, Y.~Rui, and J.~Cheng, ``Pairwise constraints based multiview
  features fusion for scene classification,'' \emph{Pattern Recognition},
  vol.~46, no.~2, pp. 483--496, 2013.

\bibitem{taigman2014deepface}
Y.~Taigman, M.~Yang, M.~Ranzato, and L.~Wolf, ``Deepface: {C}losing the gap to
  human-level performance in face verification,'' in \emph{CVPR}, 2014, pp.
  1701--1708.

\bibitem{guo2020deep}
Y.~Guo, H.~Wang, Q.~Hu, H.~Liu, L.~Liu, and M.~Bennamoun, ``Deep learning for
  3d point clouds: {A} survey,'' \emph{IEEE TPAMI}, 2020.

\bibitem{hinton2012deep}
G.~Hinton, L.~Deng, D.~Yu, G.~E. Dahl, A.-r. Mohamed \emph{et~al.}, ``Deep
  neural networks for acoustic modeling in speech recognition,'' \emph{IEEE
  Signal Processing Magazine}, vol.~29, no.~6, pp. 82--97, 2012.

\bibitem{chen2015deepdriving}
C.~Chen, A.~Seff, A.~Kornhauser, and J.~Xiao, ``Deepdriving: {L}earning
  affordance for direct perception in autonomous driving,'' in \emph{ICCV},
  2015, pp. 2722--2730.

\bibitem{esteva2017dermatologist}
A.~Esteva, B.~Kuprel, R.~A. Novoa, J.~Ko, S.~M. Swetter, H.~M. Blau, and
  S.~Thrun, ``Dermatologist level classification of skin cancer with deep
  neural networks,'' \emph{Nature}, vol. 542, no. 7639, pp. 115--118, 2017.

\bibitem{mckinney2020international}
S.~M. McKinney, M.~Sieniek, V.~Godbole, J.~Godwin, N.~Antropova \emph{et~al.},
  ``International evaluation of an {AI} system for breast cancer screening,''
  \emph{Nature}, vol. 577, no. 7788, pp. 89--94, 2020.

\bibitem{wu2016google}
Y.~Wu, M.~Schuster, Z.~Chen, Q.~V. Le, M.~Norouzi \emph{et~al.}, ``Google's
  neural machine translation system: {B}ridging the gap between human and
  machine translation,'' \emph{arXiv:1609.08144}, 2016.

\bibitem{silver2016mastering}
D.~Silver, A.~Huang, C.~J. Maddison, A.~Guez, L.~Sifre \emph{et~al.},
  ``Mastering the game of go with deep neural networks and tree search,''
  \emph{Nature}, vol. 529, no. 7587, pp. 484--489, 2016.

\bibitem{silver2017mastering}
D.~Silver, J.~Schrittwieser, K.~Simonyan, I.~Antonoglou, A.~Huang
  \emph{et~al.}, ``Mastering the game of go without human knowledge,''
  \emph{Nature}, vol. 550, no. 7676, pp. 354--359, 2017.

\bibitem{silver2018general}
D.~Silver, T.~Hubert, J.~Schrittwieser, I.~Antonoglou, M.~Lai \emph{et~al.},
  ``A general reinforcement learning algorithm that masters chess, shogi, and
  go through self-play,'' \emph{Science}, vol. 362, no. 6419, pp. 1140--1144,
  2018.

\bibitem{vinyals2019grandmaster}
O.~Vinyals, I.~Babuschkin, W.~M. Czarnecki, M.~Mathieu, A.~Dudzik
  \emph{et~al.}, ``Grandmaster level in starcraft ii using multi-agent
  reinforcement learning,'' \emph{Nature}, vol. 575, no. 7782, pp. 350--354,
  2019.

\bibitem{devries2018deep}
P.~M. DeVries, F.~Vi{\'e}gas, M.~Wattenberg, and B.~J. Meade, ``Deep learning
  of aftershock patterns following large earthquakes,'' \emph{Nature}, vol.
  560, no. 7720, pp. 632--634, 2018.

\bibitem{mit2020breakthrough}
\url{https://www.technologyreview.com/lists/technologies/2020/}.

\bibitem{stokes2020deep}
J.~M. Stokes, K.~Yang, K.~Swanson, W.~Jin, A.~Cubillos-Ruiz, N.~M. Donghia,
  C.~R. MacNair, S.~French, L.~A. Carfrae, Z.~Bloom-Ackerman \emph{et~al.}, ``A
  deep learning approach to antibiotic discovery,'' \emph{Cell}, vol. 180,
  no.~4, pp. 688--702, 2020.

\bibitem{schmidhuber2015deep}
J.~Schmidhuber, ``Deep learning in neural networks: {A}n overview,''
  \emph{Neural networks}, vol.~61, pp. 85--117, 2015.

\bibitem{zeiler2014visualizing}
M.~D. Zeiler and R.~Fergus, ``Visualizing and understanding convolutional
  networks,'' in \emph{ECCV}, 2014, pp. 818--833.

\bibitem{ioffe2017batch}
S.~Ioffe, ``Batch renormalization: {T}owards reducing minibatch dependence in
  batch-normalized models,'' in \emph{NeurIPS}, 2017, pp. 1945--1953.

\bibitem{goodfellow2016deep}
I.~Goodfellow, Y.~Bengio, and A.~Courville, \emph{Deep learning}.\hskip 1em
  plus 0.5em minus 0.4em\relax MIT press, 2016.

\bibitem{litjens2017survey}
G.~Litjens, T.~Kooi, B.~E. Bejnordi, A.~A.~A. Setio, F.~Ciompi \emph{et~al.},
  ``A survey on deep learning in medical image analysis,'' \emph{Medical image
  analysis}, vol.~42, pp. 60--88, 2017.

\bibitem{gu2018recent}
J.~Gu, Z.~Wang, J.~Kuen, L.~Ma, A.~Shahroudy, B.~Shuai, T.~Liu, X.~Wang,
  G.~Wang, J.~Cai \emph{et~al.}, ``Recent advances in convolutional neural
  networks,'' \emph{Pattern Recognition}, vol.~77, pp. 354--377, 2018.

\bibitem{pouyanfar2018survey}
S.~Pouyanfar, S.~Sadiq, Y.~Yan, H.~Tian, Y.~Tao, M.~P. Reyes, M.-L. Shyu, S.-C.
  Chen, and S.~Iyengar, ``A survey on deep learning: {A}lgorithms, techniques,
  and applications,'' \emph{Computing Surveys}, vol.~51, no.~5, pp. 1--36,
  2018.

\end{thebibliography}

\vspace{0cm}
\begin{IEEEbiographynophoto}{Delu~Zeng} received his Ph.D. degree in electronic and information engineering from South China University of Technology, China, in 2009. He is now a full professor in the School of Mathematics in South China University of Technology, China. He has been the visiting scholar of Columbia University, University of Oulu, University of Waterloo. He has been focusing his research in applied mathematics and its interdisciplinary applications. His research interests include numerical calculations, applications of partial differential equations, optimizations, machine learning and their applications in image processing, and data analysis. 
\end{IEEEbiographynophoto}
\vspace{0cm}
\begin{IEEEbiographynophoto}{Minyu~Liao} received her B.S. degree in applied mathematics from Shantou University, China, in 2018. She is currently pursuing the master's degrees in computational mathematics with South China University of Technology, China. Her research interests include computer vision, scene recognition, and deep learning.
\end{IEEEbiographynophoto}

\vspace{0cm}
\begin{IEEEbiographynophoto}{Mohammad~Tavakolian}
received the M.Sc. degree in electrical engineering from Tafresh University, Iran in 2013. Currently, he is a Ph.D. student at the Center for Machine and Signal Analysis (CMVS) of the University of Oulu, Finland. He has authored several journal and conference papers in IJCV, PRL, ICCV, ECCV, and ACCV. His research interests include representation learning, data efficient learning, computer vision, healthcare, and face analysis. 
\end{IEEEbiographynophoto}
\vspace{0cm}

\begin{IEEEbiographynophoto}{Yulan~Guo} He received his Ph.D. degrees from National University of Defense Technology (NUDT) in 2015, where he is currently an associate professor. He was a visiting Ph.D. student with the University of Western Australia from 2011 to 2014. He has authored over 90 articles in journals and conferences, such as the IEEE TPAMI and IJCV. His current research interests focus on 3D vision, particularly on 3D feature learning, 3D modeling, 3D object recognition, and scene understanding. 
\end{IEEEbiographynophoto}

\vspace{0cm}

\begin{IEEEbiographynophoto}{Bolei~Zhou} is currently an assistant professor at the Chinese University of Hong Kong, China. He received his Ph.D. from the Massachusetts Institute of Technology in 2018. He received his M.phil from the Chinese University of Hong Kong and B.Eng. degree from the Shanghai Jiao Tong University in 2010. He has authored over 70 articles in journals and conferences, such as IEEE TPAMI, ECCV, CVPR and AAAI. He is interested in understanding various human-centric properties of AI models beyond their performance, such as explainability, interpretability, steerability, generalization, fairness and bias.

\end{IEEEbiographynophoto}
\vspace{0cm}
\begin{IEEEbiographynophoto}{Dewen~Hu} received the B.S. and M.S. degrees from Xi’an Jiaotong University, China, in 1983 and 1986, respectively, and the Ph.D. degree from the National University of Defense Technology in 1999. He is currently a Professor at School of Intelligent Science, National University of Defense Technology. From October 1995 to October 1996, he was a Visiting Scholar with the University of Sheffield, U.K. His research interests include image processing, system identification and control, neural networks, and cognitive science.
\end{IEEEbiographynophoto}
\vspace{0cm}
\begin{IEEEbiographynophoto}{Matti~Pietik\"{a}inen}
received his Ph.D. degree from the University of Oulu, Finland. He is now emeritus professor with the Center for Machine Vision and Signal Analysis, University of Oulu. He is a fellow of the IEEE for fundamental contributions, \eg, to Local Binary Pattern (LBP) methodology, texture based image and video analysis, and facial image analysis. He has authored more than 350 refereed papers in international journals, books, and conferences. His papers have nearly 68,700 citations in Google Scholar (h-index 92). He was the recipient of the IAPR King-Sun Fu Prize 2018 for fundamental contributions to texture analysis and facial image analysis. 
\end{IEEEbiographynophoto}

\vspace{0cm}
\begin{IEEEbiographynophoto}{Li~Liu}
received the Ph.D. degree in information and communication engineering from the National University of Defense Technology (NUDT), China, in 2012. She is currently a professor with NUDT. She spent two years as a Visiting Student at the University of Waterloo, Canada, from 2008 to 2010. From 2015 to 2016, she spent ten months visiting the Multimedia Laboratory at the Chinese University of Hong Kong. From 2016.12 to 2018.9, she worked as a senior researcher at the Machine Vision Group at the University of Oulu, Finland. Her current research interests include computer vision, pattern recognition and machine learning. Her papers have currently over 3500+ citations in Google Scholar.
\end{IEEEbiographynophoto}

\end{document}